\documentclass[journal]{IEEEtran}
\ifCLASSINFOpdf
\else
\fi
\usepackage{url}  
\usepackage{graphicx}  

\usepackage{booktabs}       
\usepackage{amsmath}
\usepackage{amsfonts}
\usepackage{bm}
\usepackage{algorithm}
\usepackage{algorithmic}
\usepackage{multirow}
\usepackage{subfigure}
\usepackage{xcolor}
\usepackage{cases}

\begin{document}
%
\title{Underexposed Image Correction via Hybrid Priors Navigated Deep Propagation}
%
%
%

\author{Risheng Liu,~\IEEEmembership{Member,~IEEE,}
        Long Ma,
    	Yuxi Zhang,
        Xin Fan, ~\IEEEmembership{Member,~IEEE,}
    	and Zhongxuan Luo
	\thanks{R. Liu, L. Ma, Y. Zhang, X. Fan, and Z. Luo are with the DUT-RU International School
	of Information Science \& Engineering, Dalian University of Technology,
	and also with the Key Laboratory for Ubiquitous Network and
	Service Software of Liaoning Province, Dalian 116024, China. E-mail:
	rsliu@dlut.edu.cn, malone94319@gmail.com, yuxizhang@mail.dlut.edu.cn, xin.fan@dlut.edu.cn, zxluo@dlut.edu.cn.}
\thanks{Manuscript received April 19, 2005; revised August 26, 2015.}}

%
%

\markboth{Journal of \LaTeX\ Class Files,~Vol.~14, No.~8, August~2015}%
{Shell \MakeLowercase{\textit{et al.}}: Bare Demo of IEEEtran.cls for IEEE Journals}
%



\maketitle

\begin{abstract}
Enhancing visual qualities for underexposed images is an extensively concerned task that plays important roles in various areas of multimedia and computer vision. Most existing methods often fail to generate high-quality results with appropriate luminance and abundant details. To address these issues, we in this work develop a novel framework, integrating both knowledge from physical principles and implicit distributions from data to solve the underexposed image correction task.
More concretely, we propose a new perspective to formulate this task as an energy-inspired model with advanced hybrid priors. A propagation procedure navigated by the hybrid priors is well designed for simultaneously propagating the reflectance and illumination toward desired results. We conduct extensive experiments to verify the necessity of integrating both underlying principles (i.e., with knowledge) and distributions (i.e., from data) as navigated deep propagation.
Plenty of experimental results of underexposed image correction demonstrate that our proposed method performs favorably against the state-of-the-art methods on both subjective and objective assessments. Additionally, we execute the task of face detection to further verify the naturalness and practical value of underexposed image correction.
What's more, we employ our method to single image haze removal whose experimental results further demonstrate its superiorities.
\end{abstract}

\begin{IEEEkeywords}
Deep learning; Hybrid priors; Underexposed image correction; Face detection.
\end{IEEEkeywords}

%
\IEEEpeerreviewmaketitle

\section{Introduction}

High-visibility images with sufficient details of target scenes are quite essential for many multimedia and computer vision applications. However, the captured images often suffer from low-visibility due to nighttime, backlighting or some light limited scenes in most real-world scenarios. Thus, underexposed image correction is generally demanded in many practical fields.
During the past few years, various underexposed image correction techniques have been proposed. 

In the early stage, researchers tend to design the methods based on histogram modification~(\cite{wang2005Brightness,sheet2010brightness}) for underexposed image correction. This type of technique indeed improves the luminance, but it cannot work well for non-uniform illumination. Retinex-based image decomposition approaches~(\cite{guo2017lime,zhang2018high,li2018structure}) are widely used for this task at the present stage, which follow the physical law so that they achieve excellent performance. However, since they always need to design the complex prior regularization to narrow down the solution, not only increasing the time-consuming, but also resulting in the insufficient improvement of luminance and depict of details. Undoubtedly, network-based techniques~(\cite{hasinoff2017Deep,Chen2018Retinex}) can be directly adopted to settle this task. Unfortunately, the difficulty of obtaining training pairs limits the development of deep network in this task. Actually, most of existing related network-based works tend to generate the unnaturalness enhanced results in real scenarios.
\begin{figure*}[t]
	\centering
	\begin{tabular}{c@{\extracolsep{0.3em}}c@{\extracolsep{0.3em}}c@{\extracolsep{0.3em}}c@{\extracolsep{0.3em}}c}
		\includegraphics[width=0.19\linewidth]{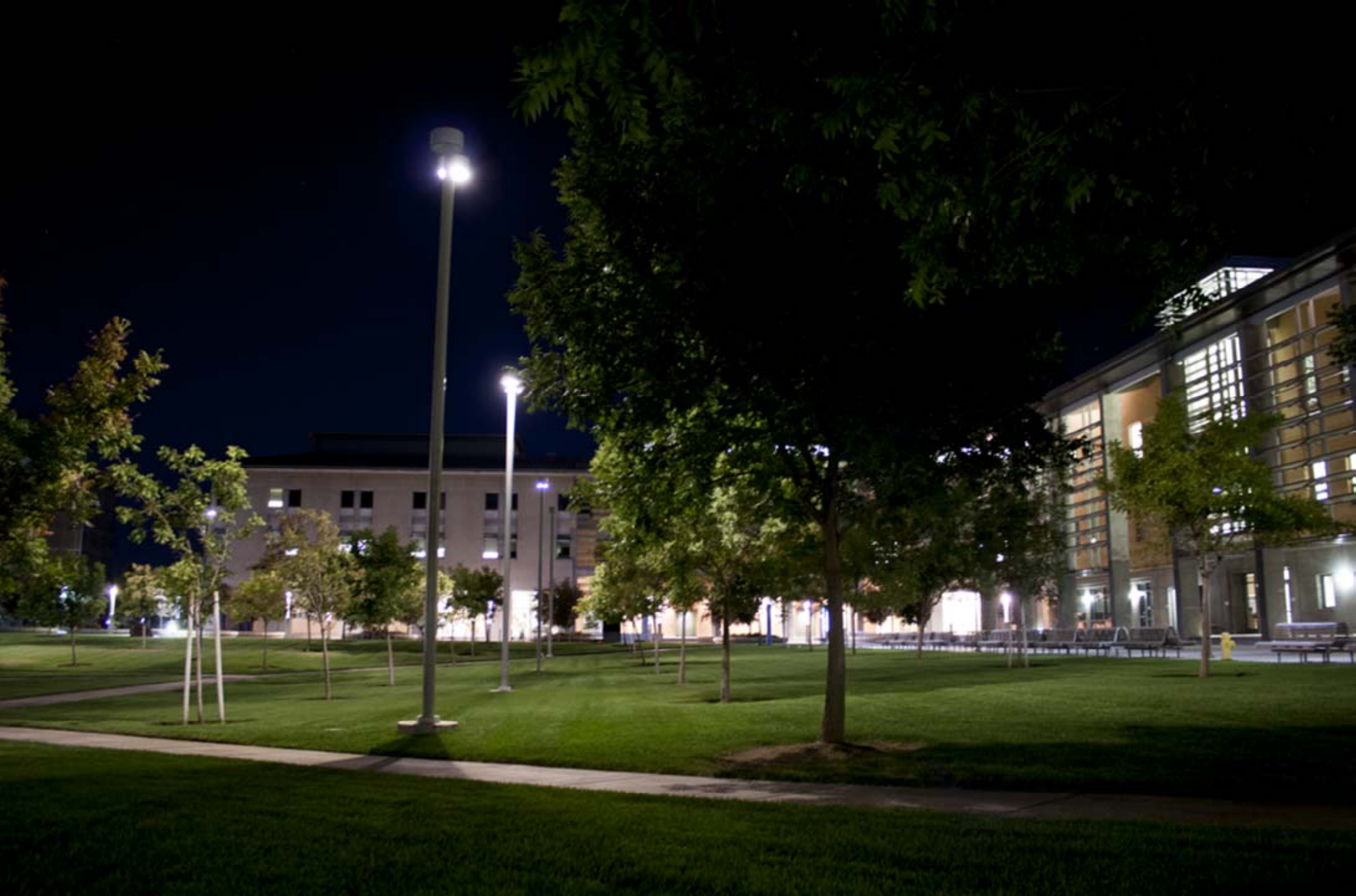}&
		\includegraphics[width=0.19\linewidth]{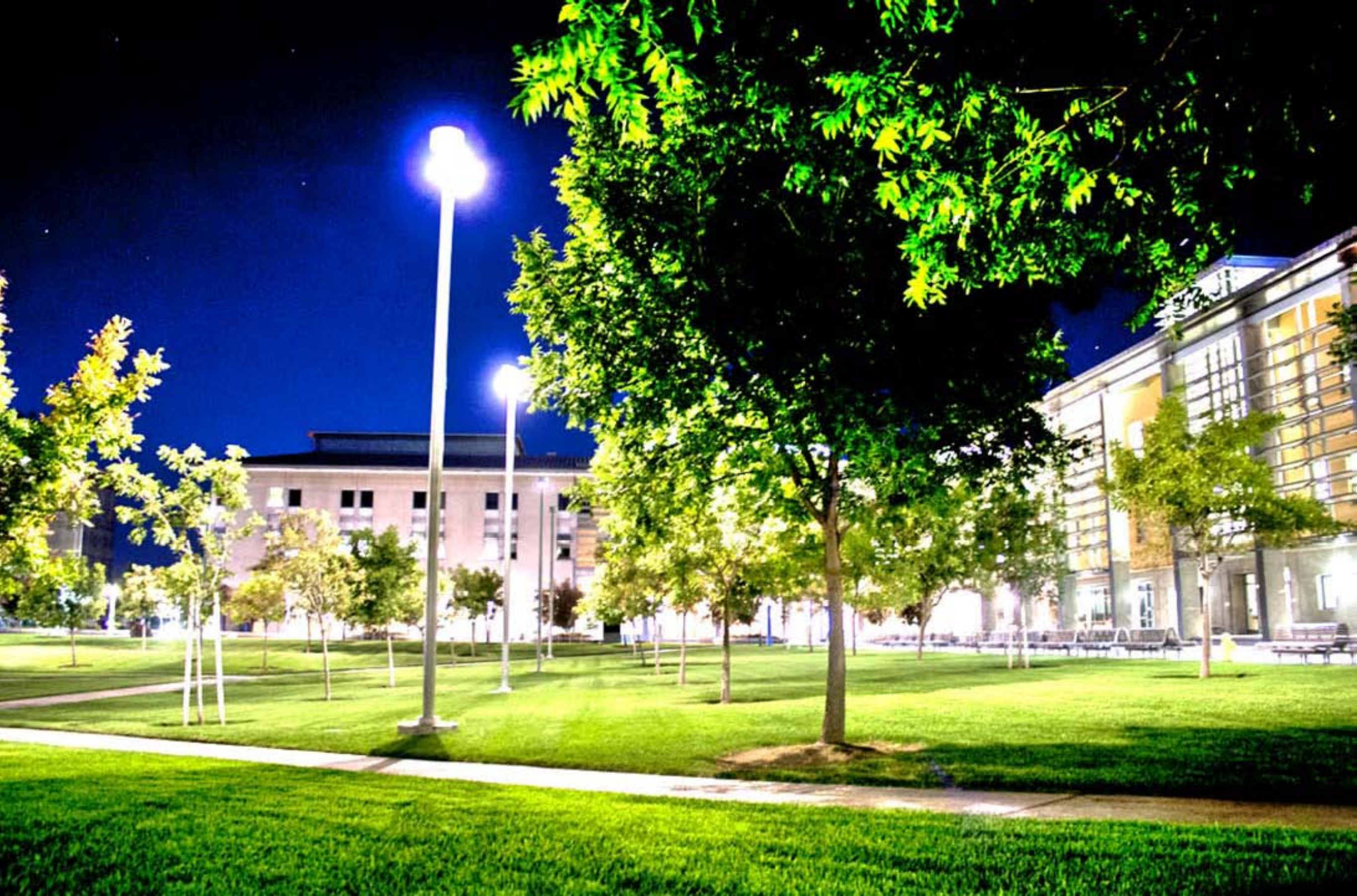}&
		\includegraphics[width=0.19\linewidth]{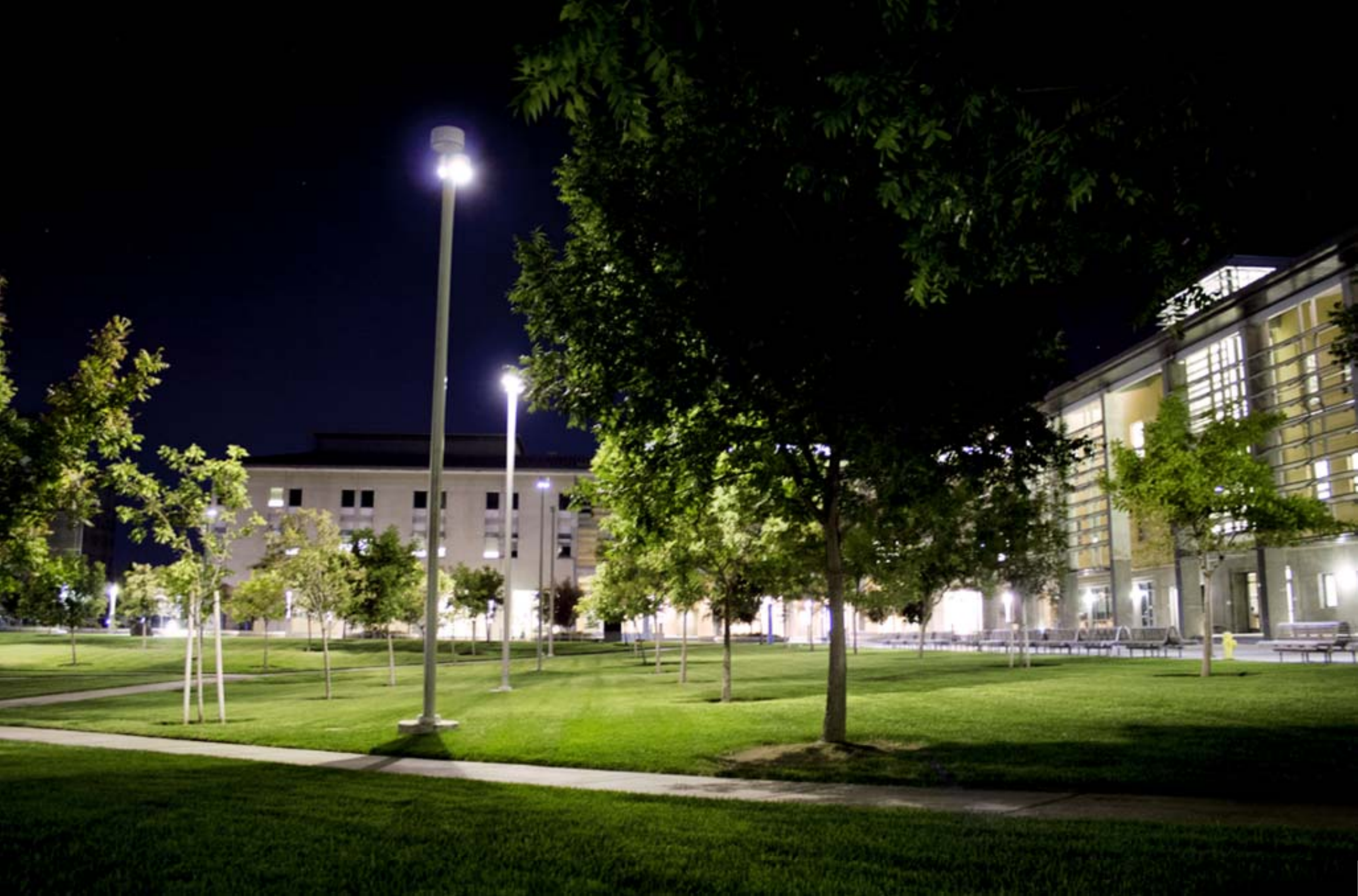}&
		\includegraphics[width=0.19\linewidth]{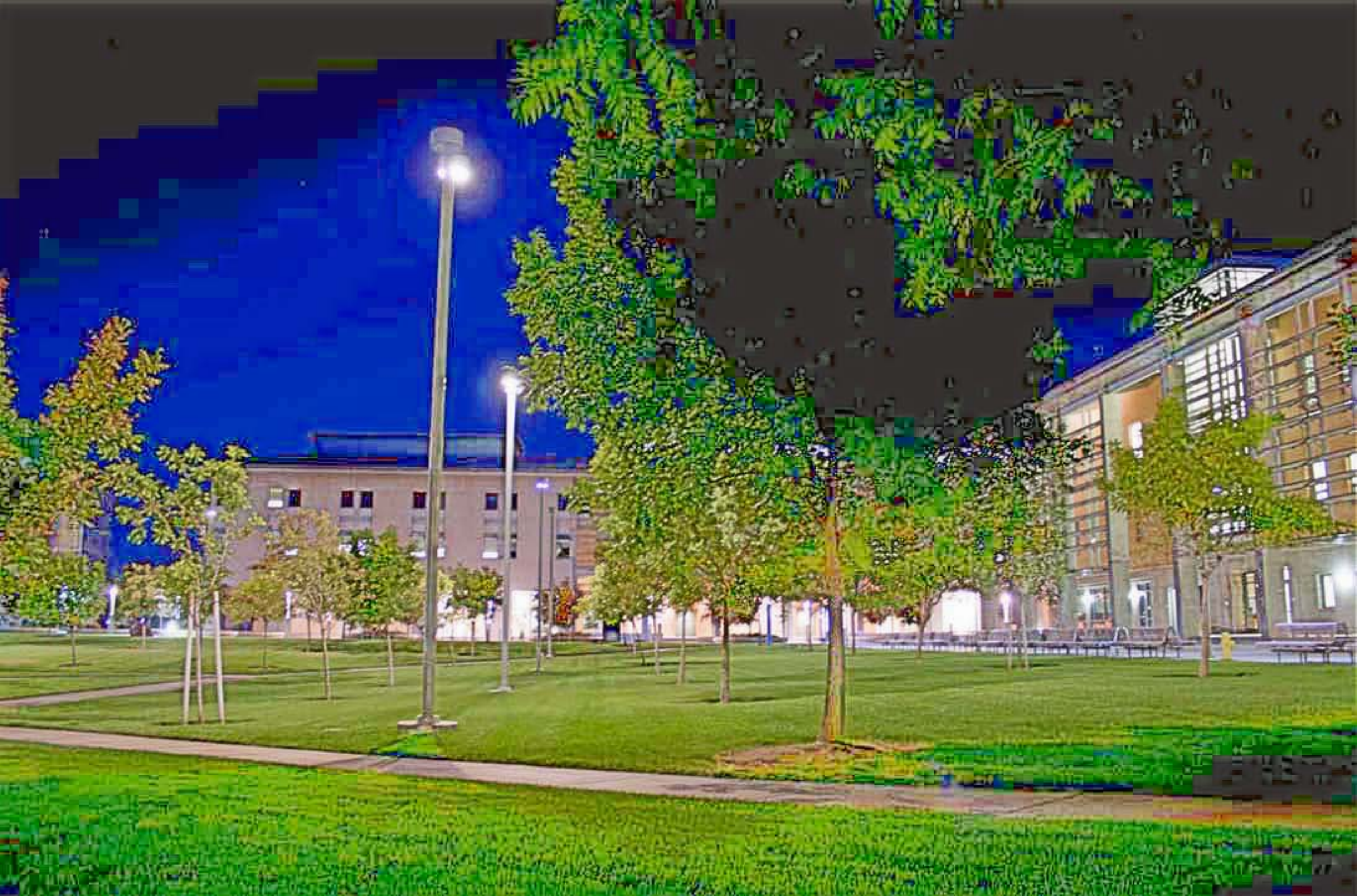}&
		\includegraphics[width=0.19\linewidth]{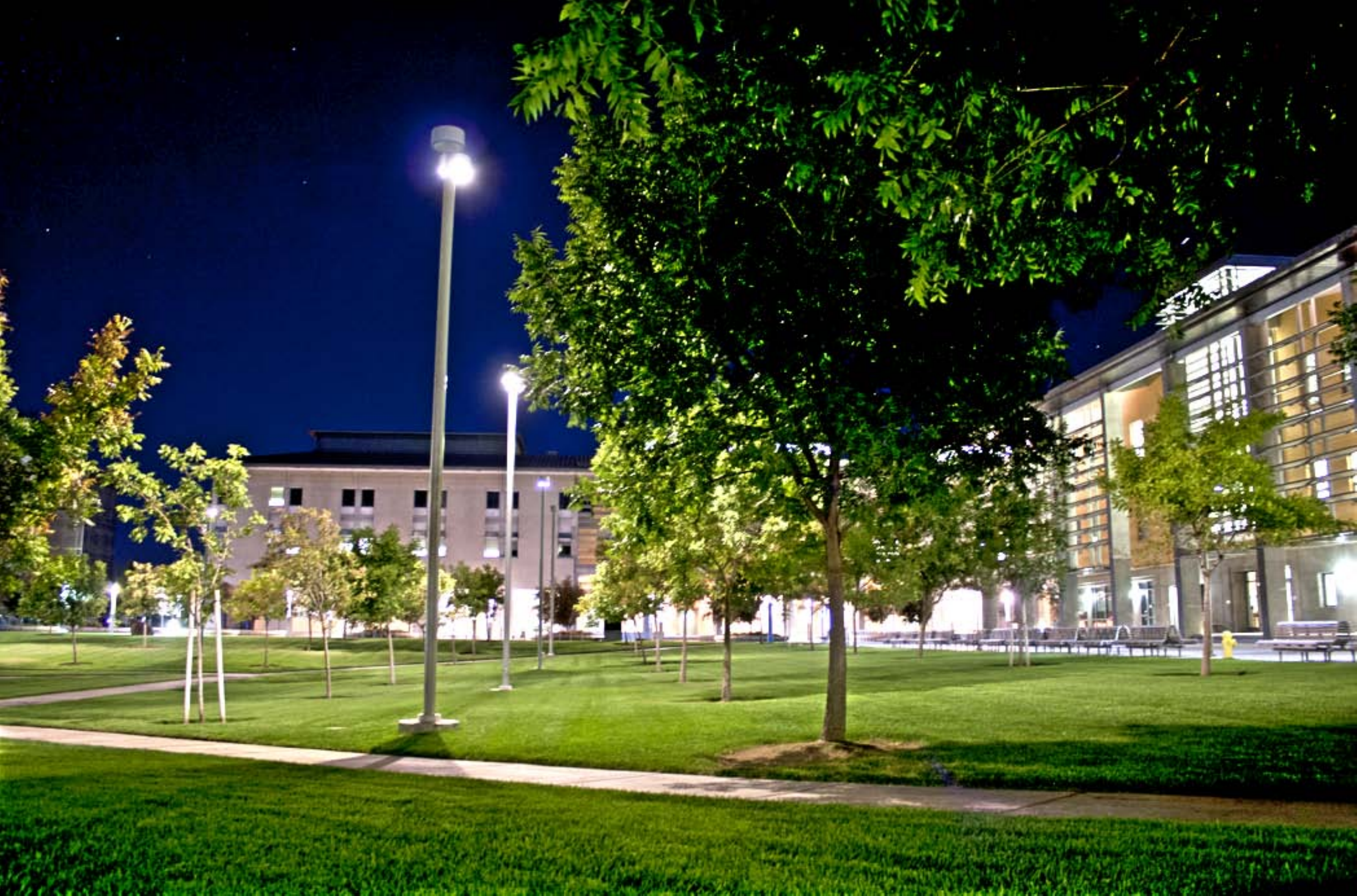}\\
		(a) Input &(b) LIME&(c) HDRNet & (d)  RetinexNet & (e) Ours \\
	\end{tabular}
	\caption{Underexposed image correction results comparison on an example image.  It can be seen that there exists severe overexposure and some details cannot be recovered in the result of LIME~\cite{guo2017lime} (Retinex decomposition based method). Most of details in the dark regions cannot be properly recovered by HDRNet~\cite{hasinoff2017Deep} (end-to-end network). The worst visual expression of RetinexNet~\cite{Chen2018Retinex} (Retinex-based end-to-end network). By comparison, we obtain the best visual performance with clear details and appropriate contrast among these compared methods.}
	\label{fig:FirstFig}
\end{figure*}
\subsection{Our Contributions}

As discussed above, to compensate for the ill-posedness of the image decomposition, strong priors for both the reflectance and illumination are required to regularize the solution space. However, designing such exact priors in hand-crafted manner is challenging and needs extremely high mathematical skills. More importantly, the purely designed priors may only suitable for the data with given distributions, thus limit their applications in more complex real-world scenarios (see Fig.~\ref{fig:FirstFig} (b)). Additionally, due to the lack of exact references for training, the networks learned by the end-to-end manners may hard to enhance all these details in the dark regions (see Fig.~\ref{fig:FirstFig} (c)) or generate the unnatural result (see Fig.~\ref{fig:FirstFig} (d)). 

In this work, we propose a novel underexposed image correction framework, in which the domain knowledge and training data are integrated to generate the hybrid priors for Retinex decomposition.
Specifically, we establish a generic energy-inspired deep propagation framework, based on the image decomposition model in Eq.~\eqref{eq:bmodel}. By introducing a schematic alternating half-quadratic splitting scheme, the fundamental propagations of the reflectance and illumination are established. To navigate the coupled iterations towards the desired solutions, we develop the hybrid priors, which consist of both explicitly designed distribution constraints (i.e., knowledge) and implicitly trained deep architectures (i.e., data). The advantage of our proposed methodology is initially verified by comparing it with one representative decomposition-based method (i.e., LIME~\cite{guo2017lime}) and two end-to-end discriminative learning approaches (i.e., HDRNet~\cite{hasinoff2017Deep}, RetinexNet~\cite{Chen2018Retinex}) on an example image in Fig.~\ref{fig:FirstFig}. 

The main contributions of our proposed method can be summarized in the following four aspects:
\begin{itemize}
	\item We provide a generic energy-inspired deep propagation perspective to formulate the image decomposition problem. It will be demonstrated that both underexposed image correction and other related vision tasks (e.g., dehazing) can be addressed within this framework. 
	\item With the flexibility of our proposed framework, we can successfully combine domain knowledge (e.g., Retinex principle, structure priors) and data-dependent architectures (e.g., learnable descent directions) to navigate the propagations of the coupled image component.
	\item To fully indicate the naturalness and practical value of our proposed method, we not only conduct extensive experiments on challenging underexposed images, but also execute the face detection to further manifest the naturalness and practical values of our method. 
	\item To evaluate the scalability of our built framework, the task of single image haze removal is considered. We present visual comparison on some challenging hazy images in real-world scenarios, which indicates our superiority.  
\end{itemize}

\begin{figure*}[t]
	\centering
	\begin{tabular}{c}
		\includegraphics[width=0.98\linewidth]{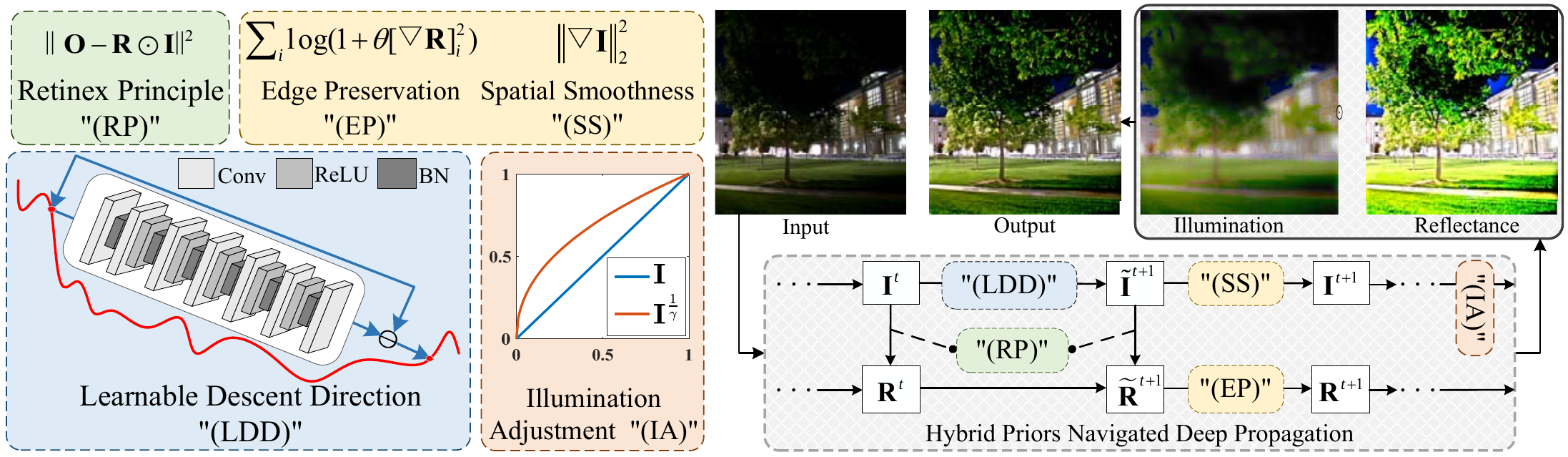}\\
	\end{tabular}
	\caption{The illustration of our propagations with hybrid priors navigation. The four dashed rectangles of left column is our core principles of designing our method. The right bottom dashed rectangle represents the iteration block in \textit{t}-th stage, i.e., the details of our algorithm. The right top row is the visual comparison of underexposed input, the enhanced output and obtained decomposition components.}
	\label{fig:Flow}
\end{figure*}

\section{Related Works}
In this section, we state a brief review of the related works. Generally, existing underexposed image correction techniques can be divided into three categories: the methods based on histogram modification, image decomposition and discriminate learning.

\textbf{Histogram Modification:} 
The histogram-based methods make efforts to modify the histogram distributions to recall visibilities of dark regions. 
Histogram equalization~\cite{cheng2004simple} is one of the most commonly used histogram modification techniques. However, it tends to result in over-enhancement. Different constraints have also been designed for brightness preservation~\cite{wang2005Brightness,sheet2010brightness,power2011Multi} and weight adjustment~\cite{yun2011Contrast}. However, these methods cannot work well for non-uniform illuminations.

\textbf{Image Decomposition:}
Retinex theory assumes that the scene in human's eyes is the product of reflectance and illumination layers~\cite{McCann2016},
in which illumination represents the light intensity and reflectance denotes the physical characteristic of objects~\cite{cai2017joint}. Given the observation $\mathbf{O}\in\mathbb{R}^{N}$,
this model can be formulated as 
\begin{equation}
\mathbf{O} = \mathbf{R}\odot\mathbf{I},
\label{eq:bmodel}
\end{equation}
where ``$\odot$" denotes pixel-wise multiplication, and $\mathbf{R}, \mathbf{I}\in\mathbb{R}^{N}$ are the reflectance and illumination parts, respectively. With basic physical principles, it is also necessary to assume that the pixel values of $\mathbf{R}$ and $\mathbf{I}$ are in definite ranges, i.e., $\mathbf{R}\in\Omega_{\mathbf{R}}:=\{\mathbf{R}|0\leq \mathbf{R}_i\leq 1, i=1,...,N\}$ and $\mathbf{I}\in\Omega_\mathbf{I}:=\{\mathbf{I}|0\leq\mathbf{I}_{i}\leq\mathbf{O}_i, i=1,...,N\}$.

Although with the additional value constraints, the  decomposition above is highly ill-posed~\cite{land1971lightness}; thus strong priors are required for both the reflectance and illumination to regularize the solution space.
The work in~\cite{kimmel2003variational} adopted $\ell_2$ regularizer to estimate the illumination in the logarithmic domain. \cite{ng2011total} adopted Total Variational (TV) based model in the logarithmic domain for intrinsic image decomposition. 
Fu \emph{et al.} directly designed
probabilistic formulations to simultaneously estimate reflectance and illumination in the image domain~\cite{fu2015probabilistic}. Furthermore, they considered a weighted variational decomposition formulation in the logarithmic domain~\cite{fu2016weighted}.
In~\cite{guo2017lime}, the illumination is refined by only preserving the main contour based on an initial illumination. Similarly, the paper~\cite{zhang2018high} proposed a perceptually bidirectional similarity to produce natural-looking results based on the illumination optimization.
The work in ~\cite{cai2017joint} combined different priors to build a complex energy model. It can be seen that these methods just adopt contrived priors to constrain their decompositions. However, it is hard to utilize human-designed priors to investigate the intrinsic structures of the underlying illumination and reflectance, especially in the image domain. This is because we are actually still not sure about the exact distributions of these latent components.


\textbf{Discriminative Learning:} Very recently, discriminative learning based methods are proposed. Especially like Convolutional Neural Networks (CNNs), which has been demonstrated that CNNs can learn realistic natural image distributions from a number of images~\cite{dmitry2018Deep}. Thus, several approaches have been proposed to apply the implicit CNN priors for low-level vision tasks, such as super-resolution~(\cite{kim2015Accurate,zhang2018residual}), deconvolution~(\cite{liu2018Proximal,liu2018GCM}), dehazing~(\cite{li2017aod}), and others~(\cite{liu2018learning,liu2019deep,liu2019convergence}). 
However, since there exist highly coupled variables and complex constraints, which cause that both synthetic and real-world datasets are all obtained difficult, thus it is extremely challenging to apply CNNs to inference the decomposition models in Eq.~\eqref{eq:bmodel}.
Up to now, there indeed exist some CNN-based approaches for addressing underexposed image correction task (\cite{hasinoff2017Deep,Chen2018Retinex, lore2017llnet,cai2018Learning, Zhang2019MM}). Actually, training dataset becomes the key restraints for the practical performance of these works. In~\cite{hasinoff2017Deep}, professional photographers are employed to generate the training pairs. \cite{lore2017llnet} proposes to train the designed network using the synthetic dataset generated from the operator of Gamma Correction. Considering the multi-exposure images, the work in~\cite{cai2018Learning} builds a large scale multi-exposure image dataset, and generates high-quality reference images based on 13 MEF and HDR algorithms for network training. The turning point arises in the work~\cite{Chen2018Retinex} which builds a new dataset, i.e. LOw-Light dataset (LOL), which is generated by adjusting the exposure time. This paper also proposes an end-to-end Retinex-based deep network which combines the Retinex theory and learnable architecture. Lately, a practical network-based algorithm is proposed in~\cite{Zhang2019MM} based on LOL dataset. Overall, these network-based approaches all tend to generate the unnaturalness performance with insufficient details and under/over exposure. The reason is that training pairs are inaccurate to depict the real distribution, e.g., LOL just considers the exposure time, while there exist many physical factors (e.g., illumination condition) in real scenarios.

In summary, training pairs are hard to generate using existing techniques for underexposure image correction, which severely limits the development of deep learning in this area. So it is intuitive to consider the learnable architecture in some deductive ways, to skip the difficulty of directly generating training pairs for this task. 

\section{The Proposed Framework}
Most existing decomposition-based models \cite{kimmel2003variational,ng2011total,fu2016weighted} are based on the logarithmic transformation. However, the side effect is that these undesired structures are amplified in the low magnitude stimuli areas and edges may become fuzzy. Therefore, in this work we directly formulate our Retinex decomposition in the image domain as the following regularized variational energy model:
\begin{equation}
	\min_{\mathbf{R}\in\Omega_\mathbf{R},\mathbf{I}\in\Omega_\mathbf{I}}f(\mathbf{I},\mathbf{R})+ \Phi(\mathbf{I})+\Psi(\mathbf{R}),
	\label{eq:energy}
\end{equation}
where $f(\mathbf{I},\mathbf{R}) = \frac{1}{2}\|\mathbf{R}\odot\mathbf{I} - \mathbf{O}\|_2^2$ is the fidelity term derived from the model Eq.~\eqref{eq:bmodel}, $\Phi$ and $\Psi$ are the prior regularization terms of illumination and reflectance, respectively. 

Notice that different from most existing decomposition-based methods, which only design the prior penalties based on their intuitions, we provide a new way to integrate knowledge and data to obtain more efficient hybrid priors for our decomposition problem in the next section.


\subsection{Hybrid Priors Navigated Deep Propagation}

Since the expression of the explicit formulations of $\Phi$ and $\Psi$ in Eq.~\eqref{eq:energy} is hard to obtain, it is indeed challenging to adopt standard iteration schemes to optimize this variational energy. Next, we will provide a new propagation framework to integrate alternating half-quadratic splitting scheme and hybrid priors to respectively obtain our desired illumination and reflectance.
\subsubsection{Illumination Propagation}
We first consider the prior term of illumination as the combination of one principled assumption (i.e., spatial smoothness) and one implicit data-dependent term as follows:
\begin{equation}
	\Phi(\mathbf{I})=\frac{\mu_\mathbf{I}}{2}\|\nabla \mathbf{I}\|_2^2 +\mathcal{D}_{\mathbf{I}}(\mathbf{I}),
	\label{eq:Imodel}
\end{equation}
where $\mu_\mathbf{I}$ is a trade-off parameter, $\|\nabla \mathbf{I}\|_2^2$ enforces the smooth constraint and $\mathcal{D}_{\mathbf{I}}$ denotes our implicit prior submodule (learned from data). 

Then utilizing half-quadratic splitting technique with an auxiliary variable $\tilde{\mathbf{I}}$ (with penalty parameter $\lambda_\mathbf{I}^{t}$), we have the following subproblem for illumination updating
\begin{equation}
	\begin{array}{l}
		(\mathbf{I}^{t+1},\tilde{\mathbf{I}}^{t+1})=\arg\min\limits_{\mathbf{I}\in\Omega_{\mathbf{I}}, \tilde{\mathbf{I}}} f(\mathbf{I},\mathbf{R}^t) + \frac{\mu_\mathbf{I}}{2}\|\nabla \mathbf{I}\|_2^2\\
		\qquad\qquad\qquad +\mathcal{D}_{\mathbf{I}}(\tilde{\mathbf{I}}) + \frac{\lambda_{\mathbf{I}}^{t}}{2}\|\mathbf{I}-\tilde{\mathbf{I}}\|^2.
	\end{array}
\end{equation}
Rather than explicitly formulating and calculating $\mathcal{D}_{\mathbf{I}}$, we directly update the auxiliary variable $\tilde{\mathbf{I}}^{t+1}$ via the following learnable descent scheme 
\begin{equation}
	\tilde{\mathbf{I}}^{t+1}=\mathbf{{I}}^{t} - \mathcal{N}(\mathbf{{I}}^{t};\Theta),
	\label{eq:Itilde}
\end{equation}
where $\mathcal{N}$ denotes the parameterized descent directions (e.g., CNN architectures) with parameters $\Theta$. We will discuss the details of these learnable architectures in the next part. It should be emphasized that
we actually provide a way to learn the guidance from training data to navigate our illumination propagation. Then we are ready to update $\mathbf{I}$ for the $t+1$-th stage.
By further reformulating the fidelity $\|\mathbf{R}\odot\mathbf{I} - \mathbf{O}\|_2^2$ as $\|\mathbf{I} - \frac{\mathbf{O}}{\mathbf{R}}\|_2^2$, we can obtain the updating scheme of $\mathbf{I}$ as
\begin{equation}
	\mathbf{I}^{t+1}=\mathcal{P}_{\Omega_{\mathbf{I}}}\left(\frac{\frac{\mathbf{O}}{\mathbf{{R}}^{t}}+\lambda_\mathbf{I}^{t}\tilde{\mathbf{I}}^{t+1}}{\lambda_\mathbf{I}^{t}+\mu_\mathbf{I}\nabla^{\top}\nabla+1}\right),\label{eq:solveI}
\end{equation}	
where $\mathcal{P}_{\Omega_{\mathbf{I}}}$ denotes the projection on $\Omega_{\mathbf{I}}$. 

\subsubsection{Reflectance Propagation}
As for the reflectance $\mathbf{R}$, we would like to preserve the sharp edge structure of the reflectance during the enhancement process. Thus we consider the following hybrid regularization term $\Psi$ as
\begin{equation}
	\Psi(\mathbf{R})=\frac{\mu_\mathbf{R}}{2}\sum\limits_i\log(1+\theta[\nabla \mathbf{R}]_i^2) +\mathcal{D}_{\mathbf{R}}(\mathbf{R}),
	\label{eq:Rmodel}
\end{equation}
where the first term is a widely used non-convex potential function (with a sparsity controlled parameter $\theta$, can be used to reveal the sharp edge structure)~\cite{Roth2009Fields}, $\mu_\mathbf{R}$ is the trade-off parameter and $\mathcal{D}_{\mathbf{R}}$ denotes the data-dependent prior for reflectance.
Here $[\nabla \mathbf{R}]_i$ denotes the $i$-th element of $\nabla\mathbf{R}$.
Using half-quadratic reformulation technique, we can obtain the $\mathbf{R}$ subproblem as
\begin{equation}
	\begin{array}{l}
		(\mathbf{R}^{t+1},\tilde{\mathbf{R}}^{t+1})=\arg\min\limits_{\mathbf{R}\in\Omega_{\mathbf{R}},\tilde{\mathbf{R}}}f(\mathbf{I}^{t+1},\mathbf{R})\\\quad+\frac{\mu_\mathbf{R}}{2}\sum\limits_i\log(1+\theta[\nabla \mathbf{R}]_i^2) +\mathcal{D}_{\mathbf{R}}(\tilde{\mathbf{R}}) + \frac{\lambda_{\mathbf{R}}^{t}}{2}\|\mathbf{R}-\tilde{\mathbf{R}}\|^2,
	\end{array}\label{eq:r-problem}
\end{equation}
where $\tilde{\mathbf{R}}$ is an auxiliary variable and $\lambda_{\mathbf{R}}^{t}$ is the penalty parameter.

Intuitively, we may follow the idea in $\tilde{\mathbf{I}}$-subproblem to introduce another network to calculate $\tilde{\mathbf{R}}$. However, by recalling the physical rule in Eq.~\eqref{eq:bmodel}, we can obtain a much simpler updating scheme for $\tilde{\mathbf{R}}$ as
\begin{equation}
	\tilde{\mathbf{R}}^{t+1}= \frac{\eta\frac{\mathbf{O}}{\mathbf{\tilde{I}}^{t+1}}+\mathbf{{R}}^{t}}{\eta+1}, \label{eq:Rtilde}
\end{equation}
where $\eta$ denotes the weight coefficient.

However, due to the non-convex potential function, we cannot obtain closed-form solution of the $\mathbf{R}$-subproblem in Eq.~\eqref{eq:r-problem}. Thus, we adopt a projected gradient type rule to update $\mathbf{R}$ as following:
\begin{equation}
	\mathbf{R}^{t+1}=\mathcal{P}_{\Omega_{\mathbf{R}}}\left(\mathbf{R}^{t} - \nabla_{\mathbf{R}} \mathit{g}(\mathbf{I}^{t+1},\mathbf{R}^{t},\tilde{\mathbf{R}}^{t+1},{\lambda_\mathbf{R}^{t}})\right),\label{eq:solveR}
\end{equation}
where  $\mathit{g}(\mathbf{I},\mathbf{R},\tilde{\mathbf{R}},{\lambda_\mathbf{R}})=f(\mathbf{{R}},\mathbf{{I}})+\frac{\lambda_\mathbf{R}^{t}}{2}\|\tilde{\mathbf{R}}-\mathbf{R}\|_2^2+\frac{\mu_\mathbf{R}}{2}\sum_i\log(1+\theta[\nabla \mathbf{R}]_i^2) $ and $\mathcal{P}_{\Omega_{\mathbf{R}}}$ is the projection on $\Omega_{\mathbf{R}}$.

\begin{algorithm}[t]
	\caption{Underexposed Image Correction via Hybrid Priors Navigated Deep Propagation}
	\label{alg:enhancement}
	\begin{algorithmic}[1]
		\STATE
		\textbf{Input:} $\mathbf{O}$, and some necessary parameters.\\
		\STATE Initialization: $\mathbf{I}^0=\mathbf{O}$, $\mathbf{R}^{0}=\mathbf{0}$.
		\FOR{$t=0:t_{\max}-1$}
		\STATE 
		Update $\tilde{\mathbf{I}}^{t+1}$ using Eq.~\eqref{eq:Itilde}.
		\STATE 
		Update $\mathbf{I}^{t+1}$ using Eq.~\eqref{eq:solveI}.
		\STATE 
		Update $\tilde{\mathbf{R}}^{t+1}$ using Eq.~\eqref{eq:Rtilde}.
		\STATE 
		Update $\mathbf{R}^{t+1}$ using Eq.~\eqref{eq:solveR}.
		\ENDFOR
		\STATE
		Obtain final enhanced result $\mathbf{O}_{enhanced}$ (i.e., Eq.~\eqref{eq:enhanced}).\\
		\STATE
		\textbf{Output:} $\mathbf{O}_{enhanced}$.
	\end{algorithmic}\label{alg:dhp}
\end{algorithm}

\subsubsection{Illumination Adjustment}

It is known that the illumination contains the lightness information. So the underexposed image correction task now reduces to the problem of adjusting the illumination to generate high-visually reconstructions. Gamma correction is a common measure to encode and decode the luminance by taking advantage of the non-linear manner in which humans perceive light and color~\cite{poynton2012digital}.
So we adopt the following Gamma correction operation to adjust our obtained illumination:
\begin{equation}
	\mathbf{O}_{e}= \mathbf{R}\odot\mathbf{I}^{\frac{1}{\gamma}}
	\label{eq:enhanced}
\end{equation}
where $\mathbf{O}_{e}$ denotes the final enhanced result. $\gamma>0$ is a tunning parameter (empirically designed as 2.2). 
Now we are ready to summarize our algorithm in Alg.~\ref{alg:enhancement} and Fig.~\ref{fig:Flow}.
\begin{figure}[t]
	\begin{tabular}{c@{\extracolsep{0.1em}}c@{\extracolsep{0.1em}}c@{\extracolsep{0.1em}}c}
		{Illumination}&{Reflectance}&{Output}&{Zoomed-in}\\
		\includegraphics[width=0.11\textwidth]{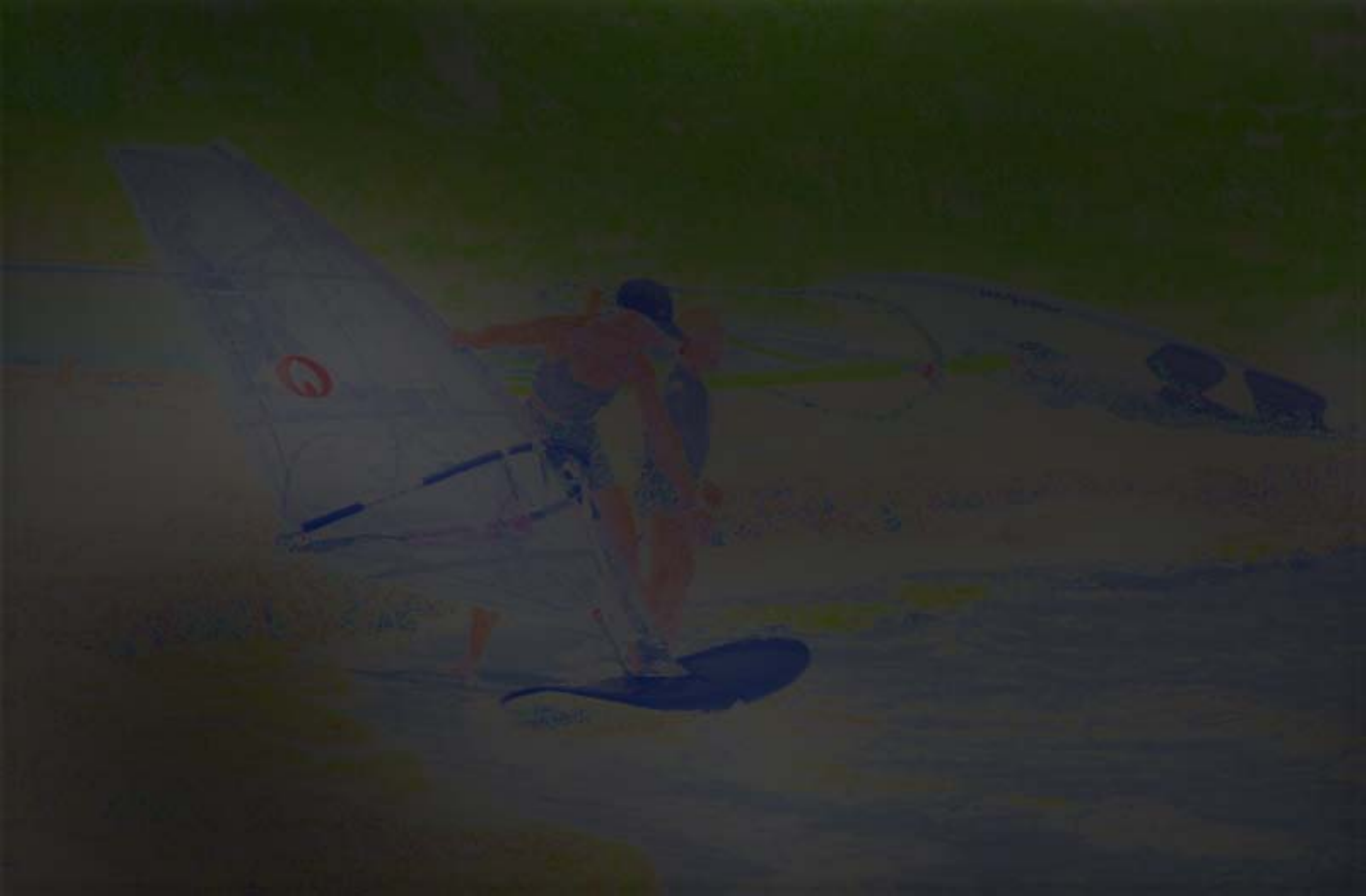}&
		\includegraphics[width=0.11\textwidth]{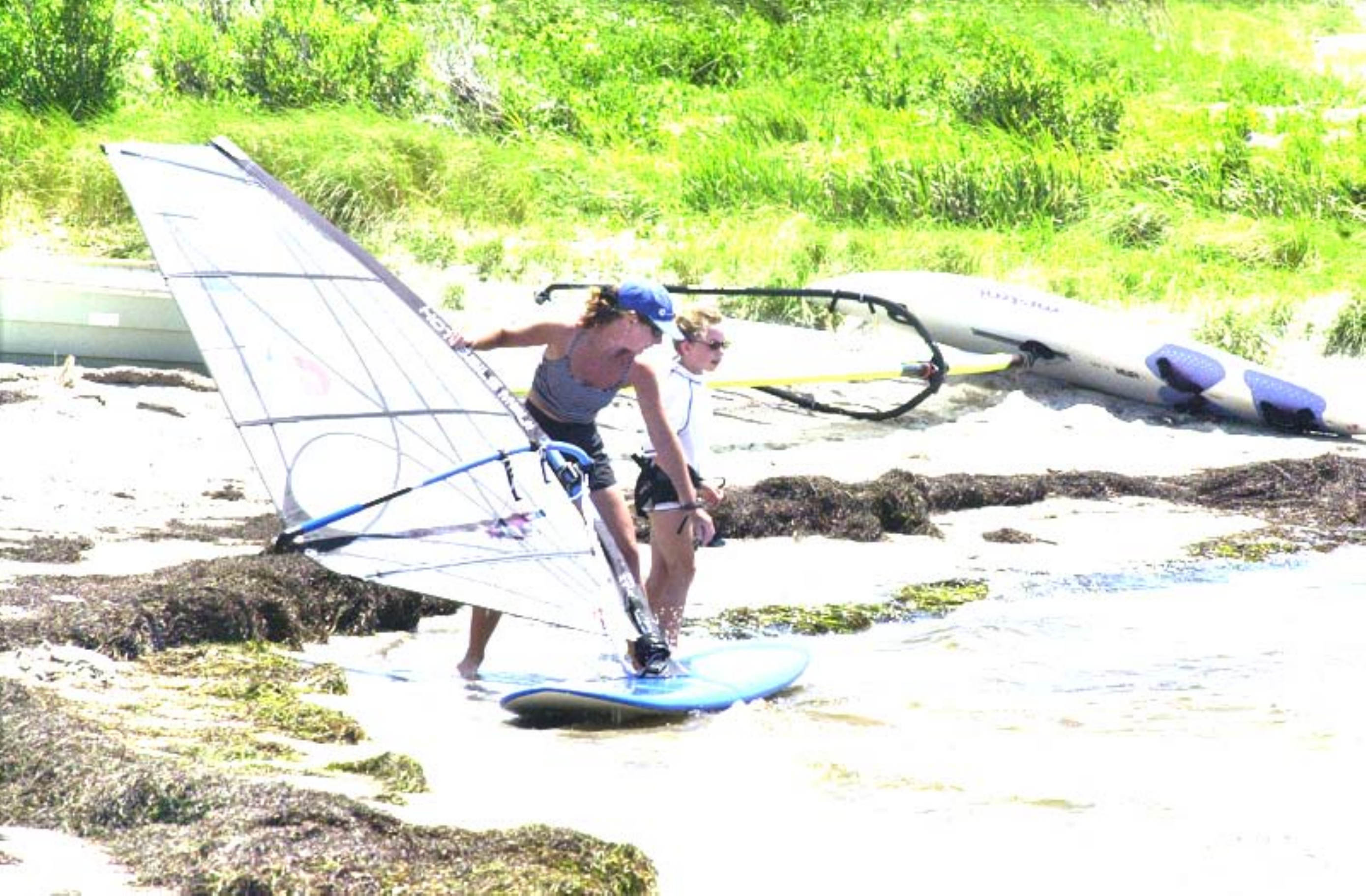}&
		\includegraphics[width=0.11\textwidth]{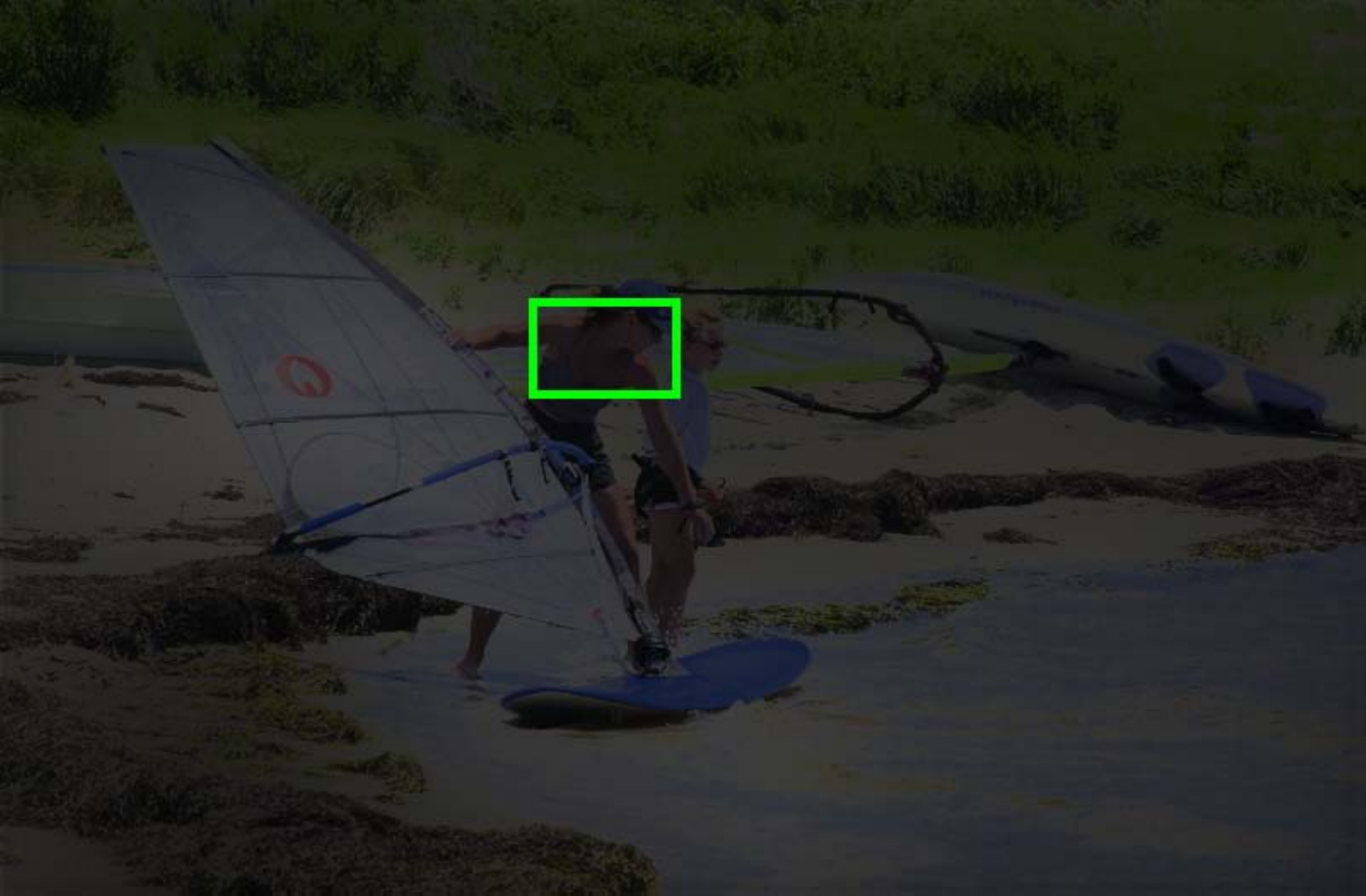}&
		\includegraphics[width=0.115\textwidth]{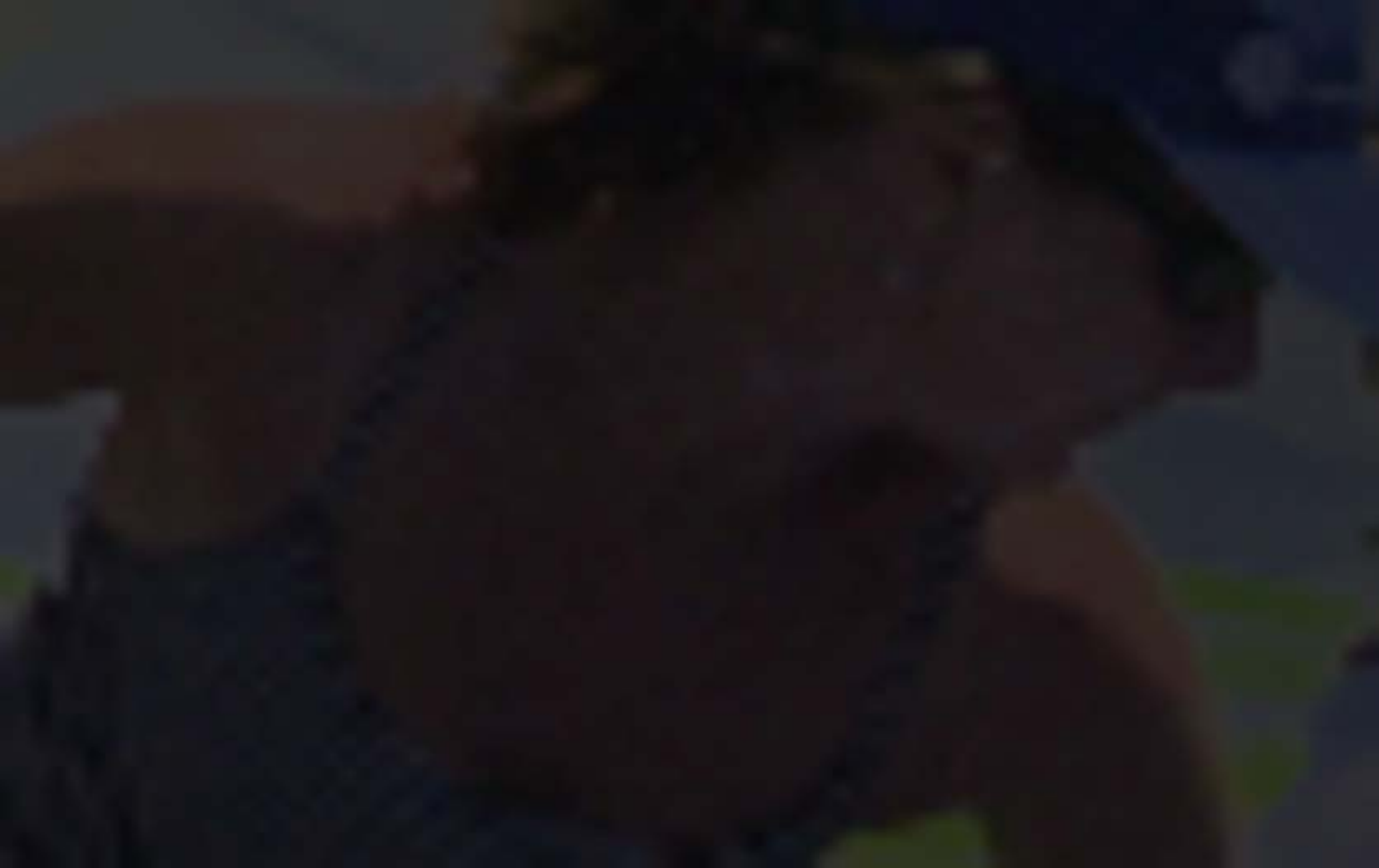}\\
		\multicolumn{4}{c}{(a)  ``(RP)"+``(EP)"+``(SS)"}\\
		\includegraphics[width=0.11\textwidth]{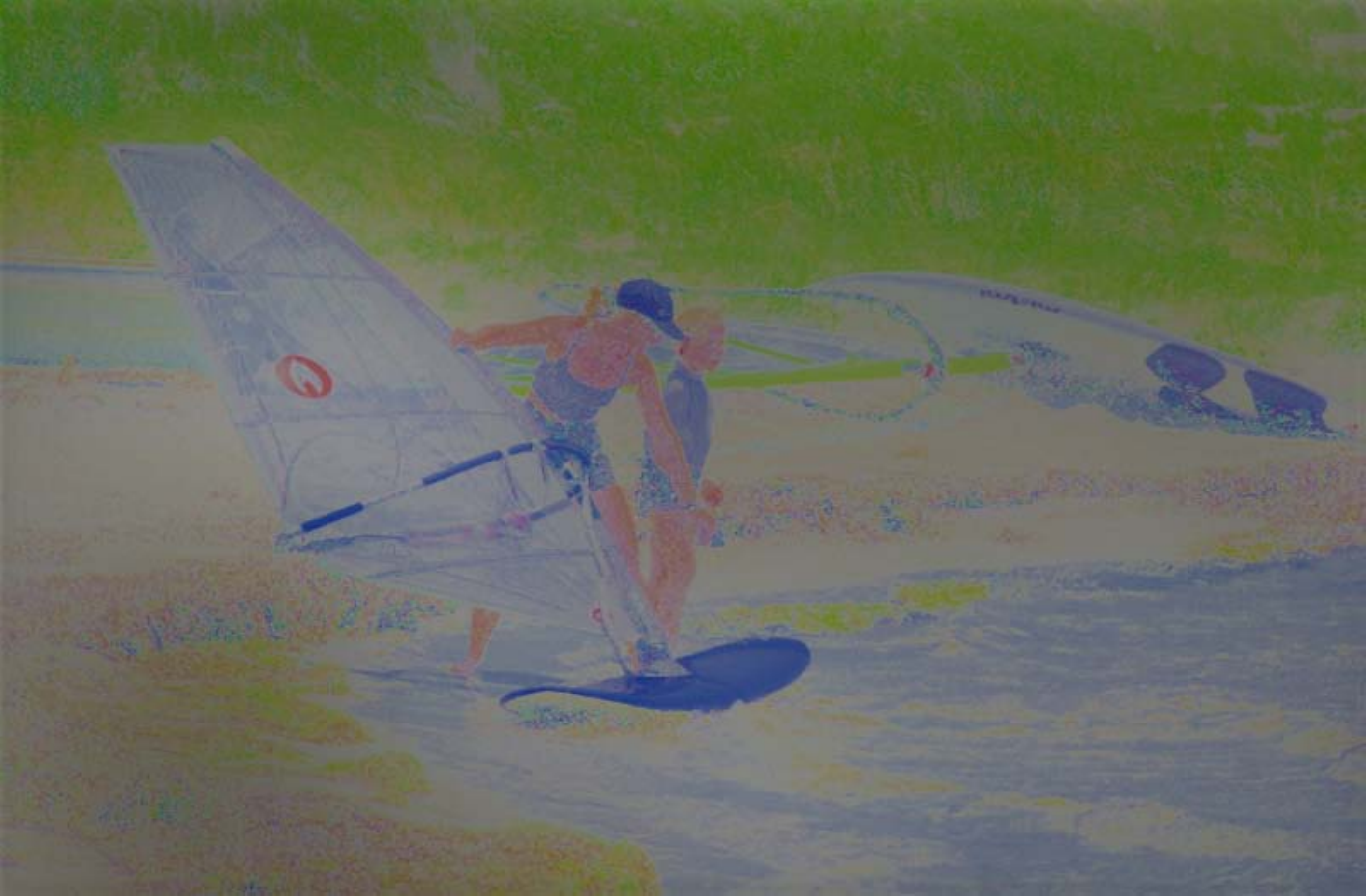}&
		\includegraphics[width=0.11\textwidth]{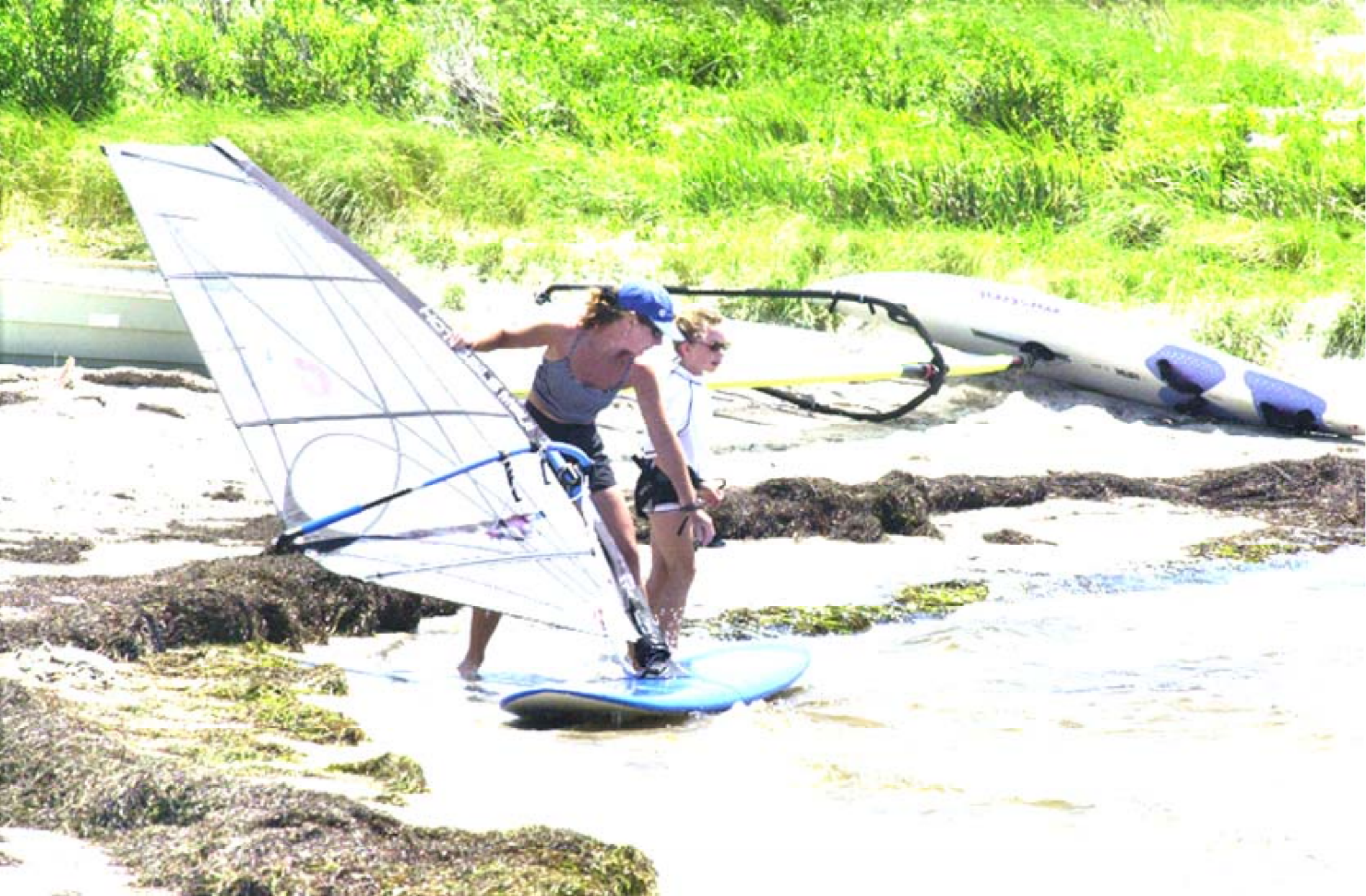}&
		\includegraphics[width=0.11\textwidth]{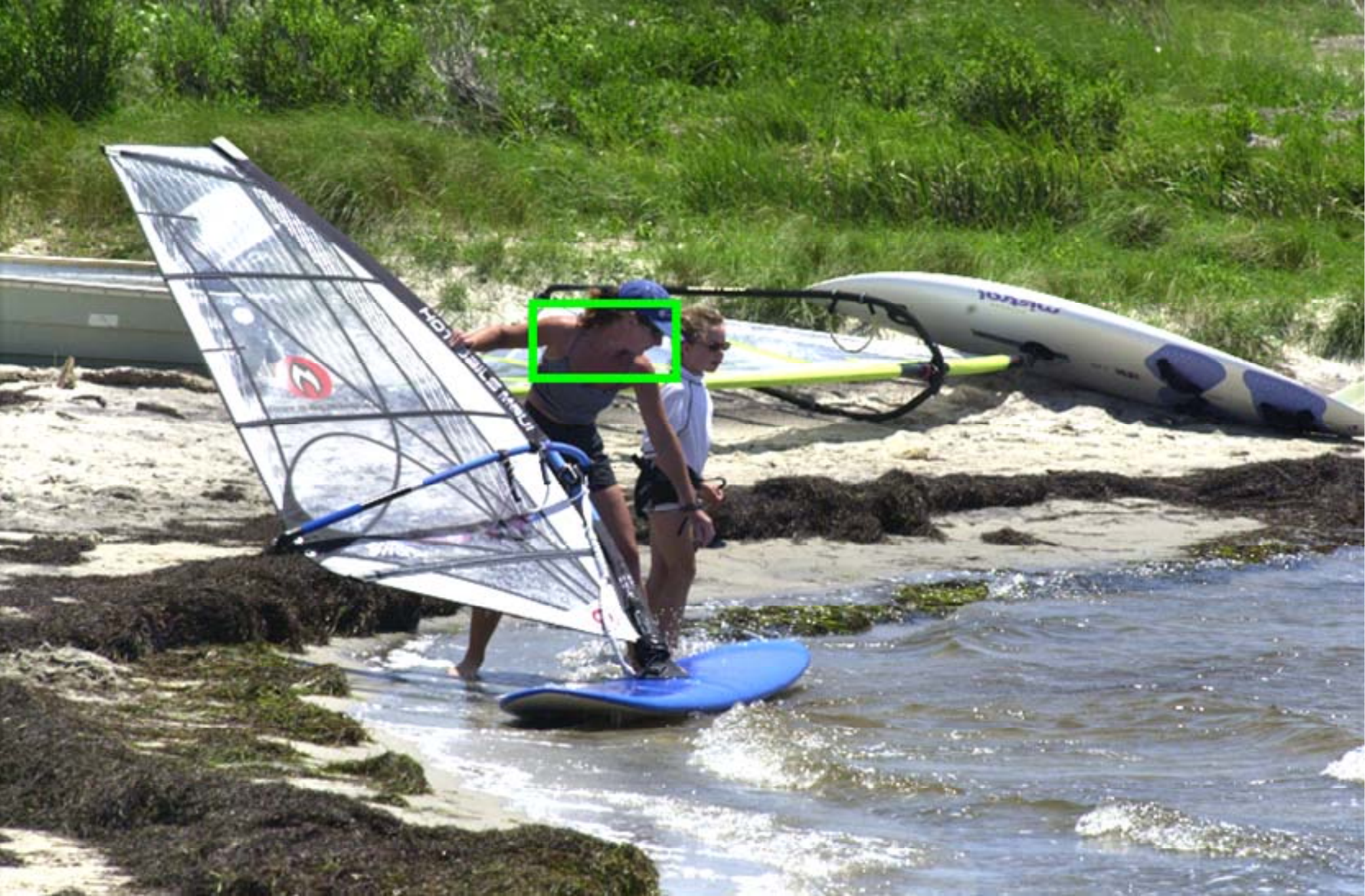}&
		\includegraphics[width=0.115\textwidth]{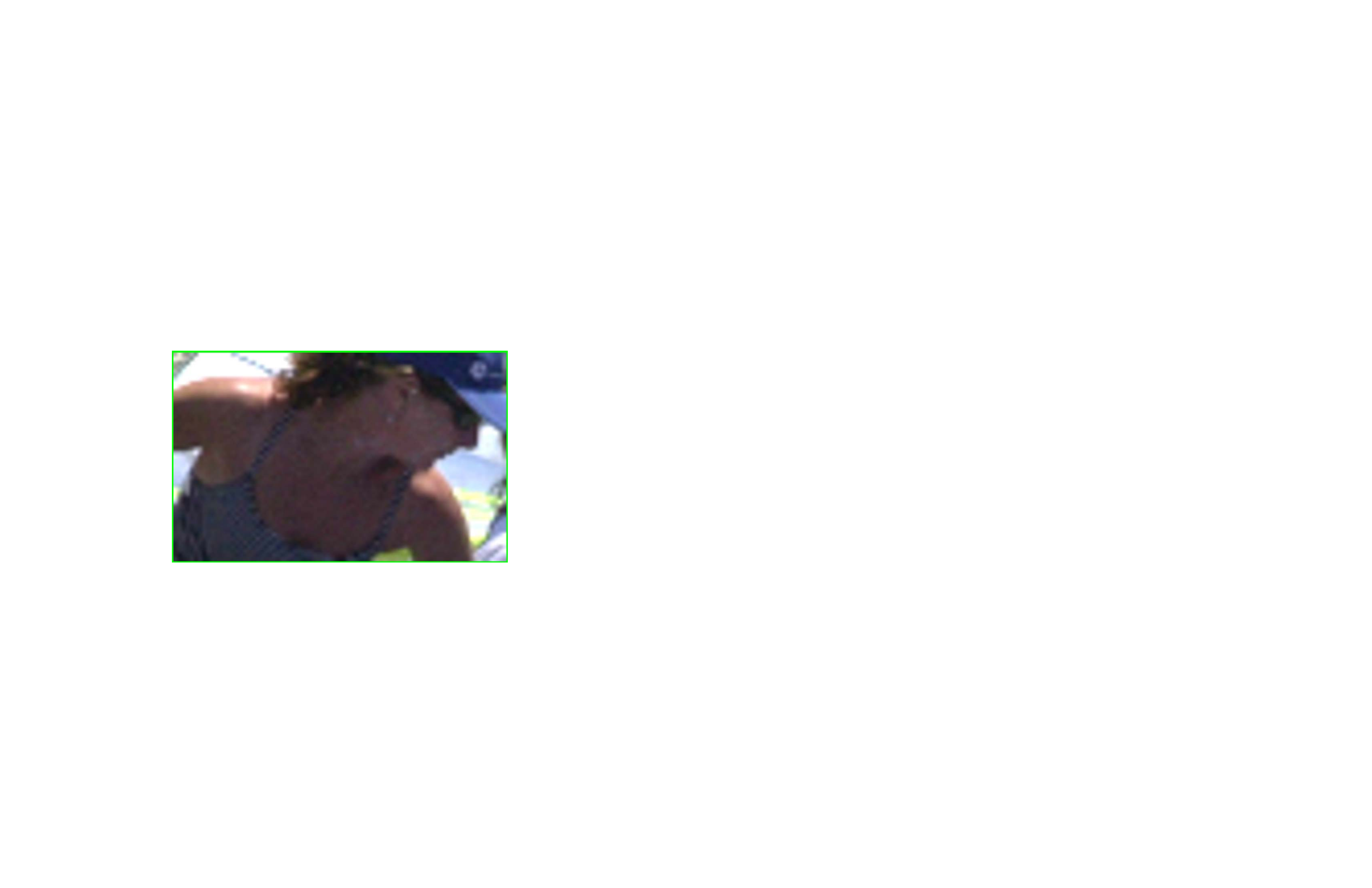}\\
		\multicolumn{4}{c}{(b) ``(RP)"+``(EP)"+``(SS)"+``(IA)"}\\
		\includegraphics[width=0.11\textwidth]{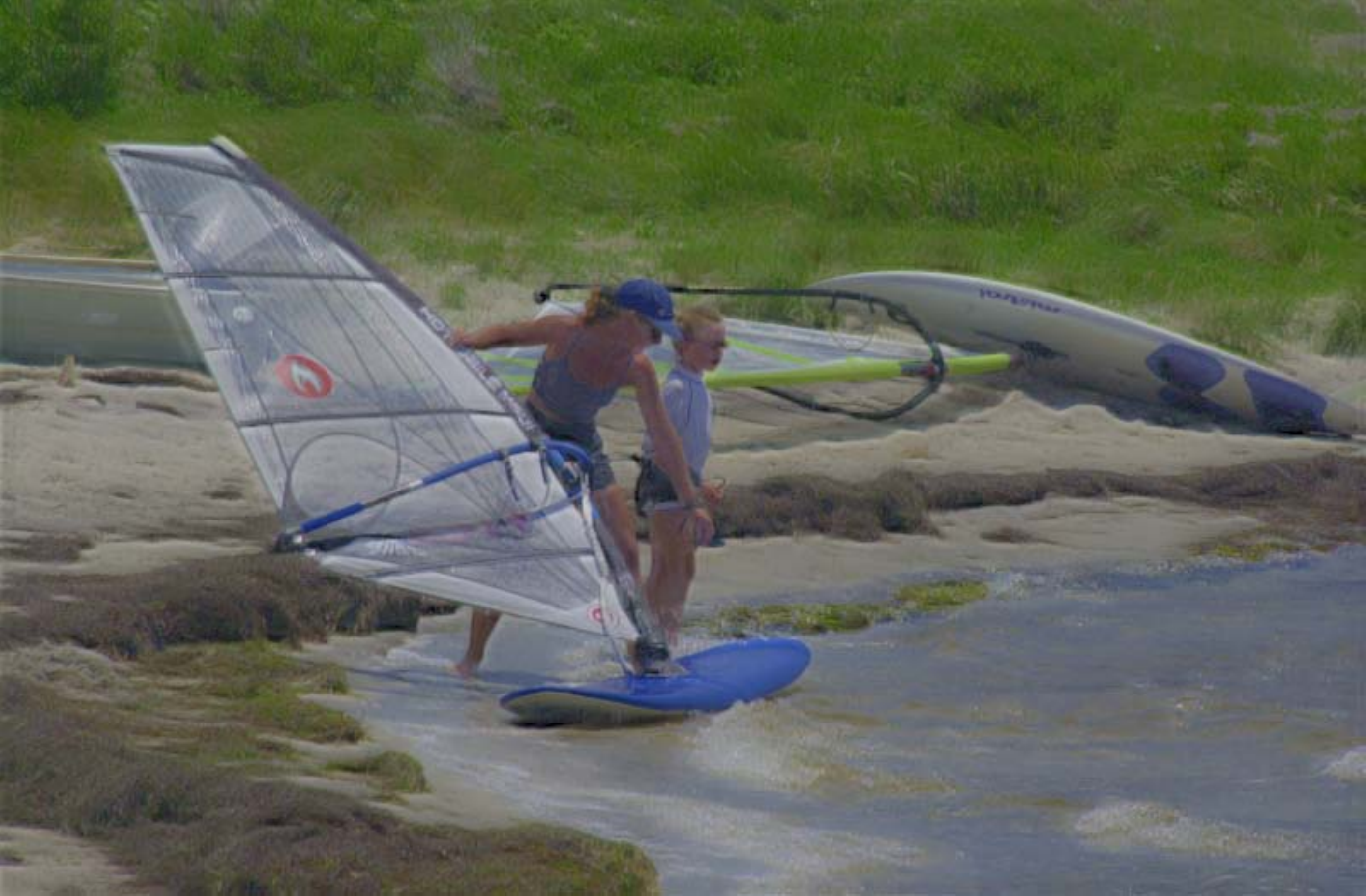}&
		\includegraphics[width=0.11\textwidth]{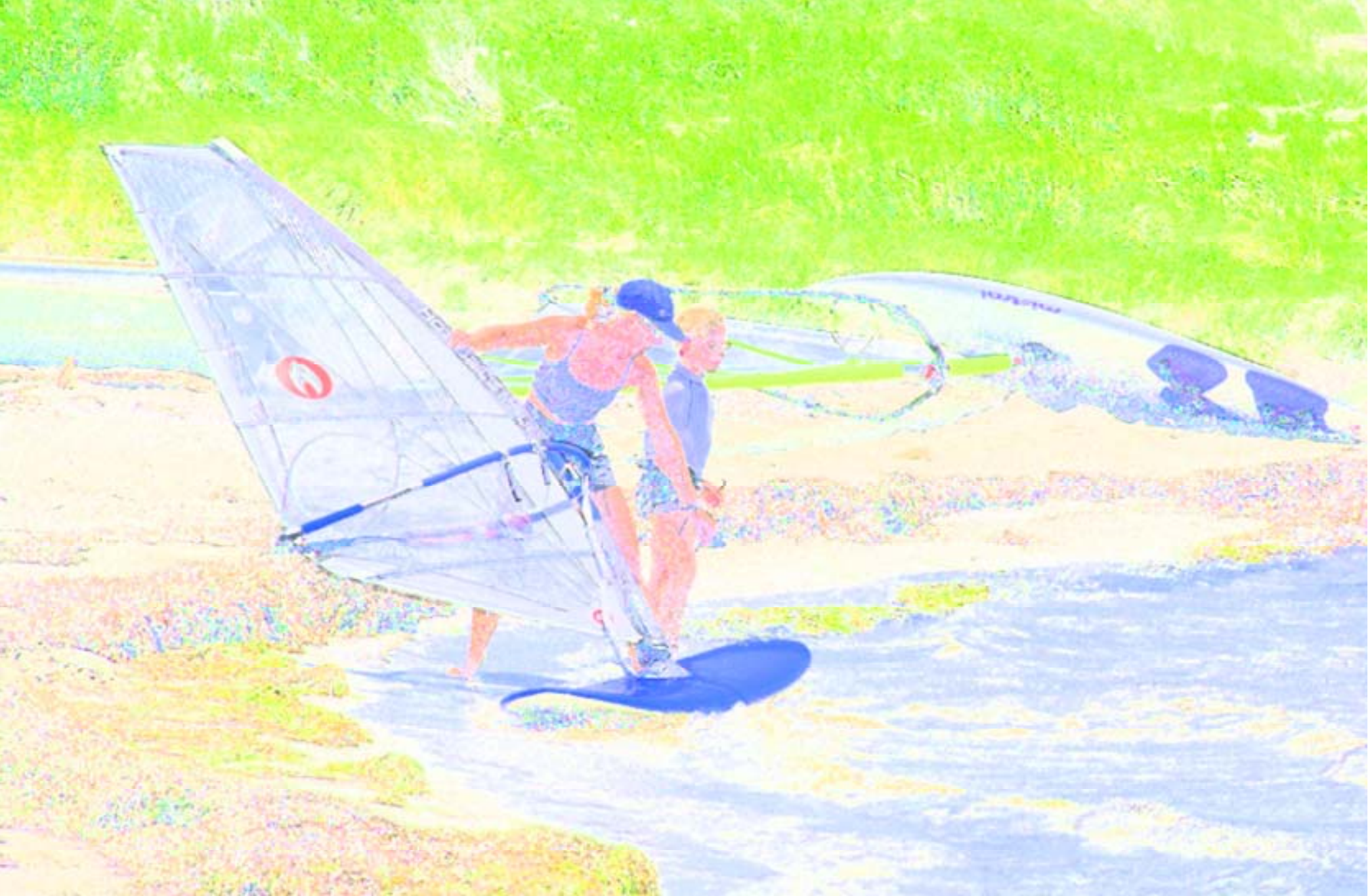}&
		\includegraphics[width=0.11\textwidth]{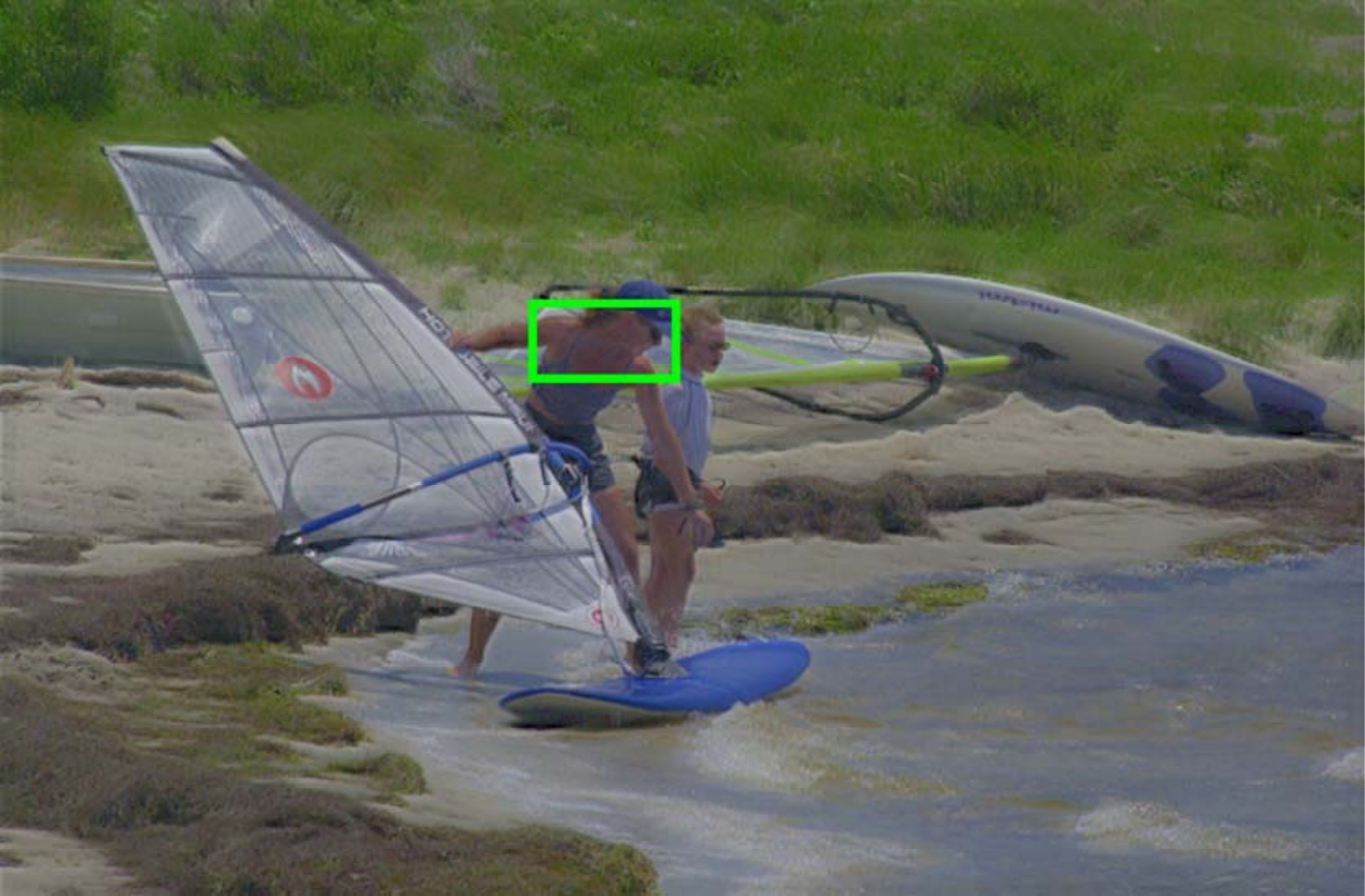}&
		\includegraphics[width=0.115\textwidth]{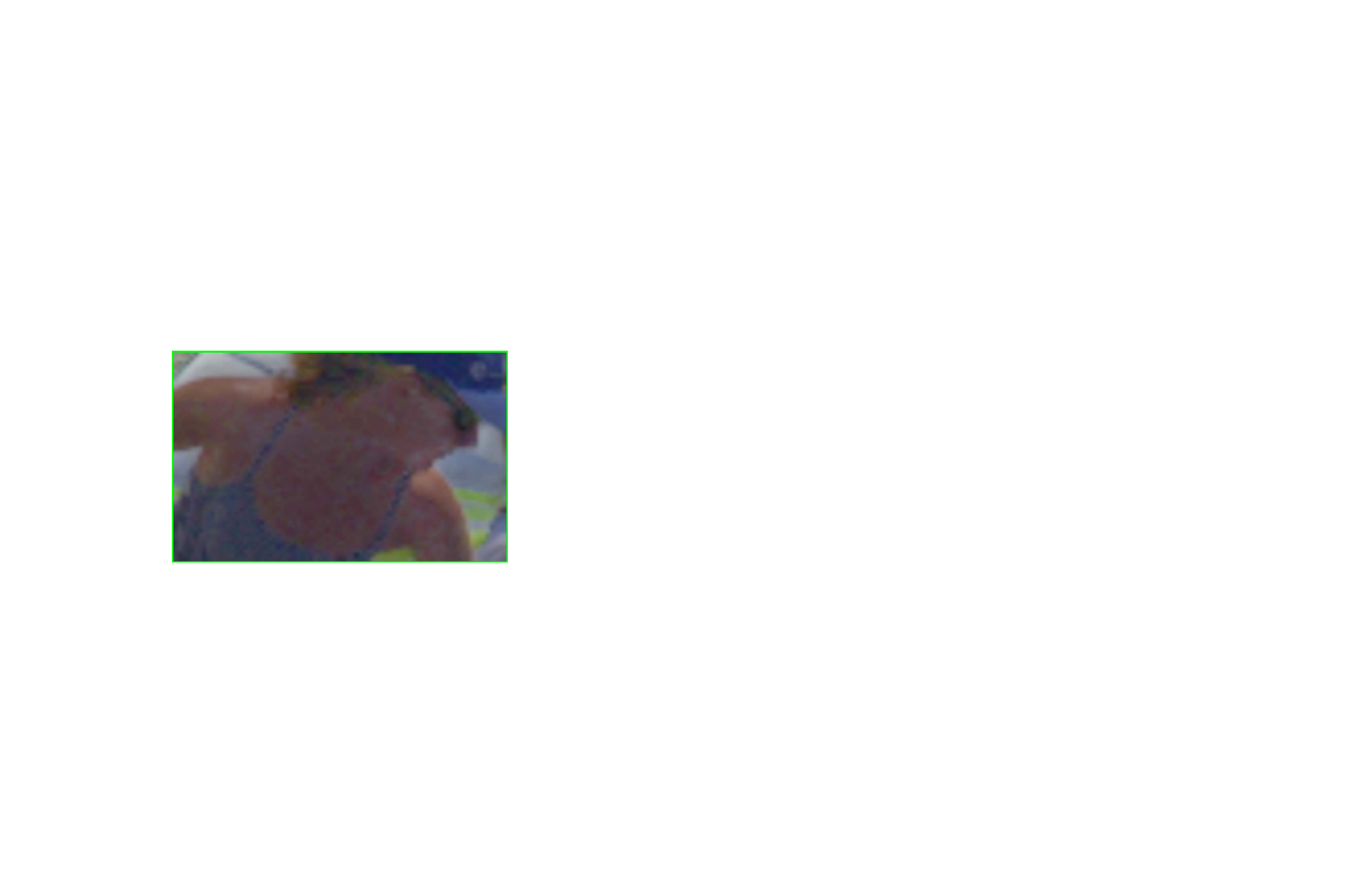}\\
		\multicolumn{4}{c}{(c)  ``(RP)"+``(LDD)"+``(IA)"}\\
		\includegraphics[width=0.11\textwidth]{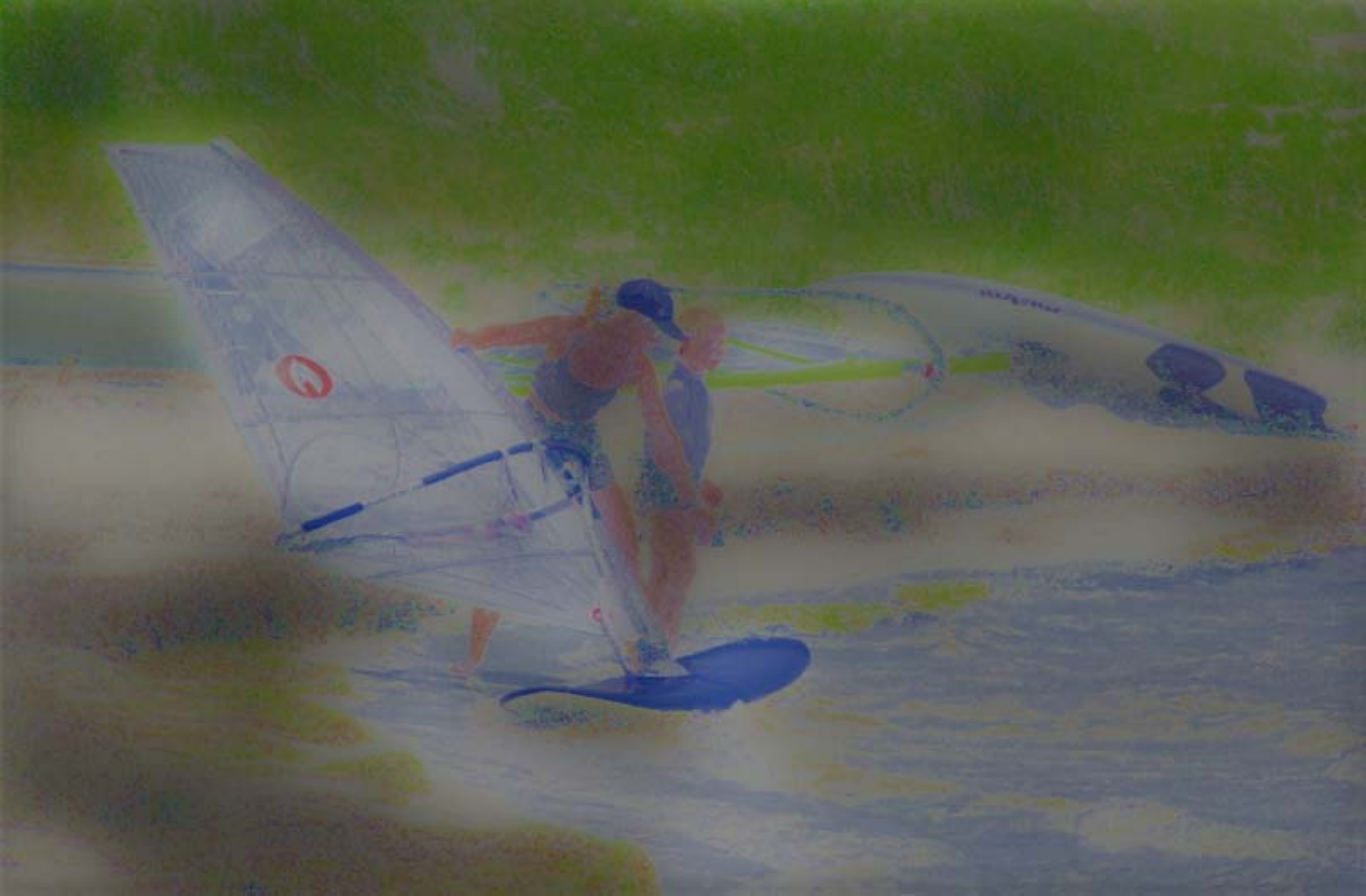}&
		\includegraphics[width=0.11\textwidth]{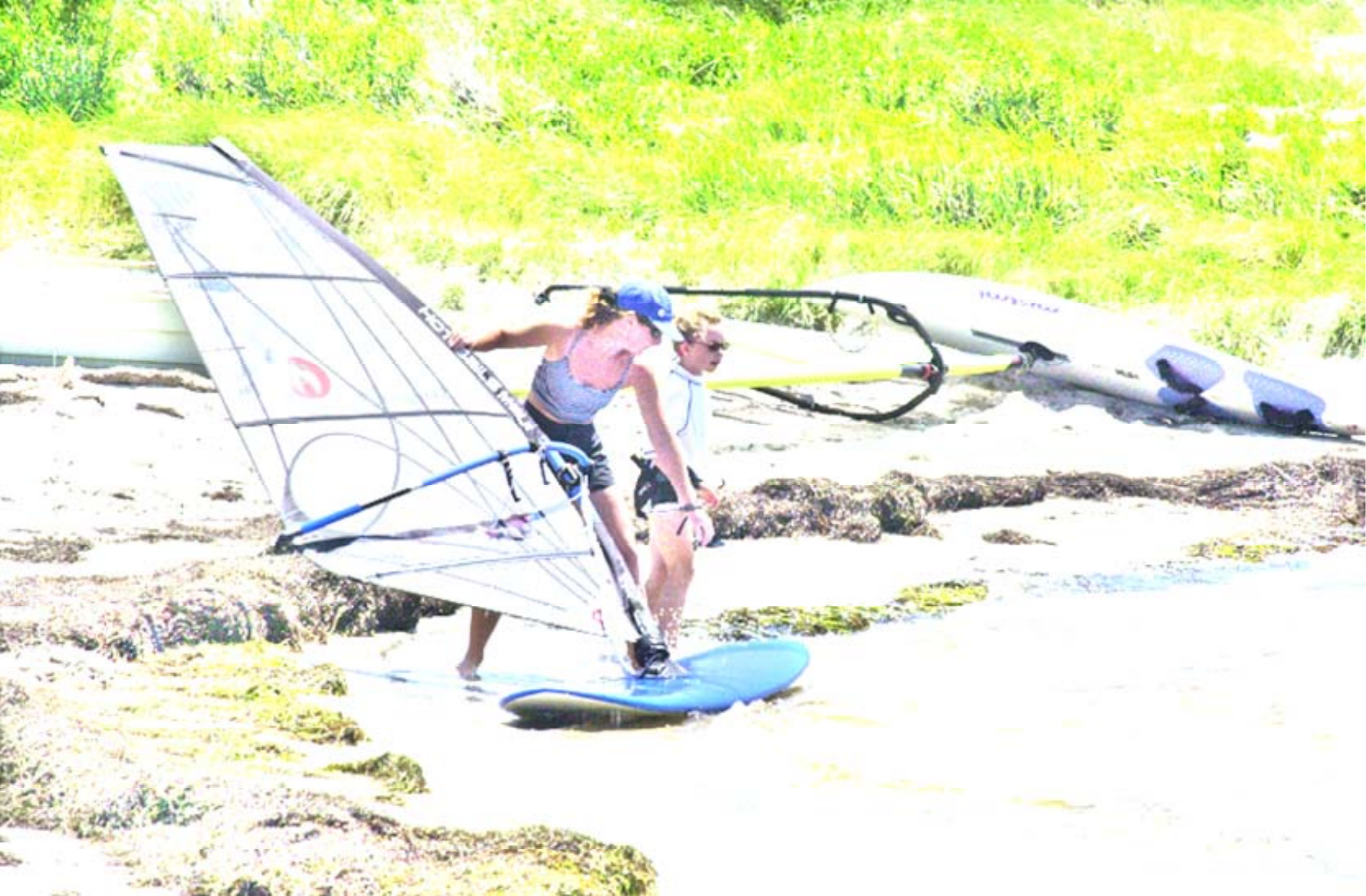}&
		\includegraphics[width=0.11\textwidth]{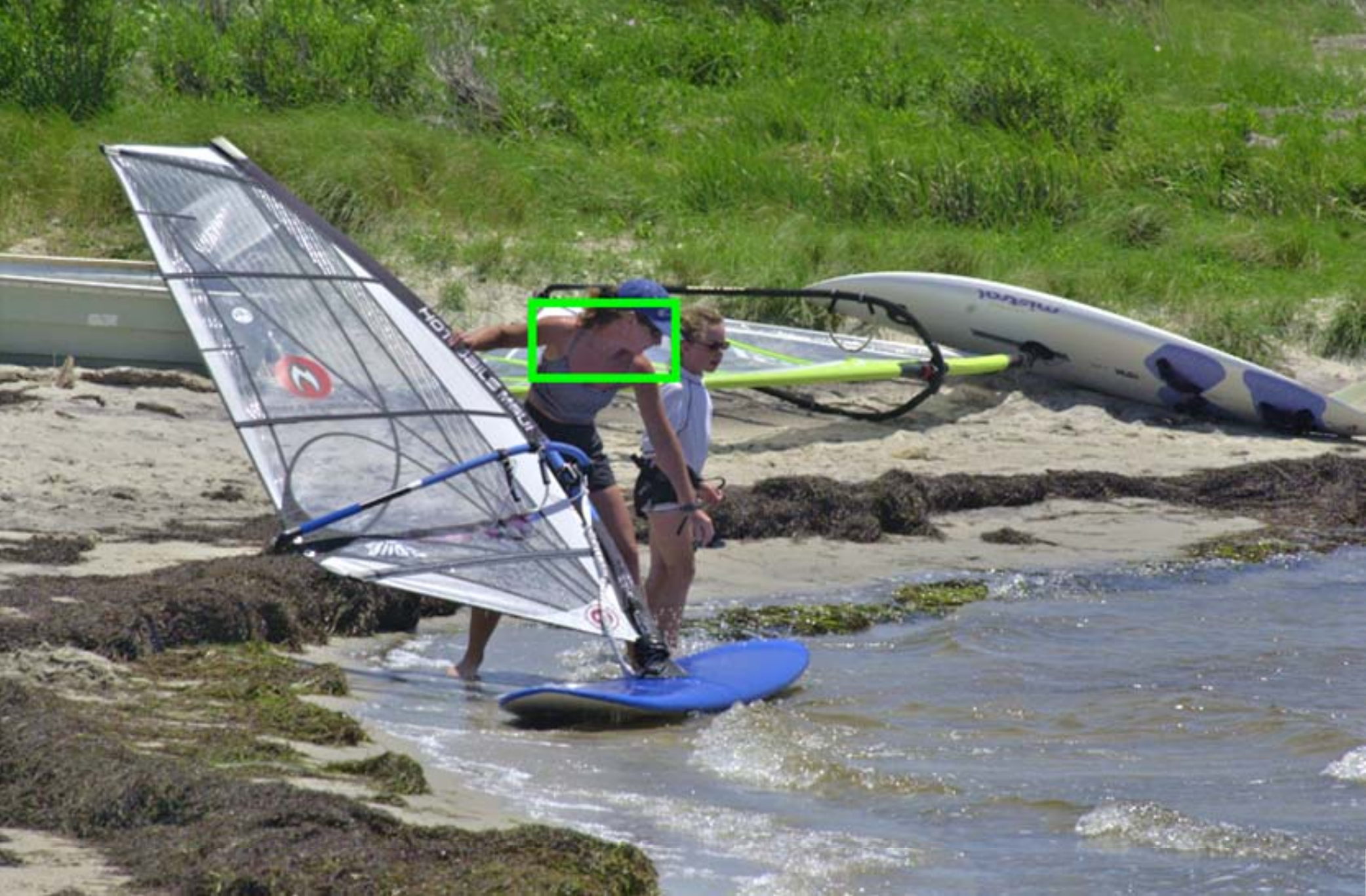}&
		\includegraphics[width=0.115\textwidth]{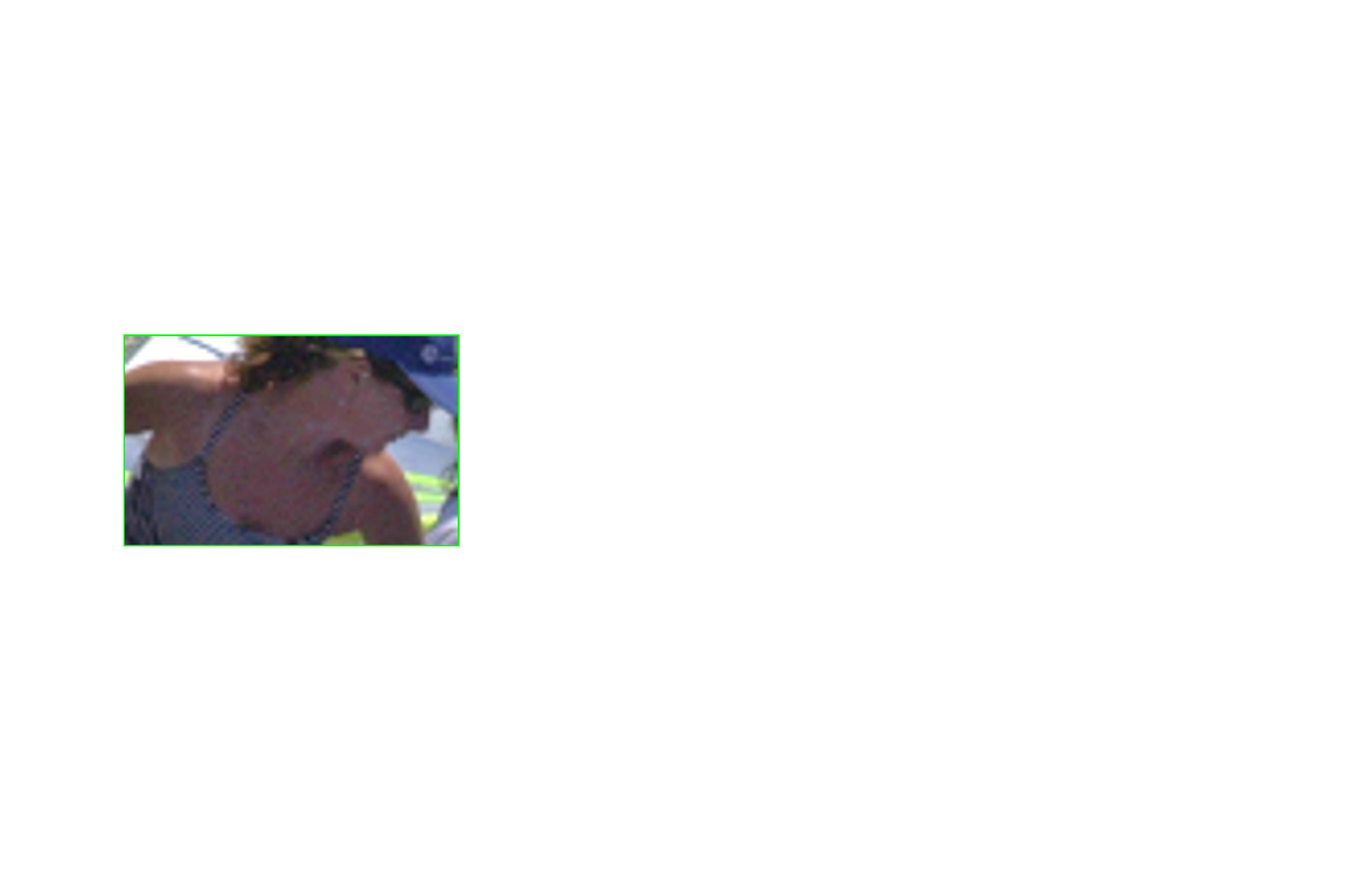}\\
		\multicolumn{4}{c}{(d) Ours ( ``(RP)"+``(EP)"+``(SS)"+``(LDD)"+``(IA)")}\\
	\end{tabular}
	\caption{Visual comparisons of our method with different prior strategies.  Notice that all the abbreviations of this figure come from the Fig.~\ref{fig:Flow}. }
	\label{fig:PriorsChange}
\end{figure}

To demonstrate the necessity of our propagation with hybrid priors navigation, we provide an illustrative comparison of different updating strategies for the decomposition and the corresponding enhancement performance in Fig.~\ref{fig:PriorsChange}. Concretely, we consider three different variations of priors, i.e. only data-dependent (i.e., ``(LDD)"), only knowledge-based (i.e., ``(EP)"+``(SS)") and our proposed hybrid prior (i.e., ``(LDD)"+``(EP)"+``(SS)"). Moreover, we also consider a condition of only knowledge-based prior without illumination adjustment (i.e., ``(IA)"). It evidents that the illumination adjustment is essential to adjust the luminance of illumination as the subfigures (a) and (b) of Fig.~\ref{fig:PriorsChange} show. Obviously, some details are missing in the reflectance generated by the knowledge-based prior, thus, there exist some details cannot be recovered in the enhanced results. We observe that the reflectance estimated by data-dependent prior is over-smooth, which leads to the absence of some structural information in the enhanced result. In contrast, the reflectance obtained by the proposed hybrid prior is felicitous and the enhanced result has the distinguished enhancement effects (see zoomed-in regions). 

\subsection{Learnable Architecture} 
As for the learnable architecture, we would like to point out that we actually produce a learnable descent direction derived from the data distributions (see ''(LDD)'' in Fig.~\ref{fig:Flow}), to assist searching our desired solution. Additionally, since the knowledge-based submodule in our hybrid priors has the ability to roughly estimate the latent image structures, the main left task to the data-dependent submodule should be refining the rich details and removing small corrections. Therefore, it is essential to adopt a denoising-type strategy for our learnable architecture. That is, we generate the training image pairs by adding different levels of Gaussian noises to simulate the corruptions and consider the clear images as the outputs of the architecture.

Specifically, a simple CNN architecture is adopted as our learnable architecture, which consists of 7 dilated convolution layers with 64 kernels, acting on a kernel of size 3. We set a ReLU as nonlinear activation function in between two convolution layers, batch normalizations are also introduced for convolution operations from 2nd to 6th linear layers. We adopt mean square error as our training loss.
As for the training data, we randomly select 800 images from ImageNet database~\cite{krizhevsky2012imagenet}. We crop them into small patches of size 35$\times$35, and also augment our training set by rotation and flip.

\begin{figure}[t]
	\centering
	\begin{tabular}{c@{\extracolsep{0.3em}}c@{\extracolsep{0.3em}}c@{\extracolsep{0.3em}}c}
		\includegraphics[width=.23\linewidth]{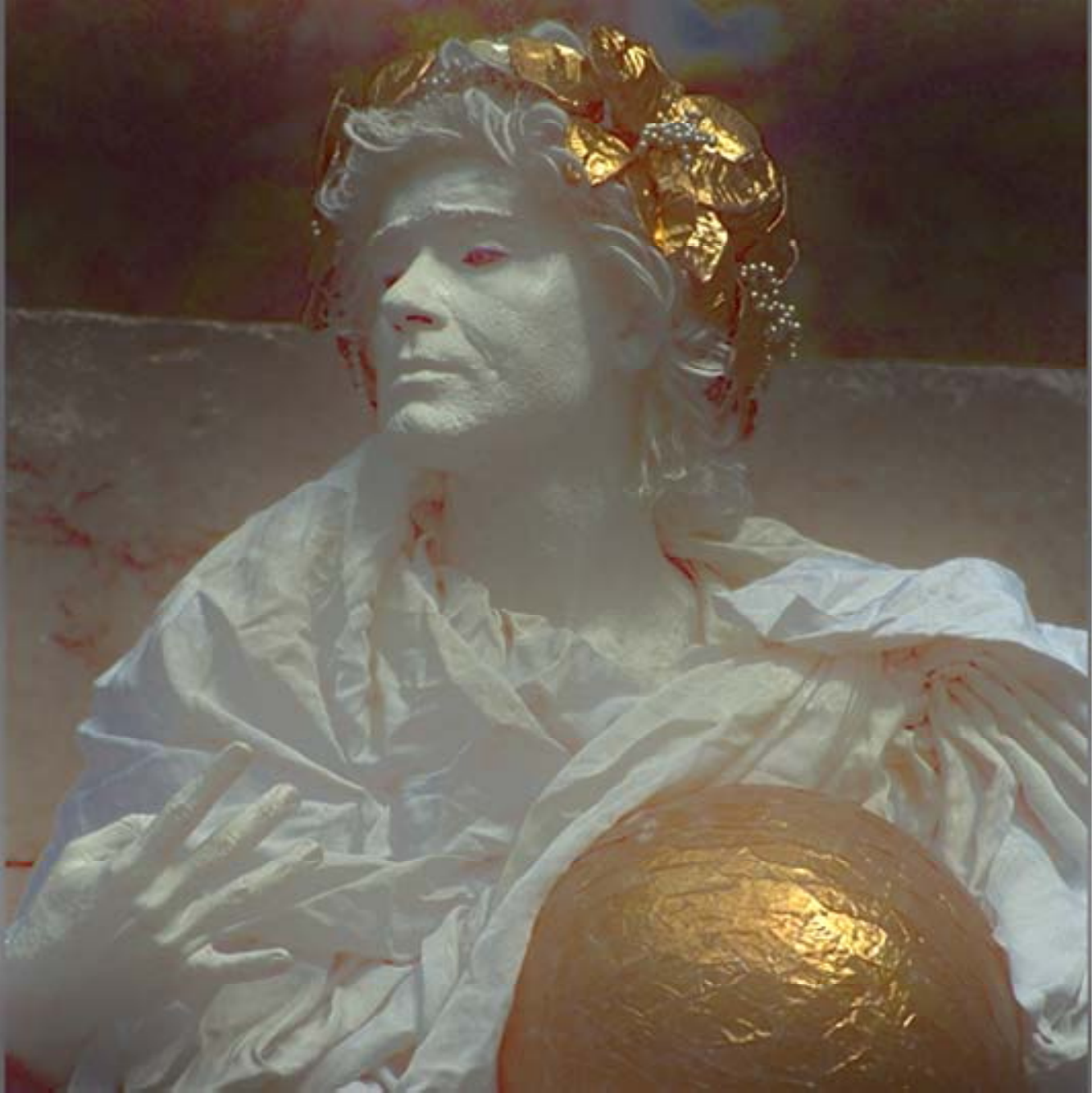}&
		\includegraphics[width=.23\linewidth]{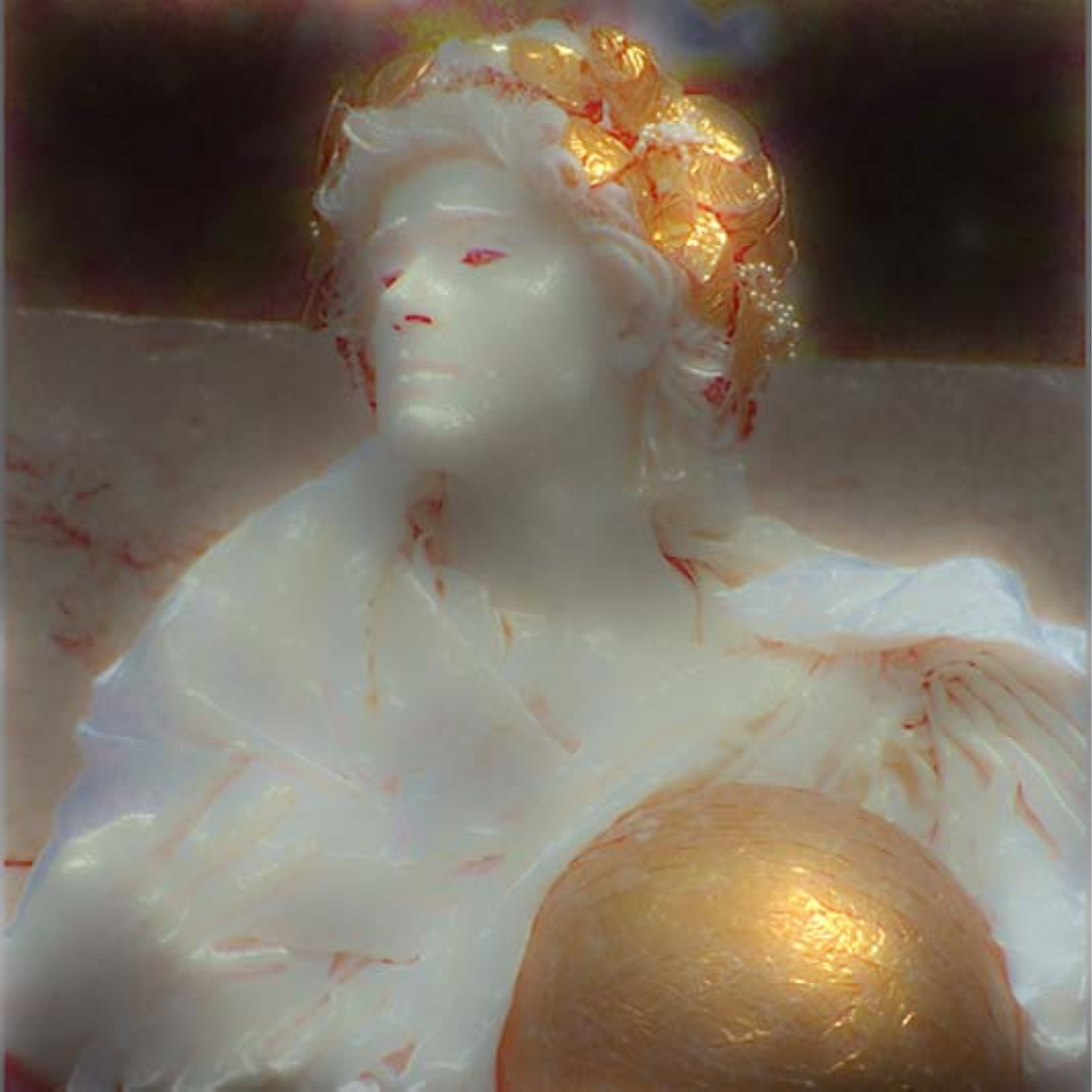}&
		\includegraphics[width=.23\linewidth]{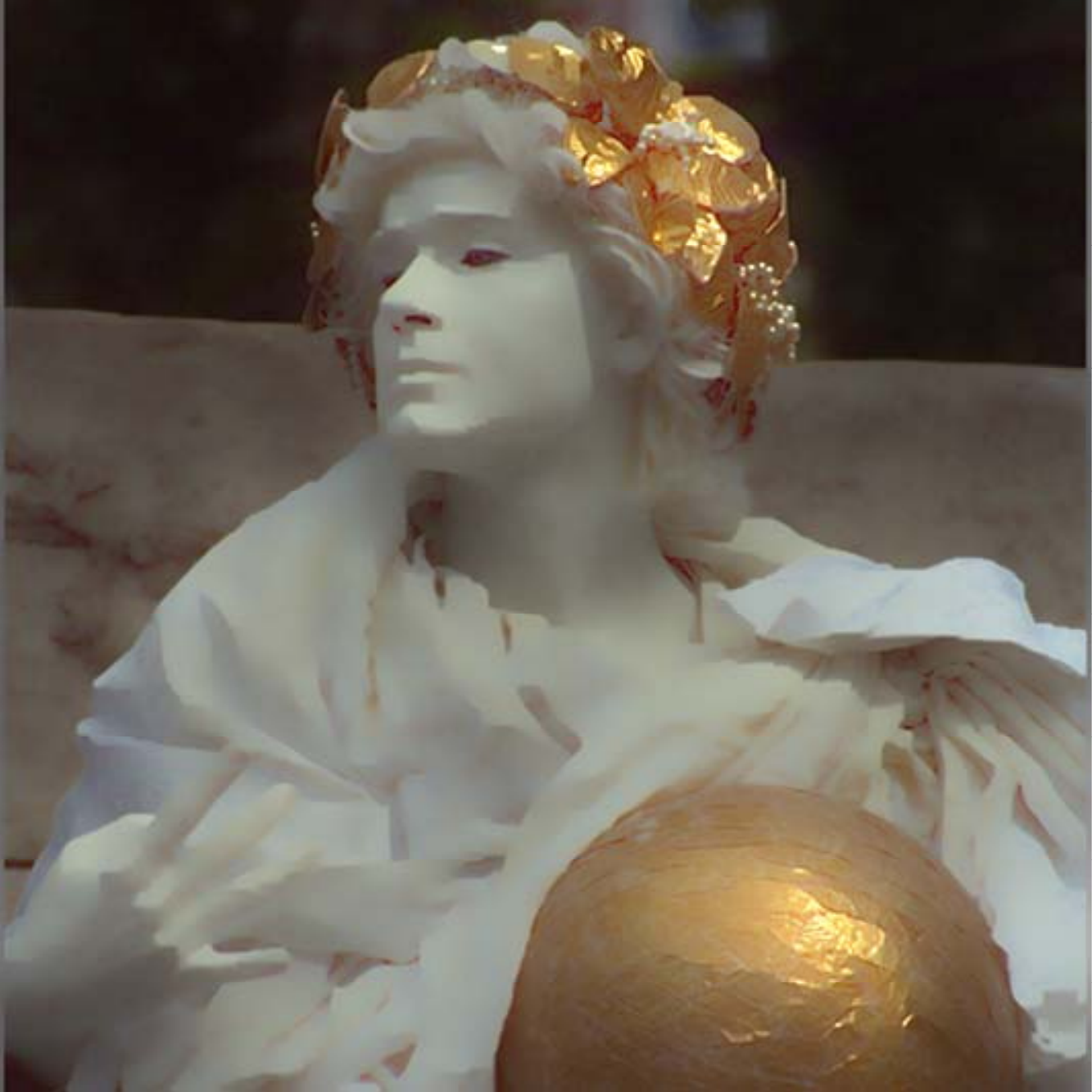}&
		\includegraphics[width=.23\linewidth]{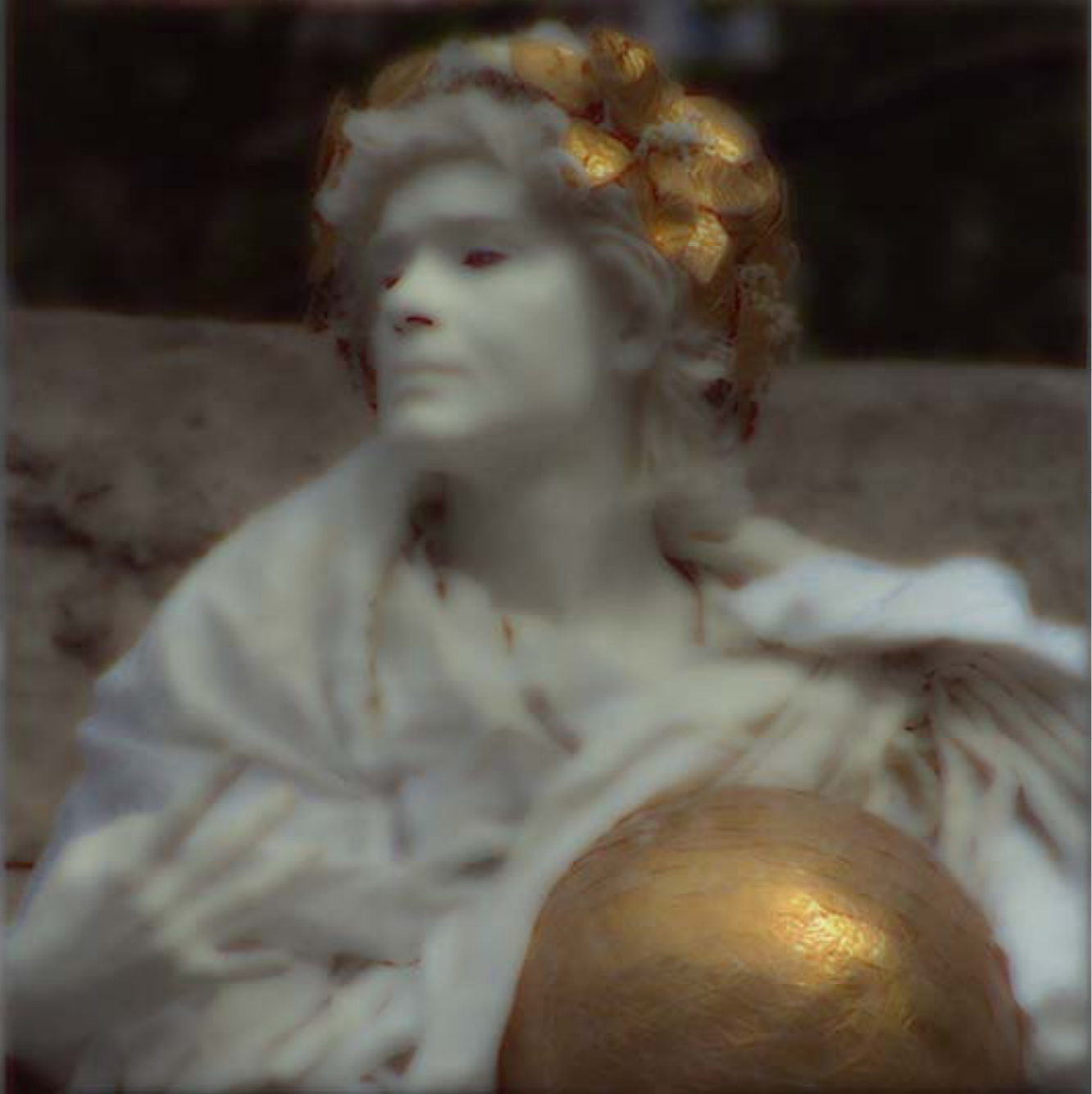}\\
		\includegraphics[width=.23\linewidth]{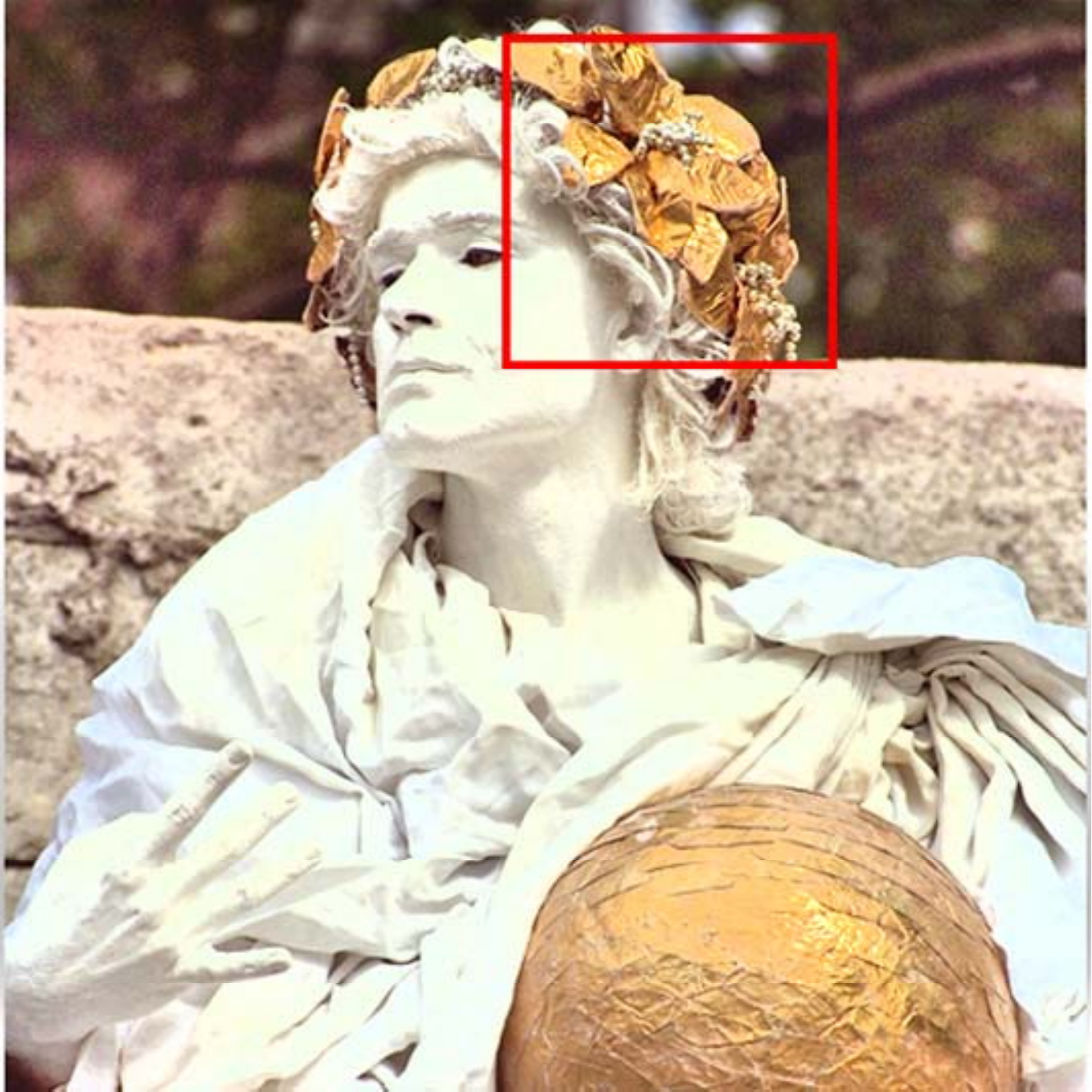}&
		\includegraphics[width=.23\linewidth]{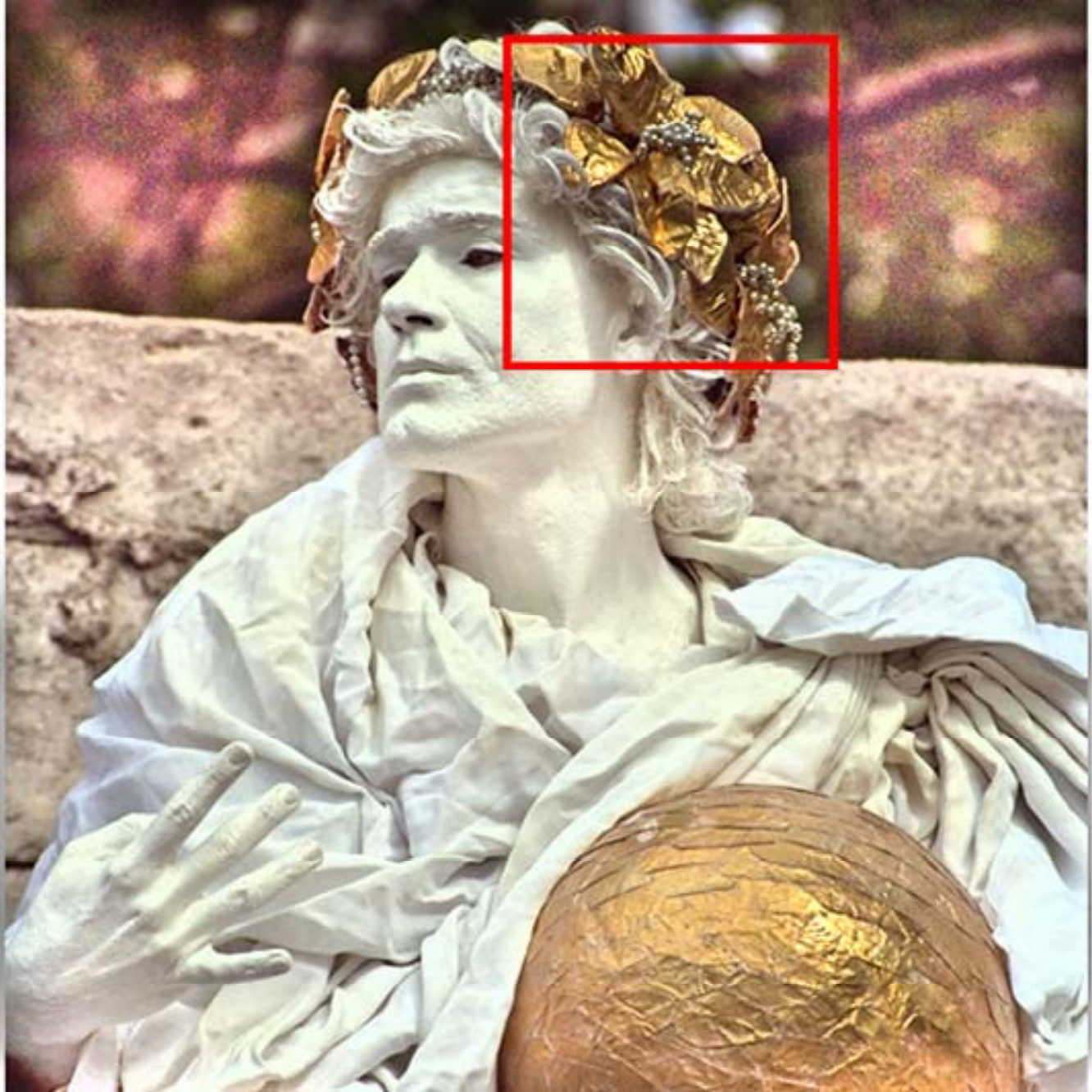}&
		\includegraphics[width=.23\linewidth]{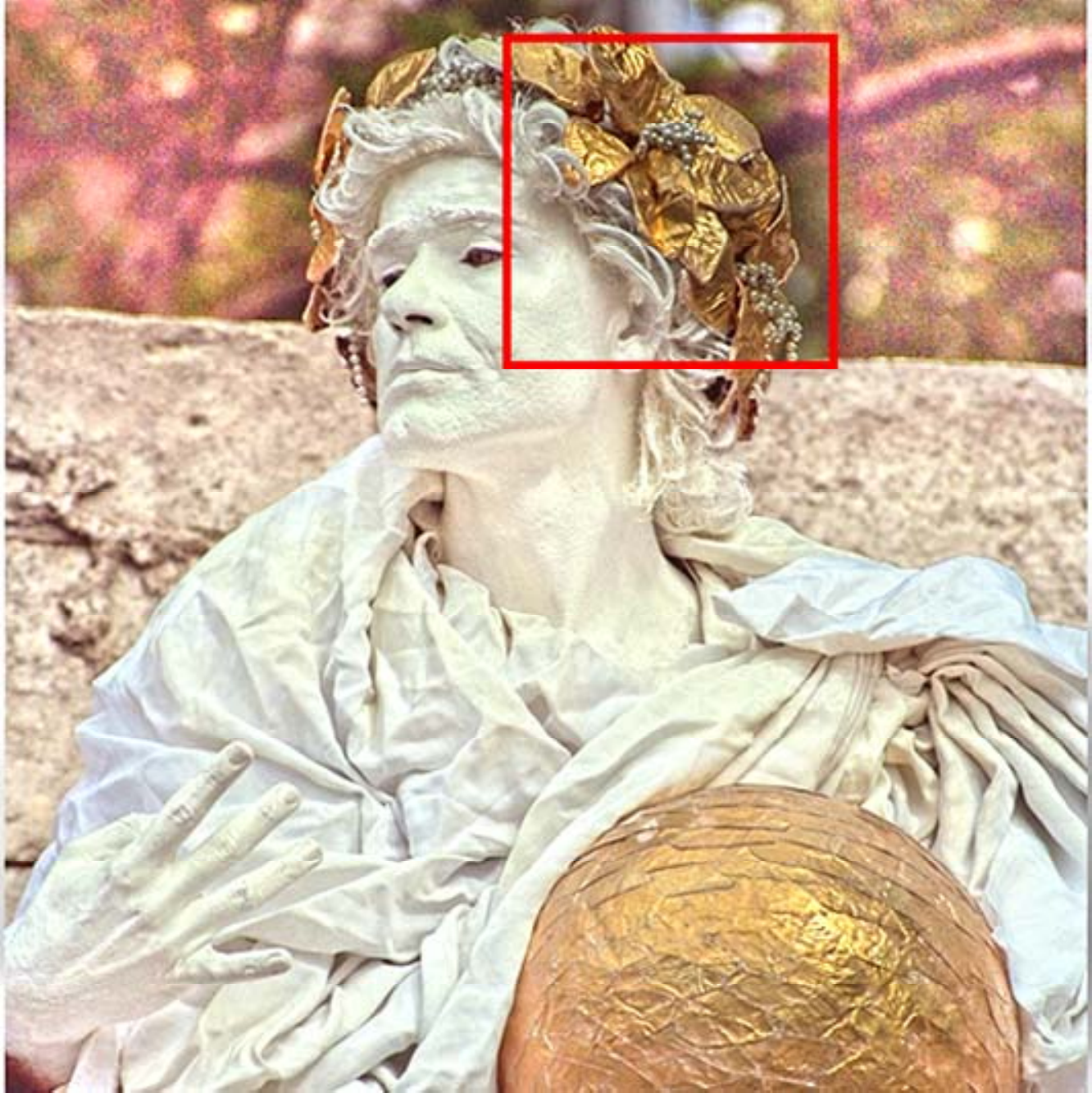}&
		\includegraphics[width=.23\linewidth]{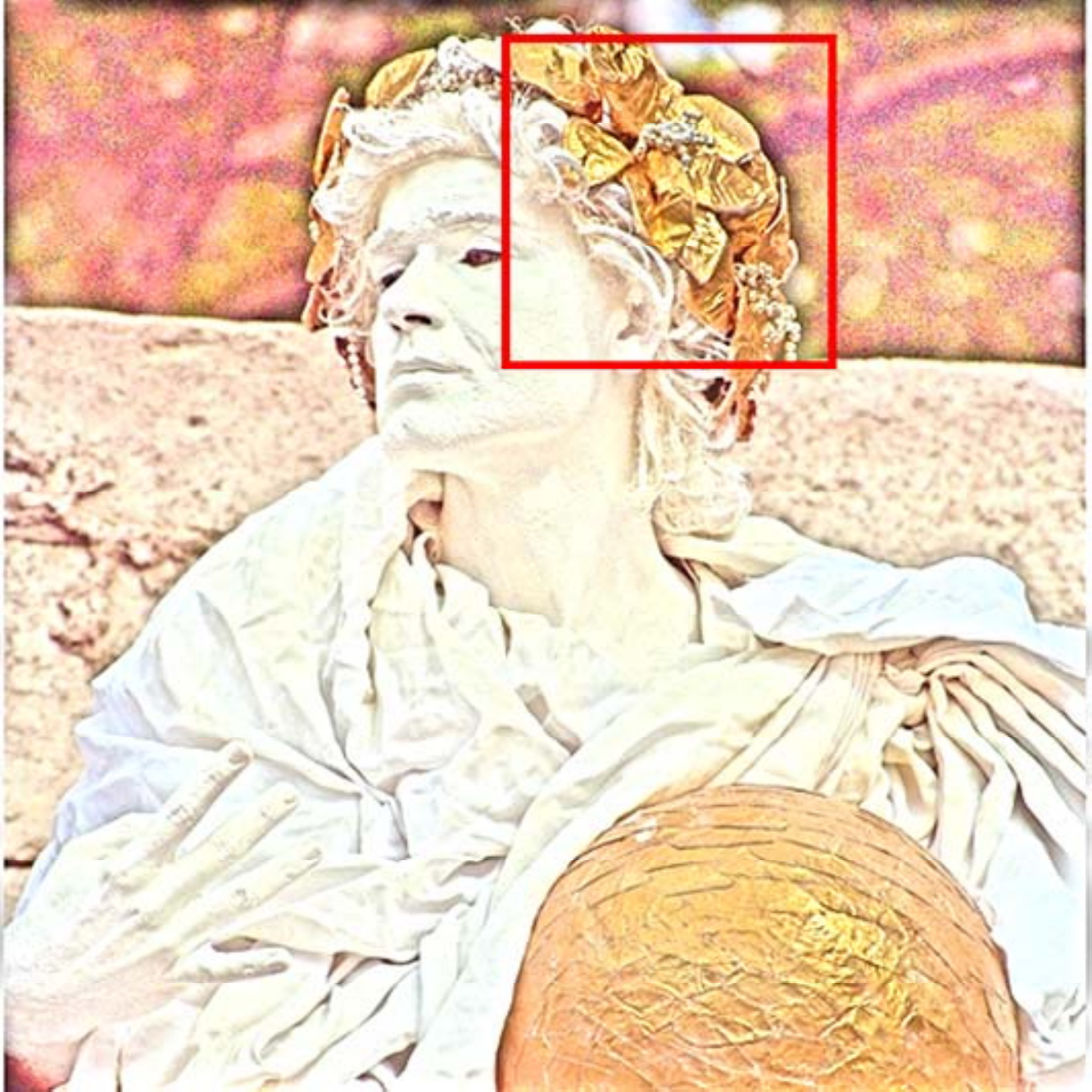}\\
		\includegraphics[width=.23\linewidth]{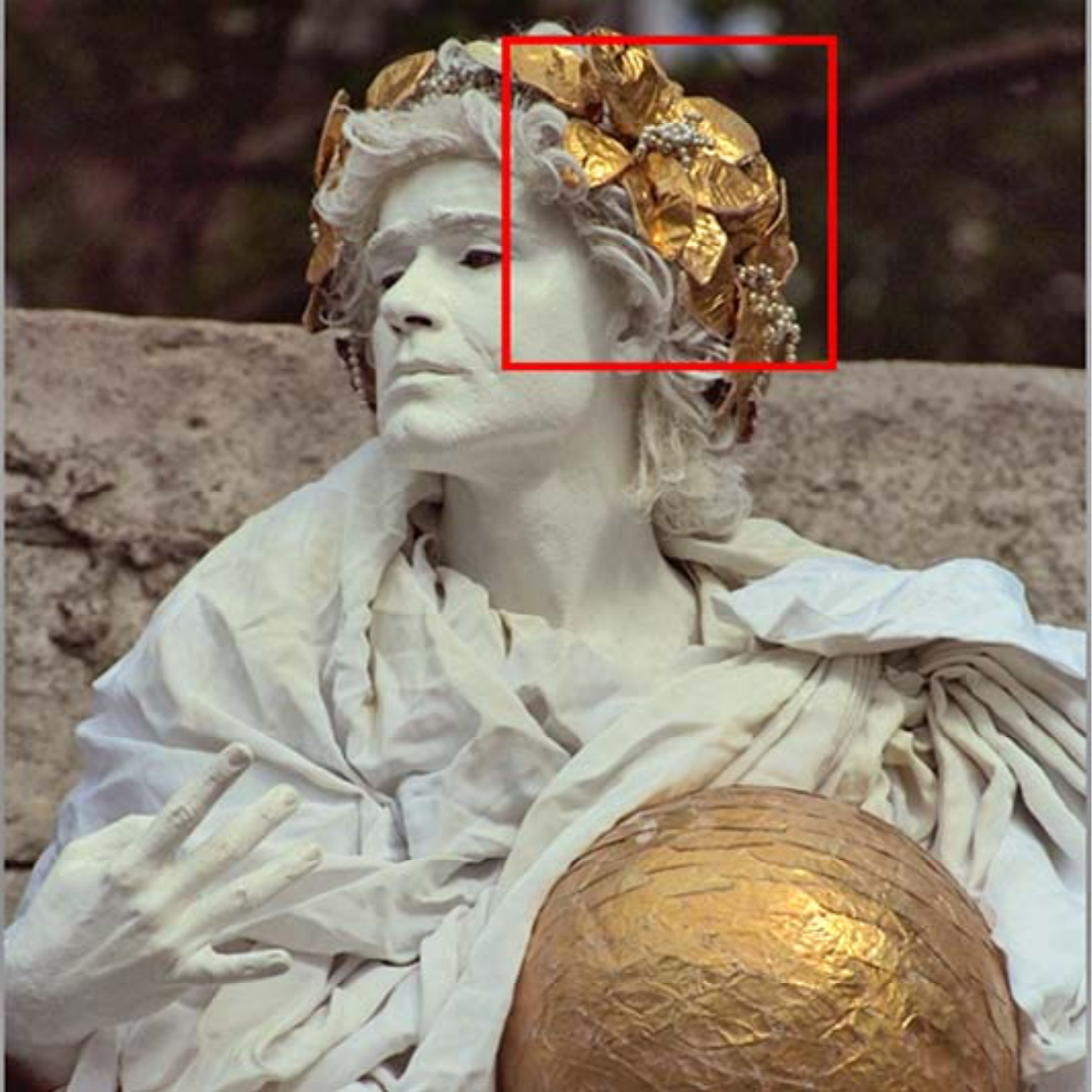}&
		\includegraphics[width=.23\linewidth]{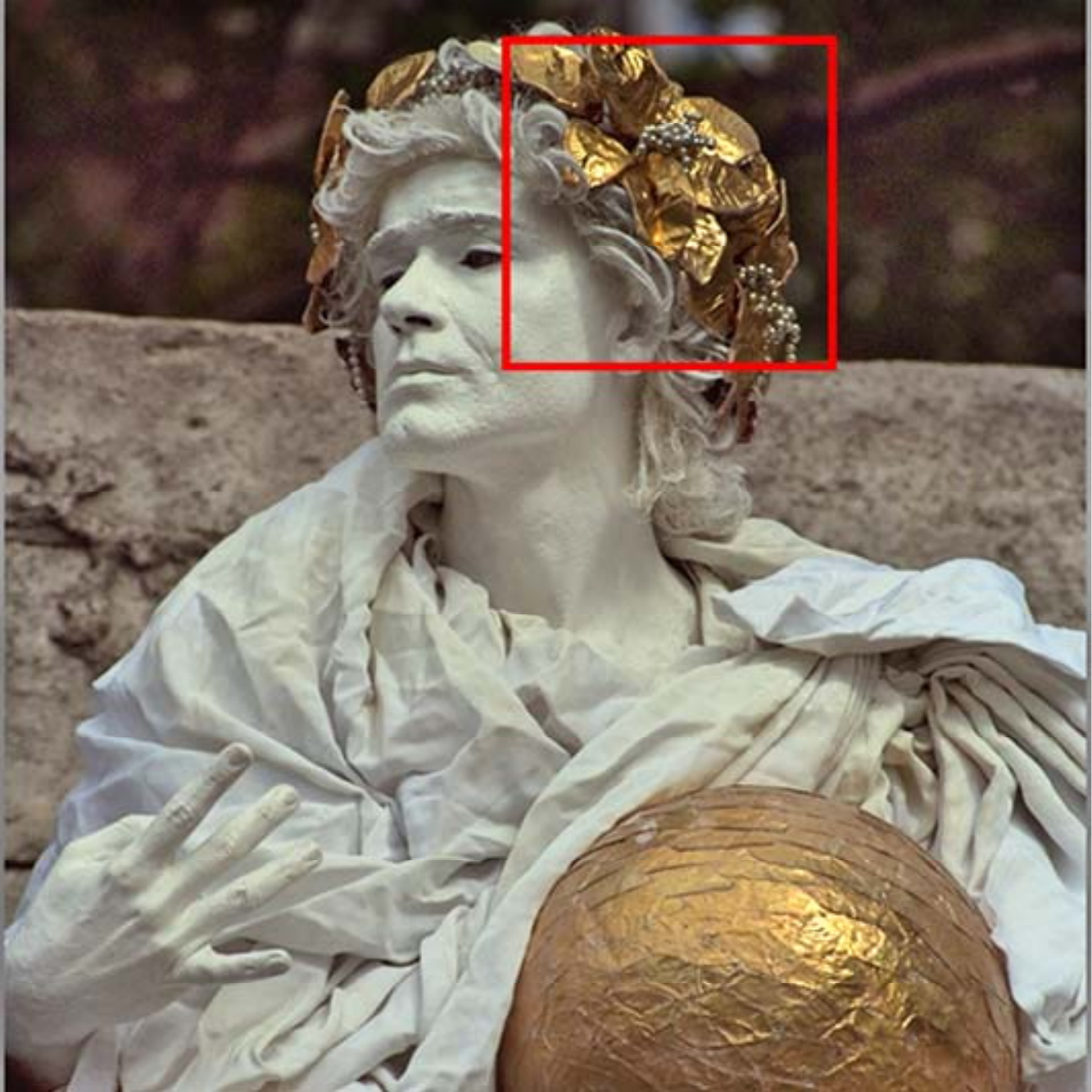}&
		\includegraphics[width=.23\linewidth]{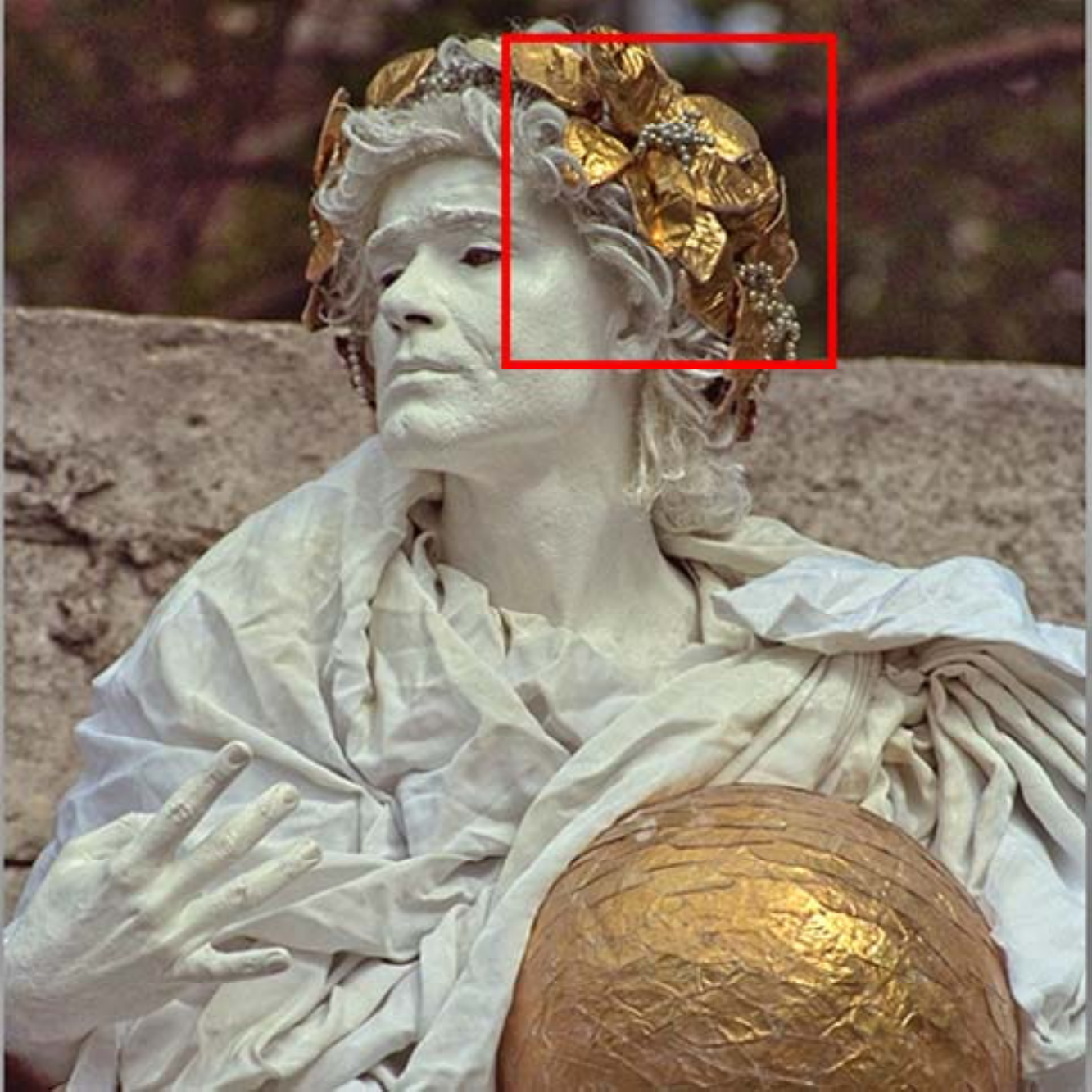}&
		\includegraphics[width=.23\linewidth]{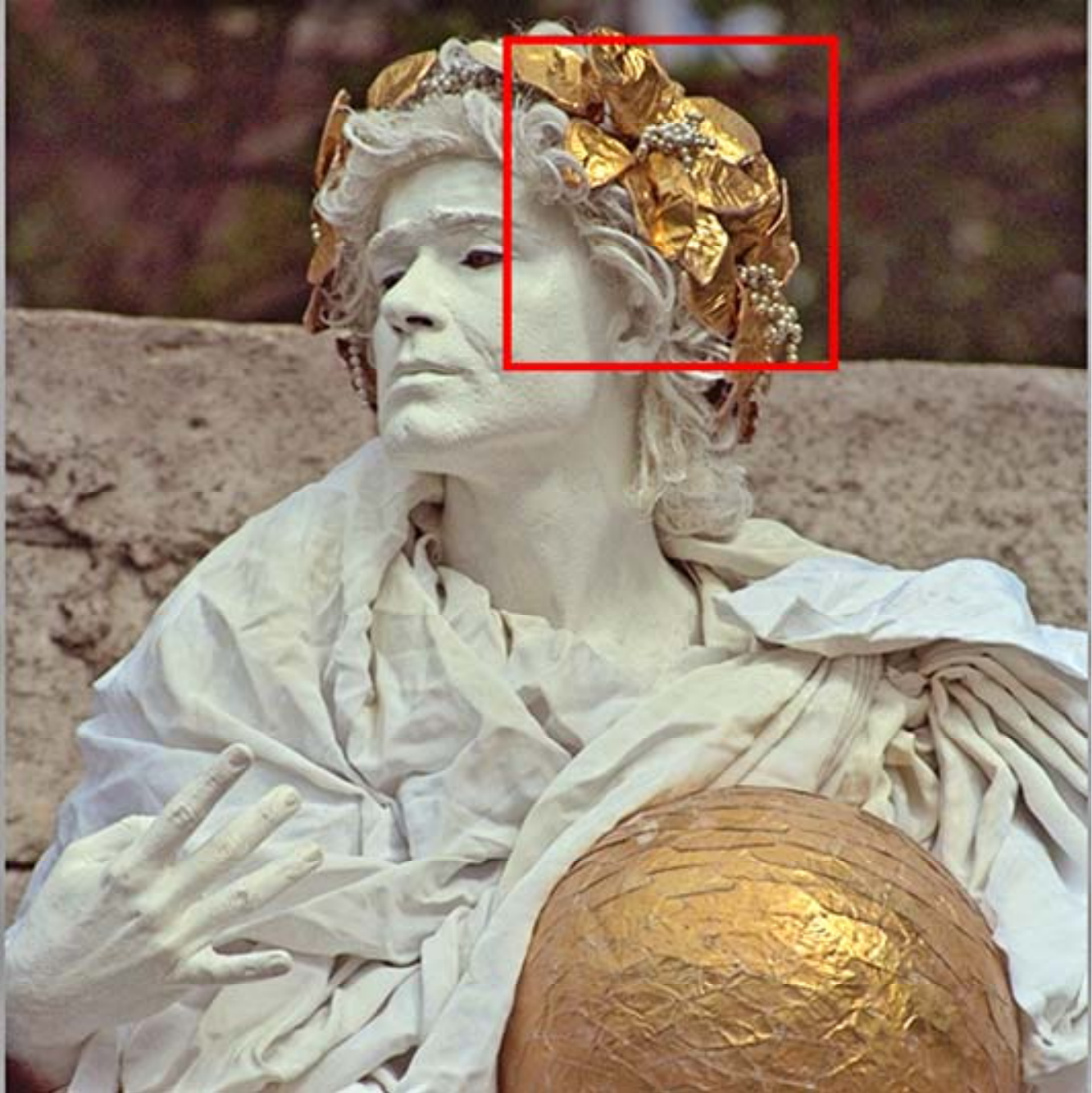}\\
		{SRIE}&{WVM}&{JIEP}&{Ours}\\
	\end{tabular}
	\caption{Comparing the image components for Retinex decomposition based methods (i.e., SRIE~\cite{fu2015probabilistic}, WVM~\cite{fu2016weighted}, JIEP~\cite{cai2017joint} and Ours). The illumination, reflectance and enhanced components are plotted on the top, middle, and bottom row, respectively. }
	\label{fig:decomp}
\end{figure}

\section{Experimental Results}\label{sec:exp}
In this section, we conduct a series of experiments to evaluate our algorithm. 
Since the reference images are unavailable, it is hard to assess the quantitative performance using standard metrics (e.g., PSNR). Thus we follow most exiting works to adopt Natural Image Quality Evaluator (NIQE)~\cite{mittal2013making} as our quantitative metric in all experiments. 
Please notice that the lower value of NIQE indicates a higher image quality. The decomposition is applied for the V-channel in the HSV (Hue, Saturation and Value) space, and
then transform it back to the RGB domain.
All these experiments are conducted on a PC with Intel Core i7-8700 CPU at 3.70GHz, 32 GB RAM and an NVIDIA GeForce GTX 1080 Ti 11GB GPU.

\begin{figure}[t]
	\centering	
	\begin{tabular}{c@{\extracolsep{0.3em}}c}
		\includegraphics[width=0.23\textwidth]{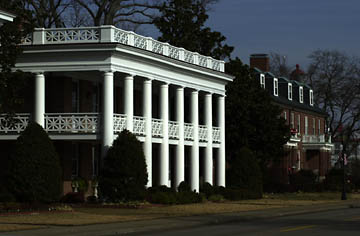}&
		\includegraphics[width=0.23\textwidth]{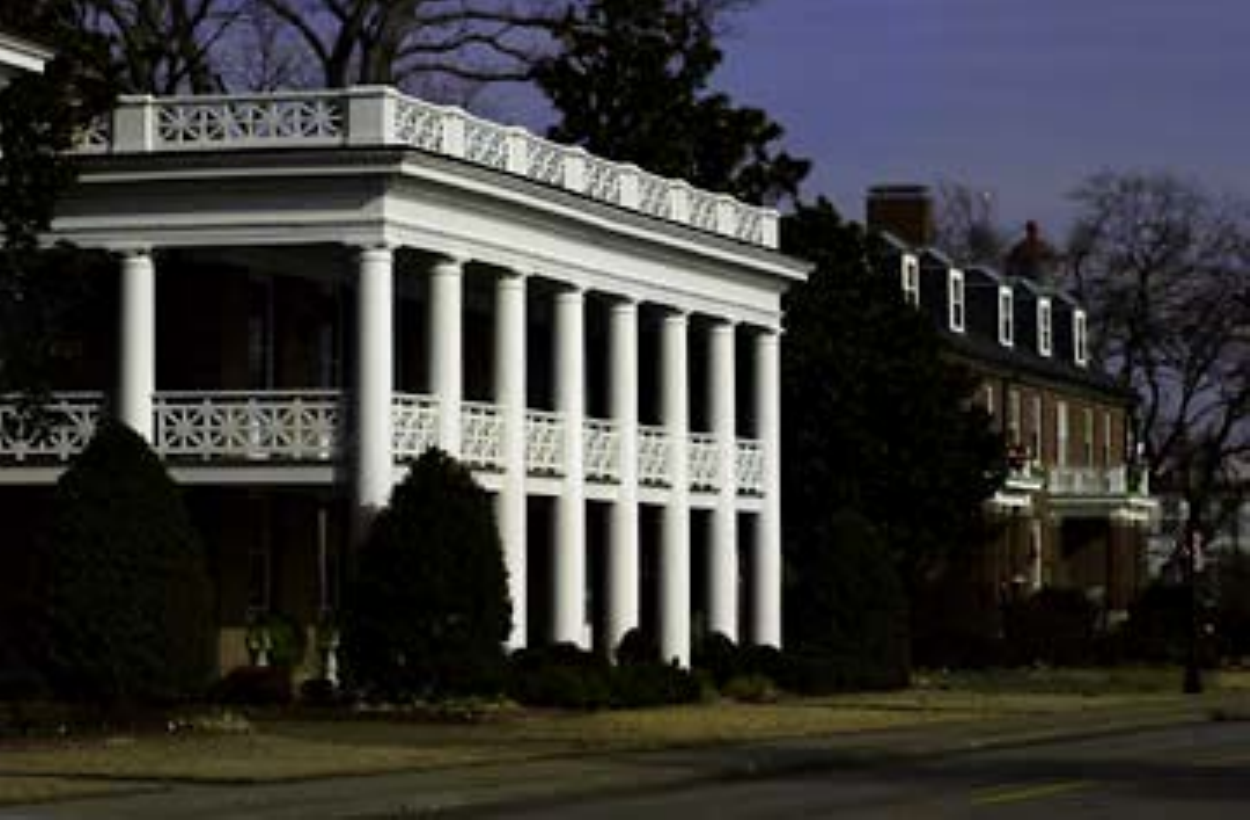}\\
		{Input}&{HDRNet}\\
		\includegraphics[width=0.23\textwidth]{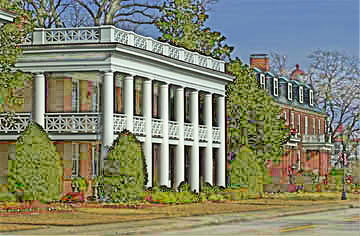}&
		\includegraphics[width=0.23\textwidth]{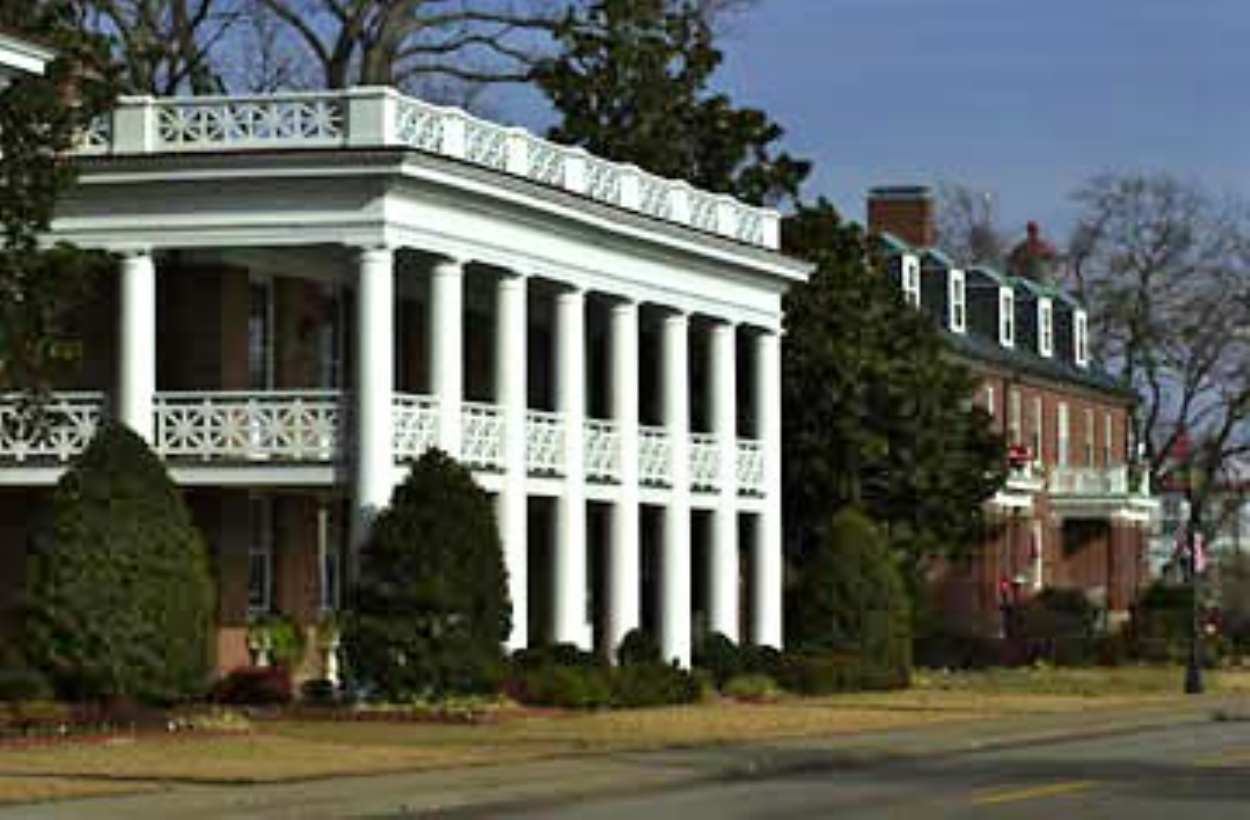}\\
		{RetinexNet}&{Ours}\\
	\end{tabular}
	\caption{Comparing the performance of two kinds of end-to-end learning-based method (i.e., HDRNet~\cite{hasinoff2017Deep}, RetinexNet~\cite{Chen2018Retinex}) and our proposed hybrid prior method.}
	\label{fig:VC0}
\end{figure}

\begin{figure*}[t]
	\centering
	\begin{tabular}{c@{\extracolsep{2em}}c}
		\includegraphics[width=0.45\linewidth]{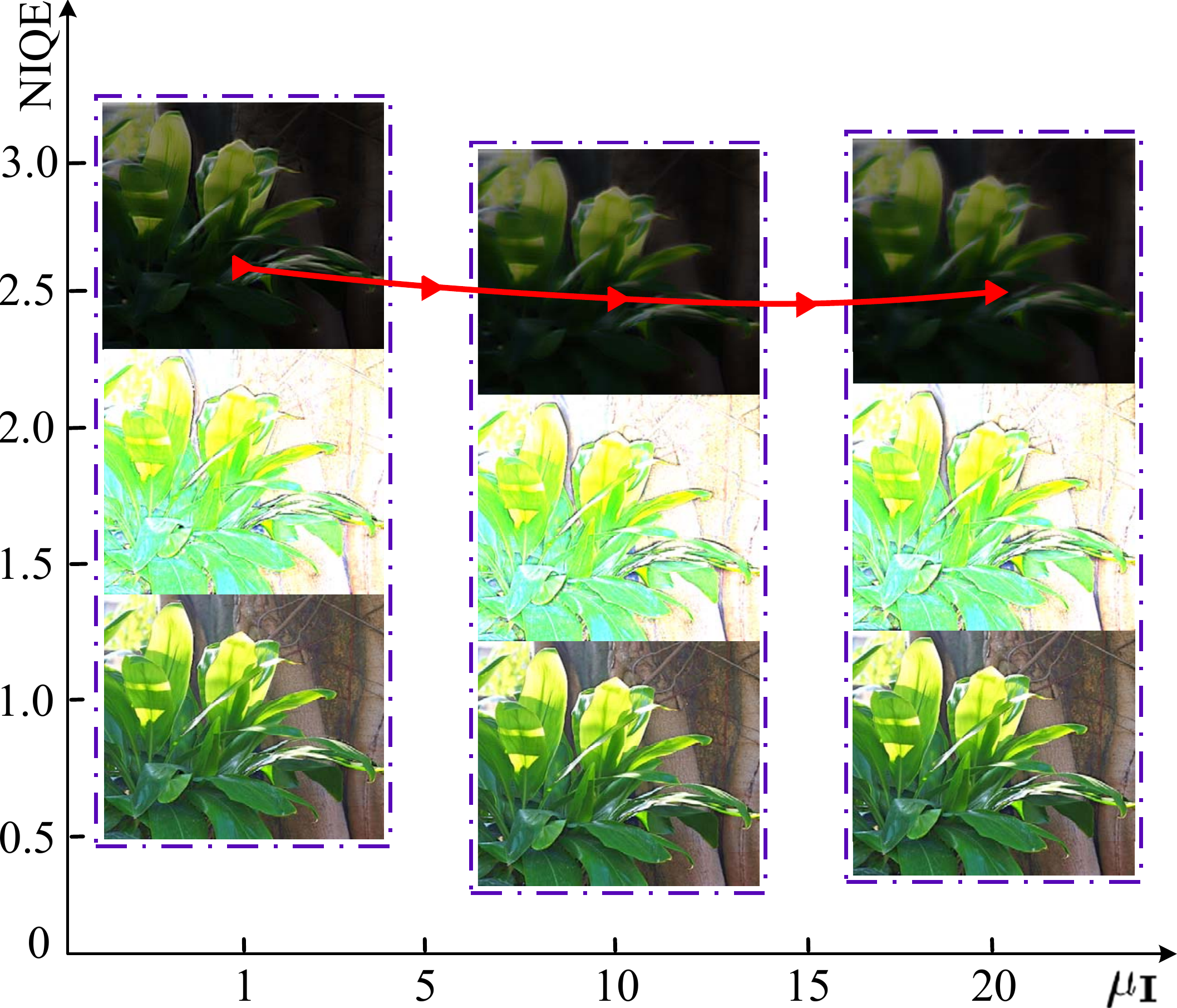}&
		\includegraphics[width=0.45\linewidth]{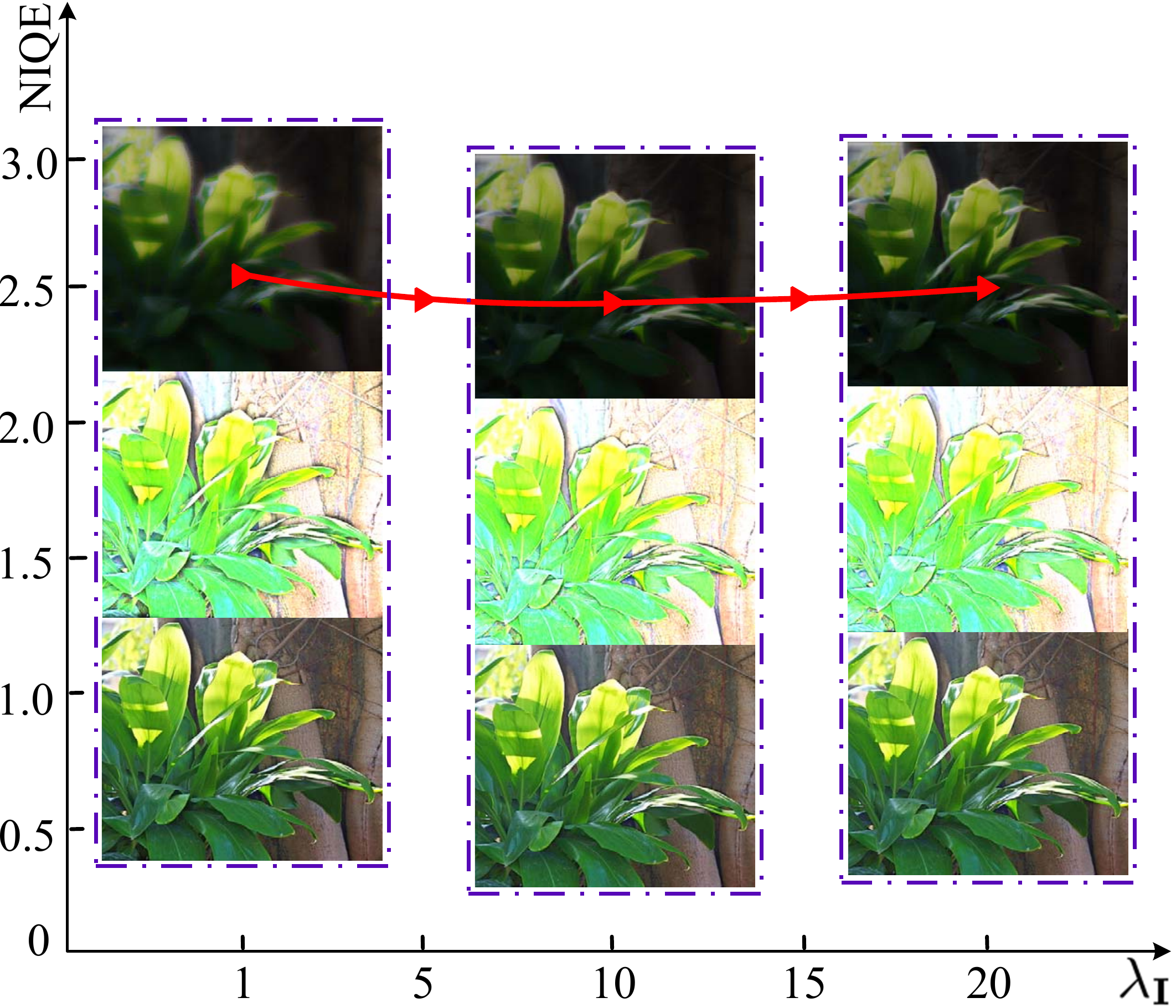}\\
		(a) NIQE vs. $\mu_{\mathbf{I}}$ &(b) NIQE vs. $\lambda_{\mathbf{I}}$\\
	\end{tabular}
	\caption{The visual performances and quantitative results of our method w.r.t the settings of algorithmic parameters. The NIQE (lower is better) curves of $\mu_\mathbf{I}$ and $\lambda_\mathbf{I}$ are illustrated in subfigures (a) and (b), respectively. With each parameter setting, we plot illumination (top), reflectance (middle) and the final enhanced result (bottom) in the dashed rectangles.}
	\label{fig:IlluminationParameters}
\end{figure*}
\begin{figure*}[t]
	\centering
	\begin{tabular}{c@{\extracolsep{0.5em}}c@{\extracolsep{0.5em}}c}
		\includegraphics[width=0.32\linewidth]{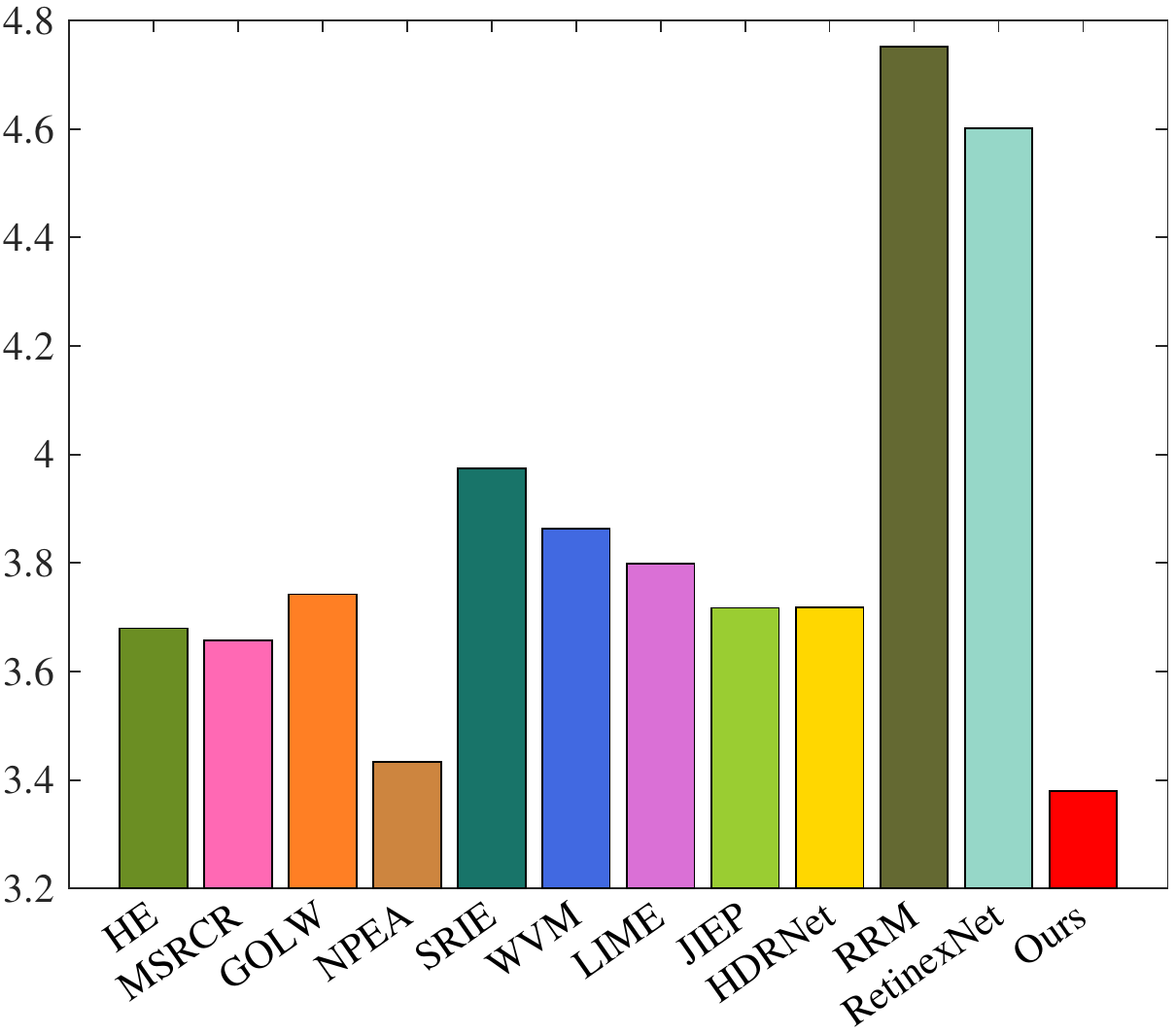}&
		\includegraphics[width=0.32\linewidth]{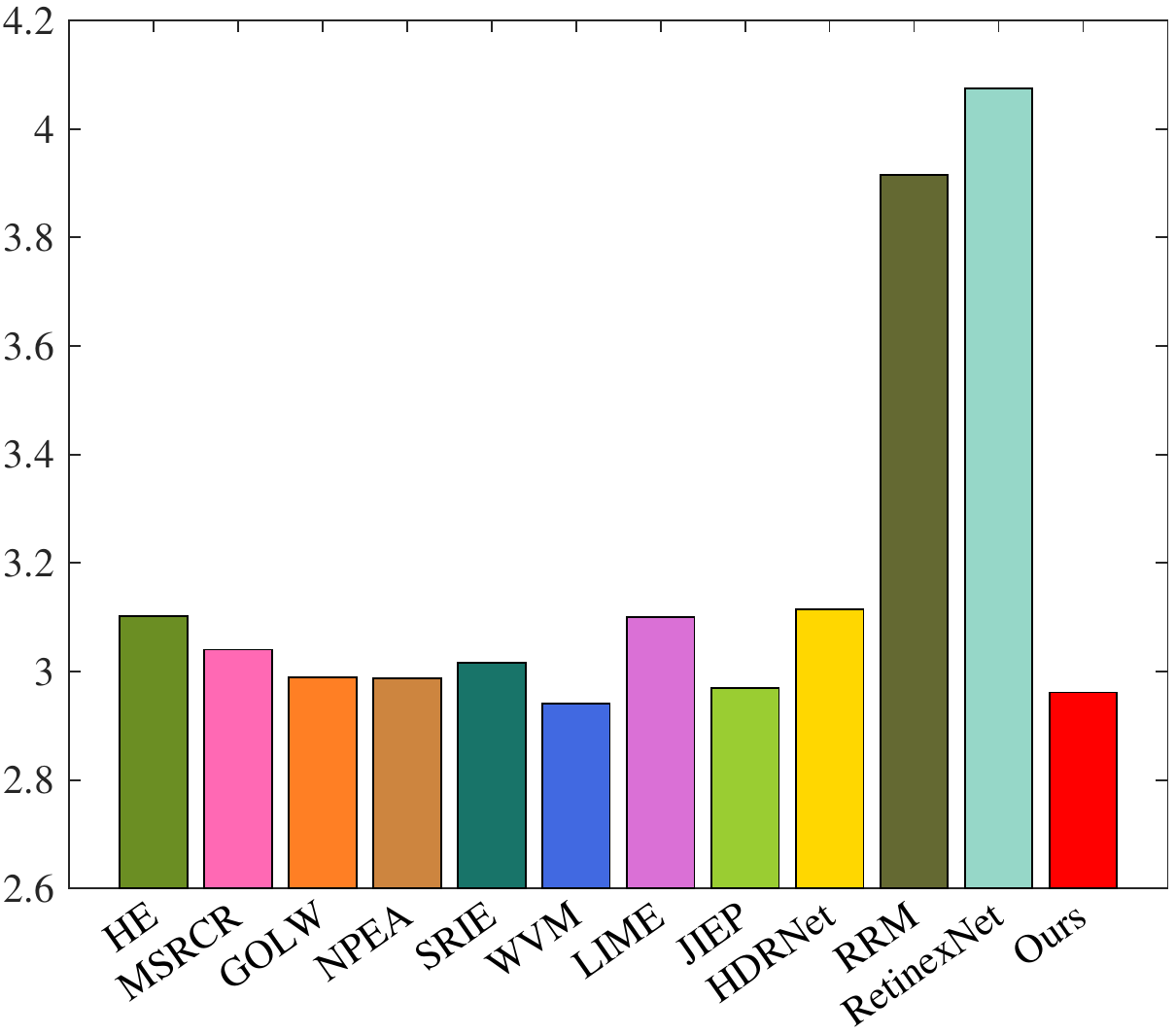}&
		\includegraphics[width=0.32\linewidth]{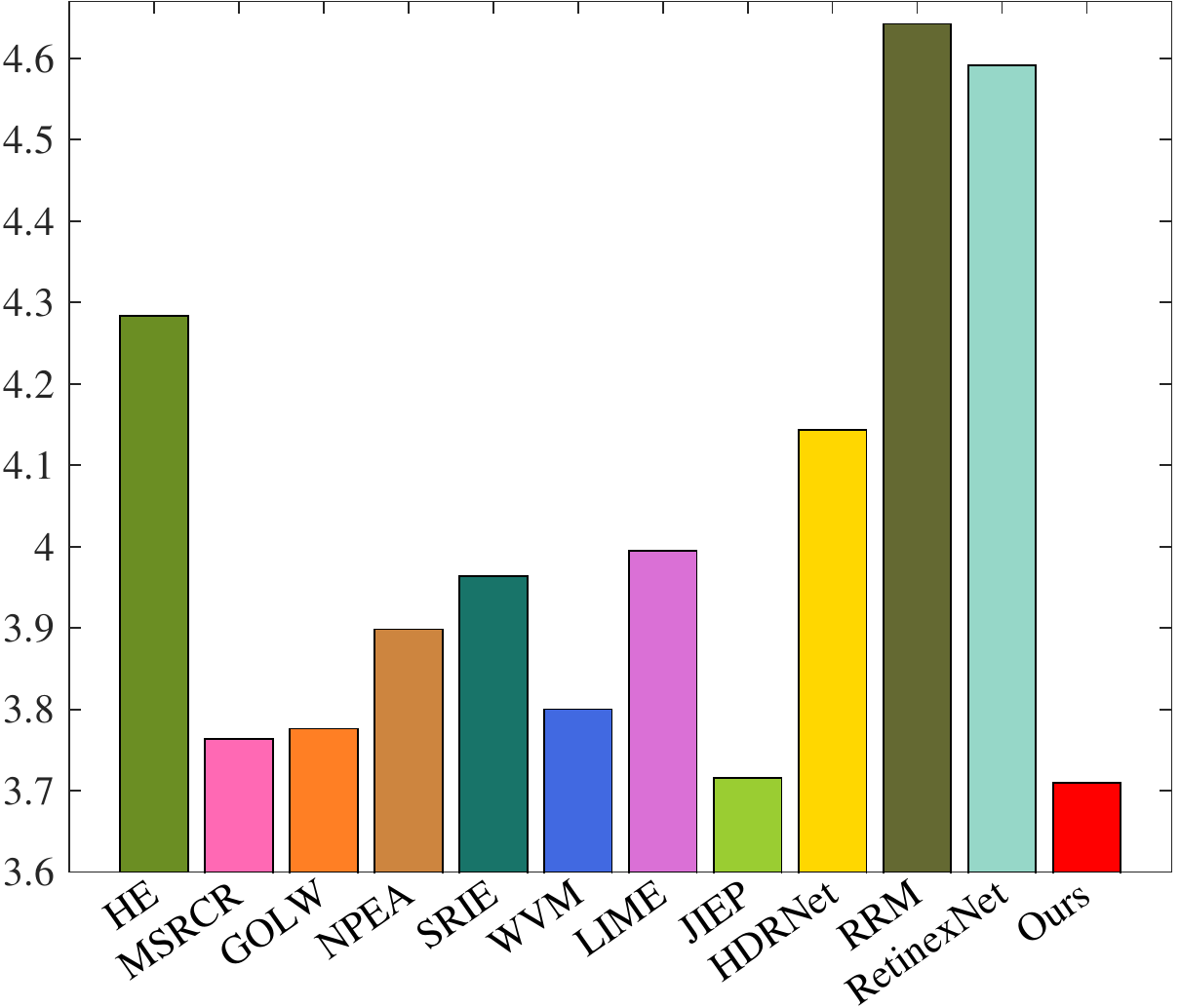}\\
		(a) NASA& (b) NPE& (c) LIME\\
	\end{tabular}
	\caption{Quantitative performance (i.e., NIQE, lower is better) on three different benchmark databases.}
	\label{fig:rescomp}
\end{figure*}

\begin{figure*}[!htb]
	\centering
	\begin{tabular}{c@{\extracolsep{0.5em}}c@{\extracolsep{0.5em}}c@{\extracolsep{0.5em}}c}
		\includegraphics[width=.24\linewidth]{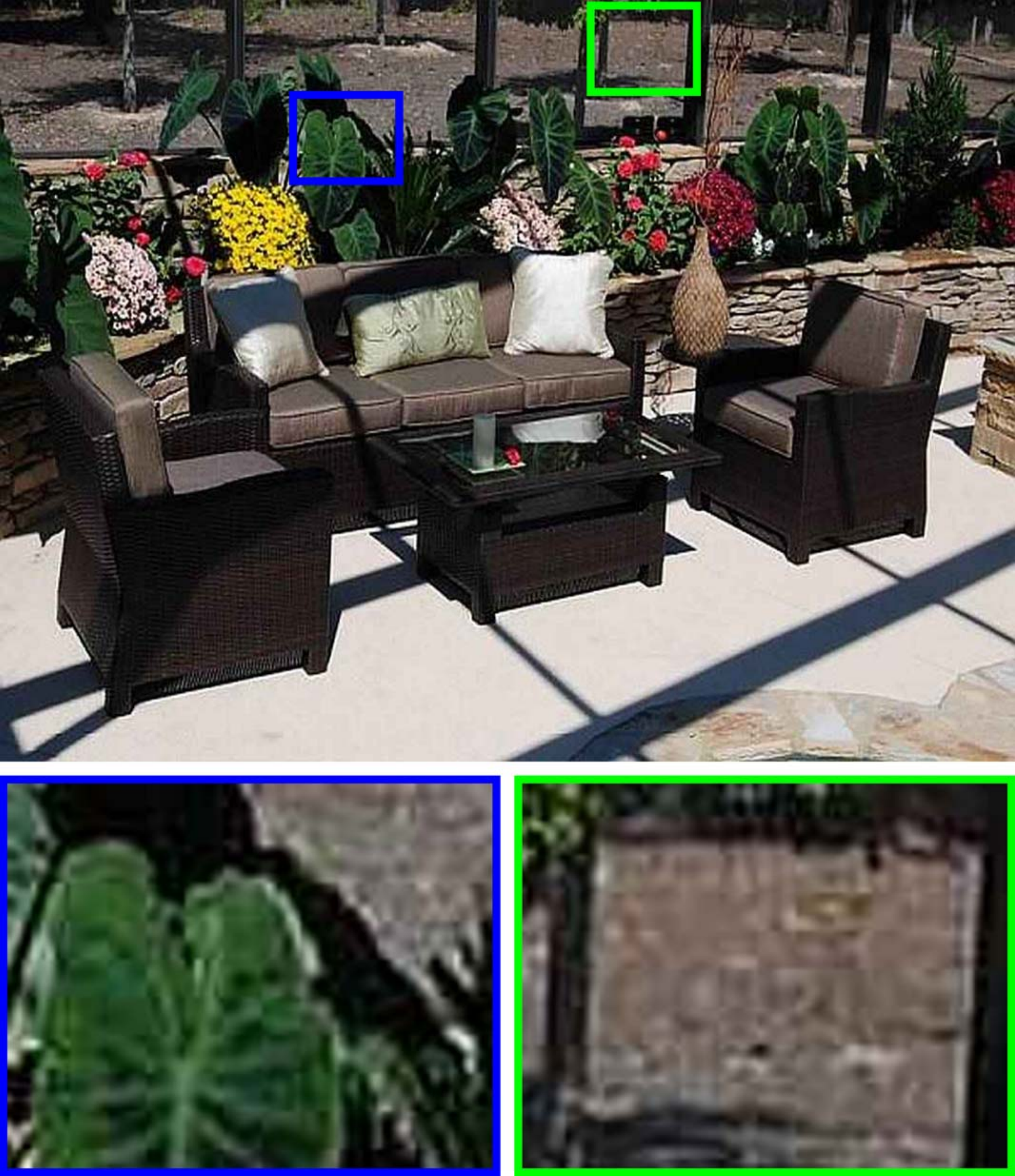}&
		\includegraphics[width=.24\linewidth]{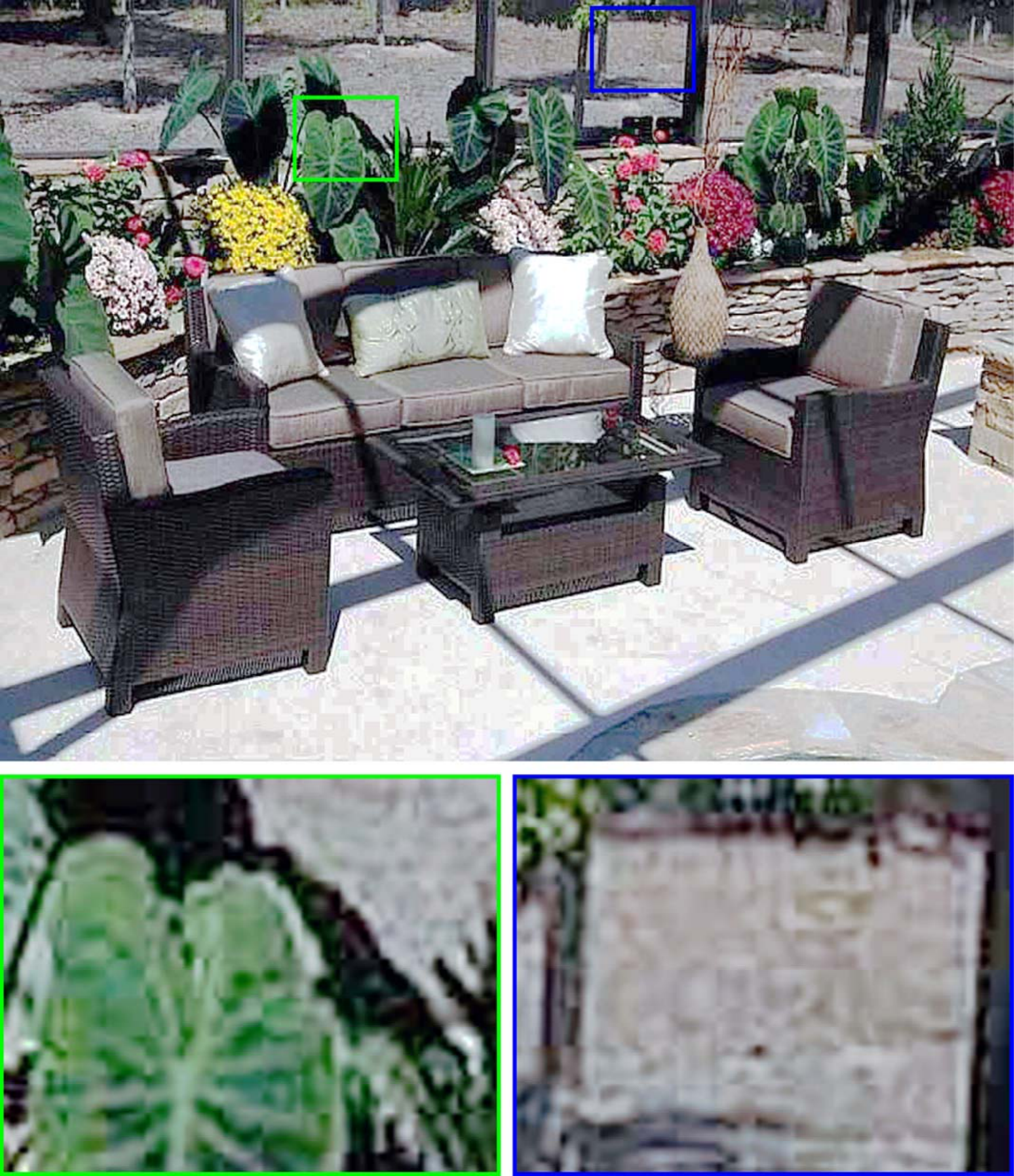}&
		\includegraphics[width=.24\linewidth]{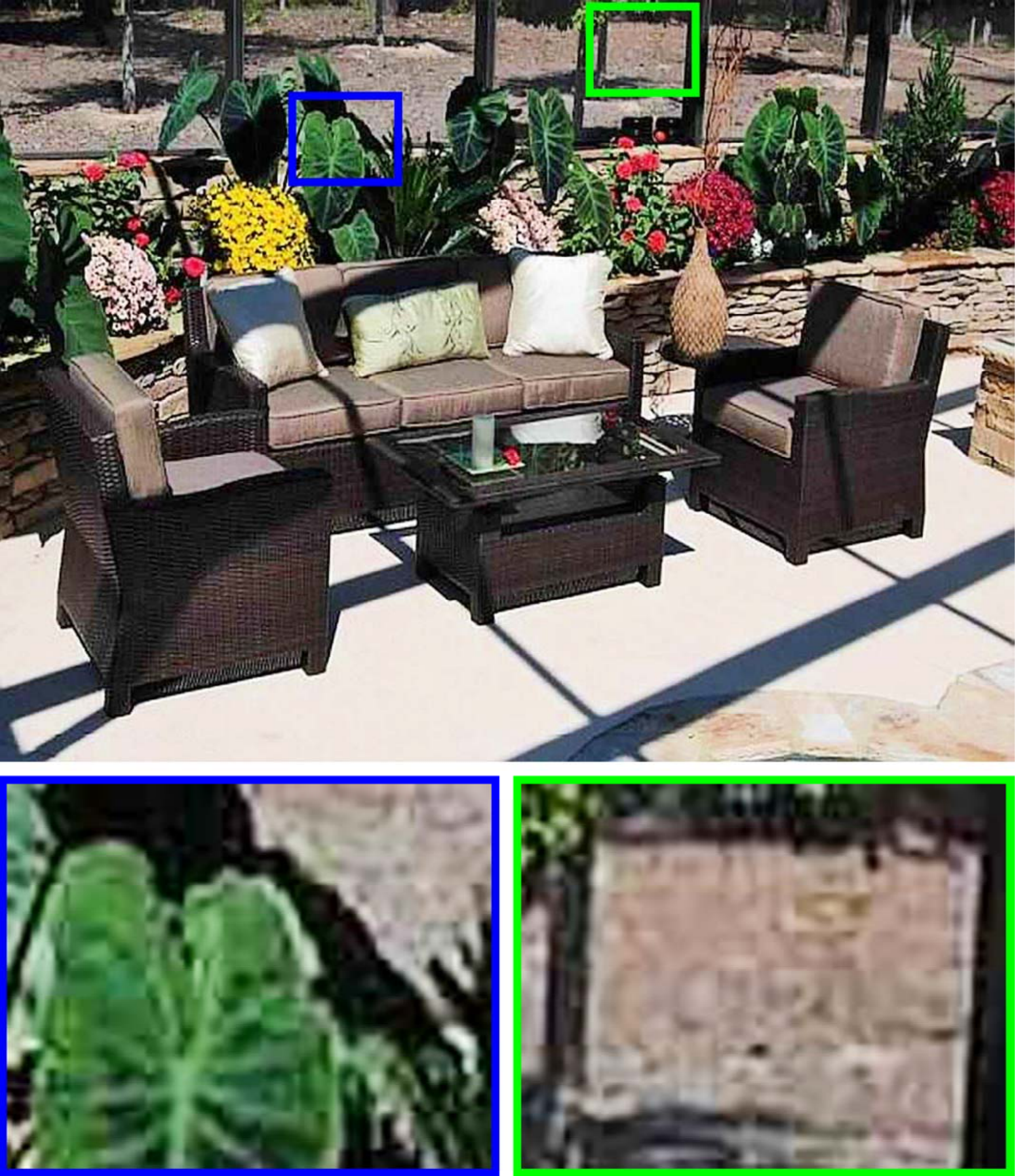}&
		\includegraphics[width=.24\linewidth]{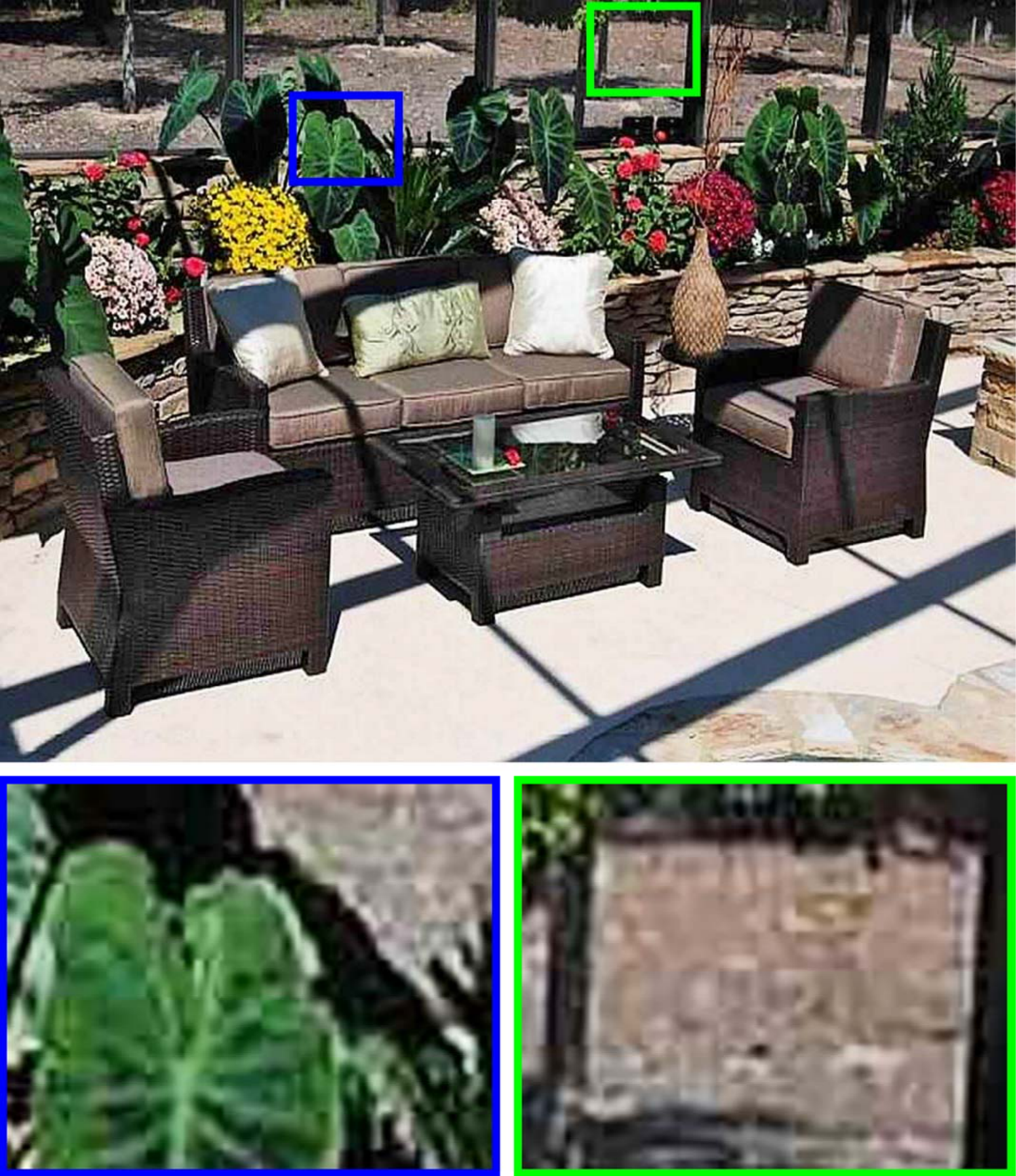}\\
		{Input} &{HE} ({3.9318})&{SRIE} ({4.0469})&{WVM} ({3.9181})\\
		\includegraphics[width=.24\linewidth]{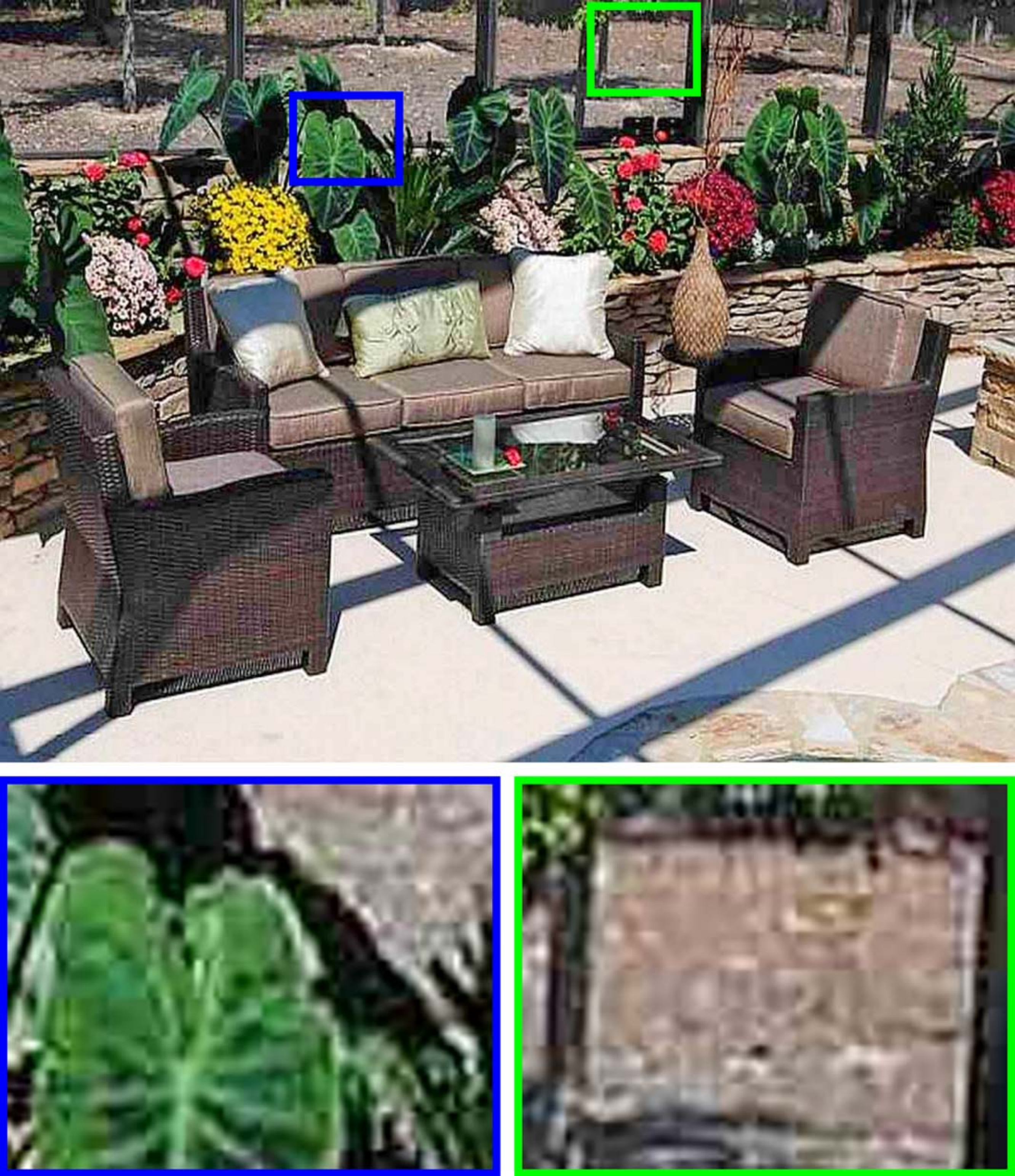}&
		\includegraphics[width=.24\linewidth]{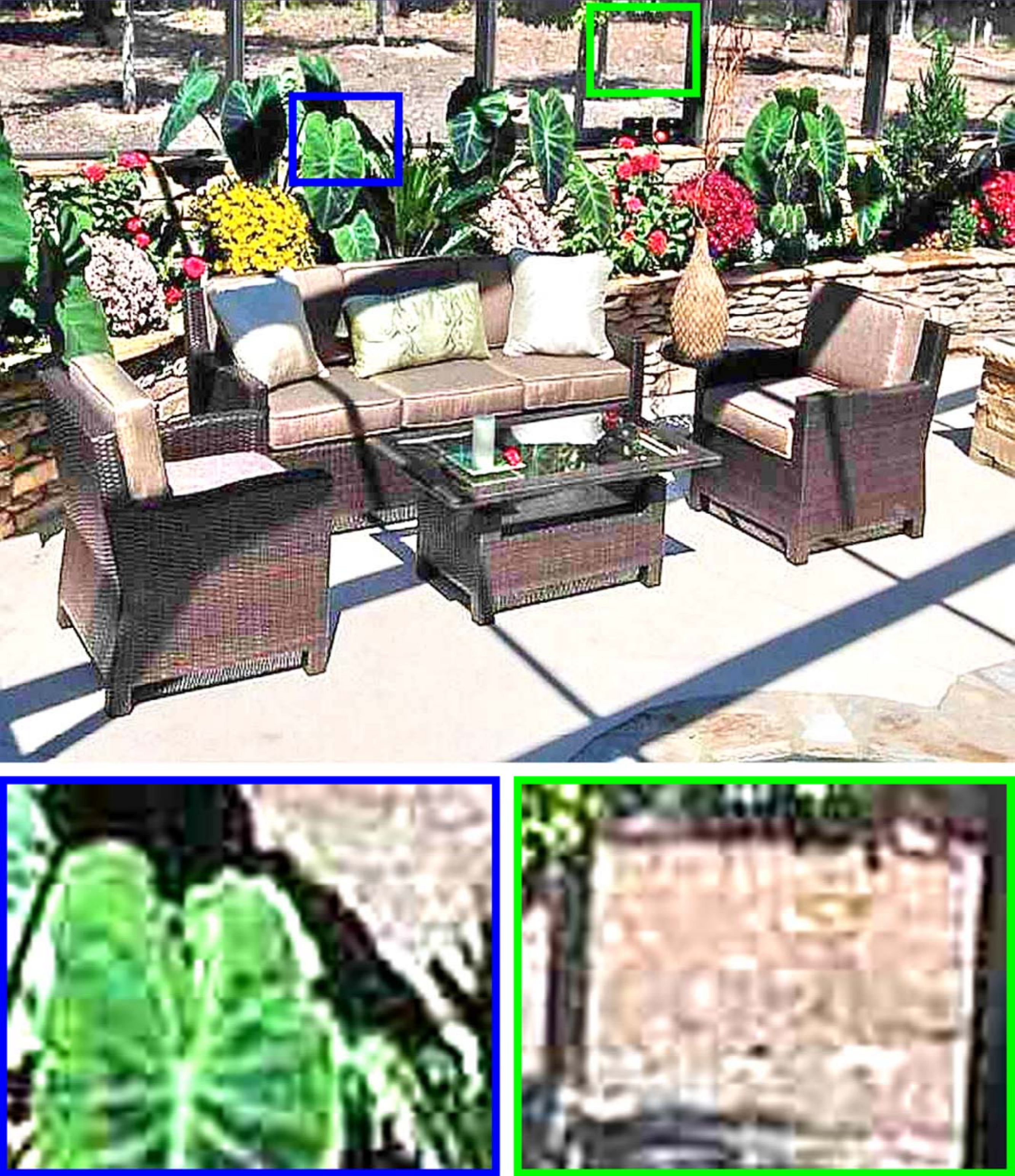}&
		\includegraphics[width=.24\linewidth]{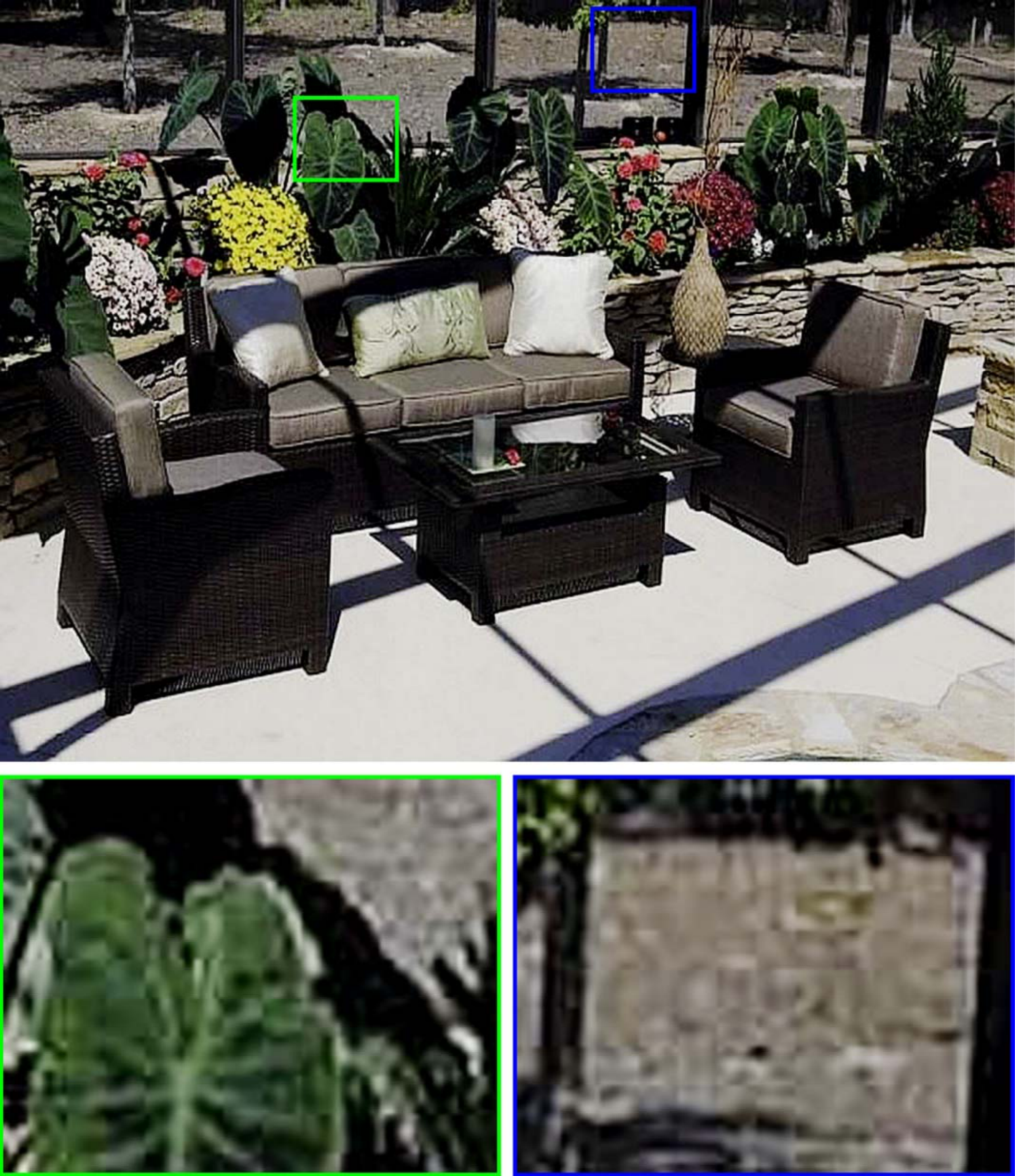}&
		\includegraphics[width=.24\linewidth]{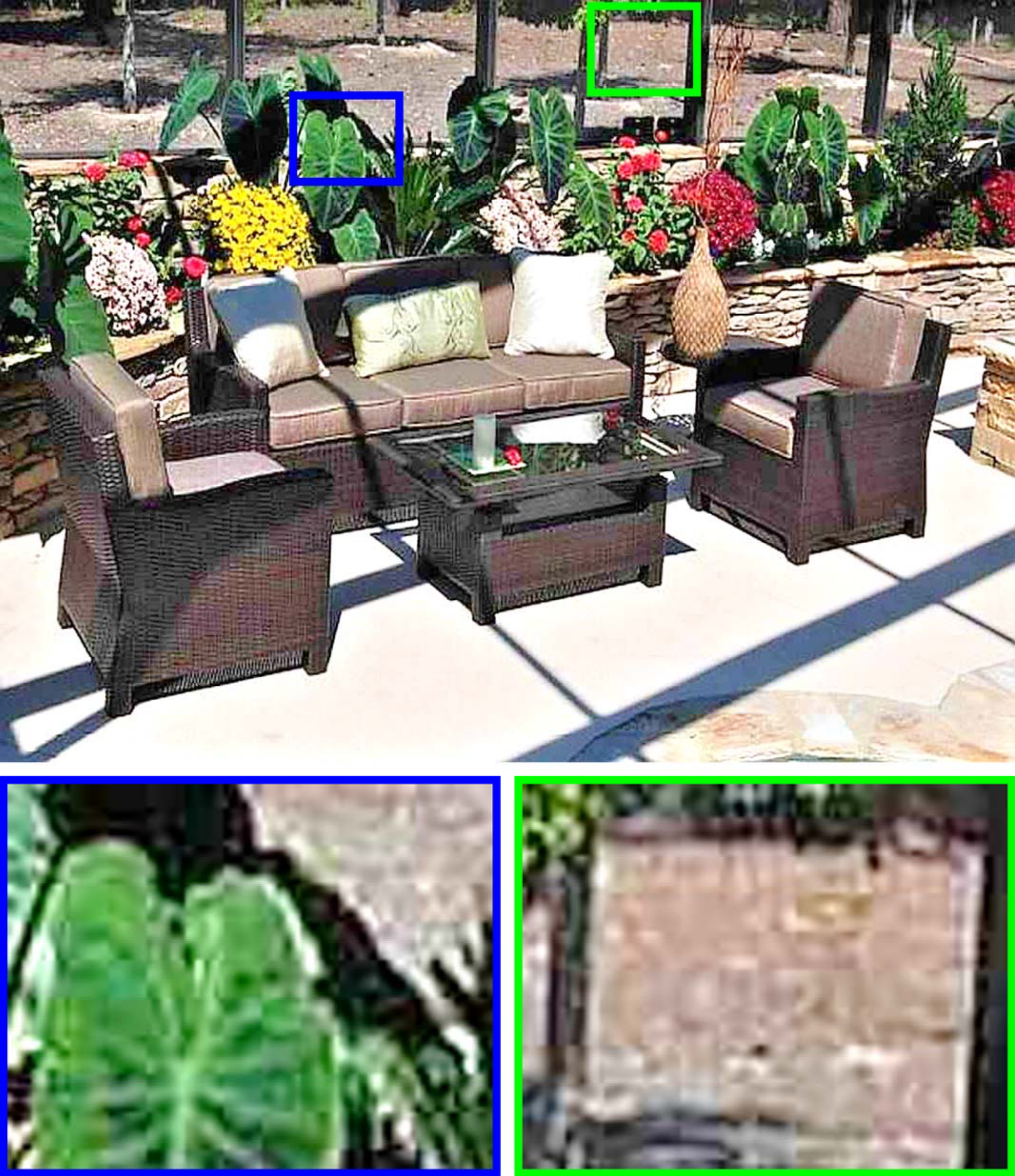}\\
		{JIEP} ({4.2309})&LIME ({4.3105})&{HDRNet} ({4.1184})&{Ours} ({\textbf{3.8763}})\\ 
	\end{tabular}
	\caption{Comparisons on an example in Non-uniform dataset. The NIQE scores are reported below each image.}
	\label{fig:nonuniform}
\end{figure*}

\begin{figure*}[!htb]
	\centering
	\begin{tabular}{c@{\extracolsep{0.5em}}c@{\extracolsep{0.5em}}c}
		\includegraphics[width=.32\linewidth]{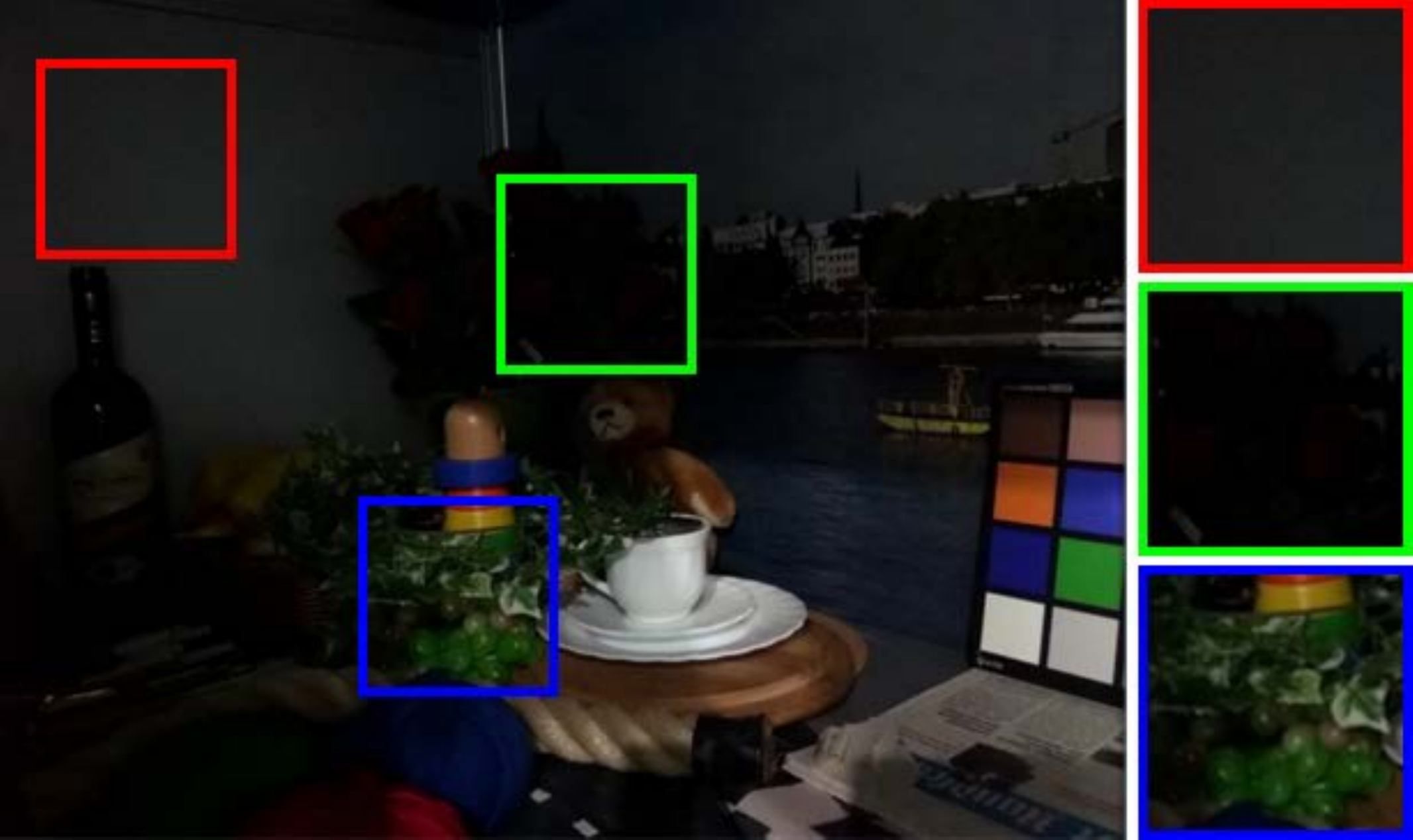}&
		\includegraphics[width=.32\linewidth]{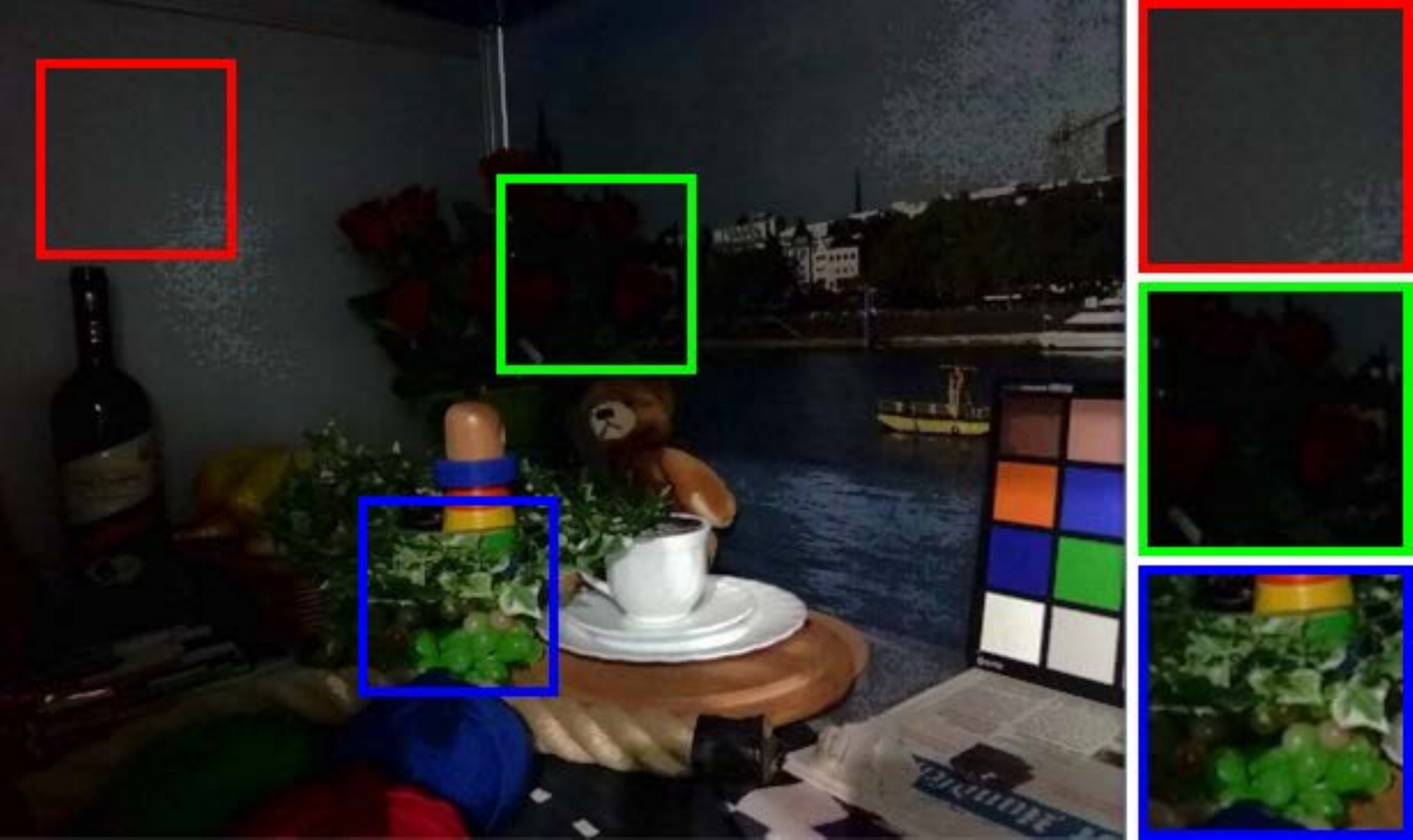}&
		\includegraphics[width=.32\linewidth]{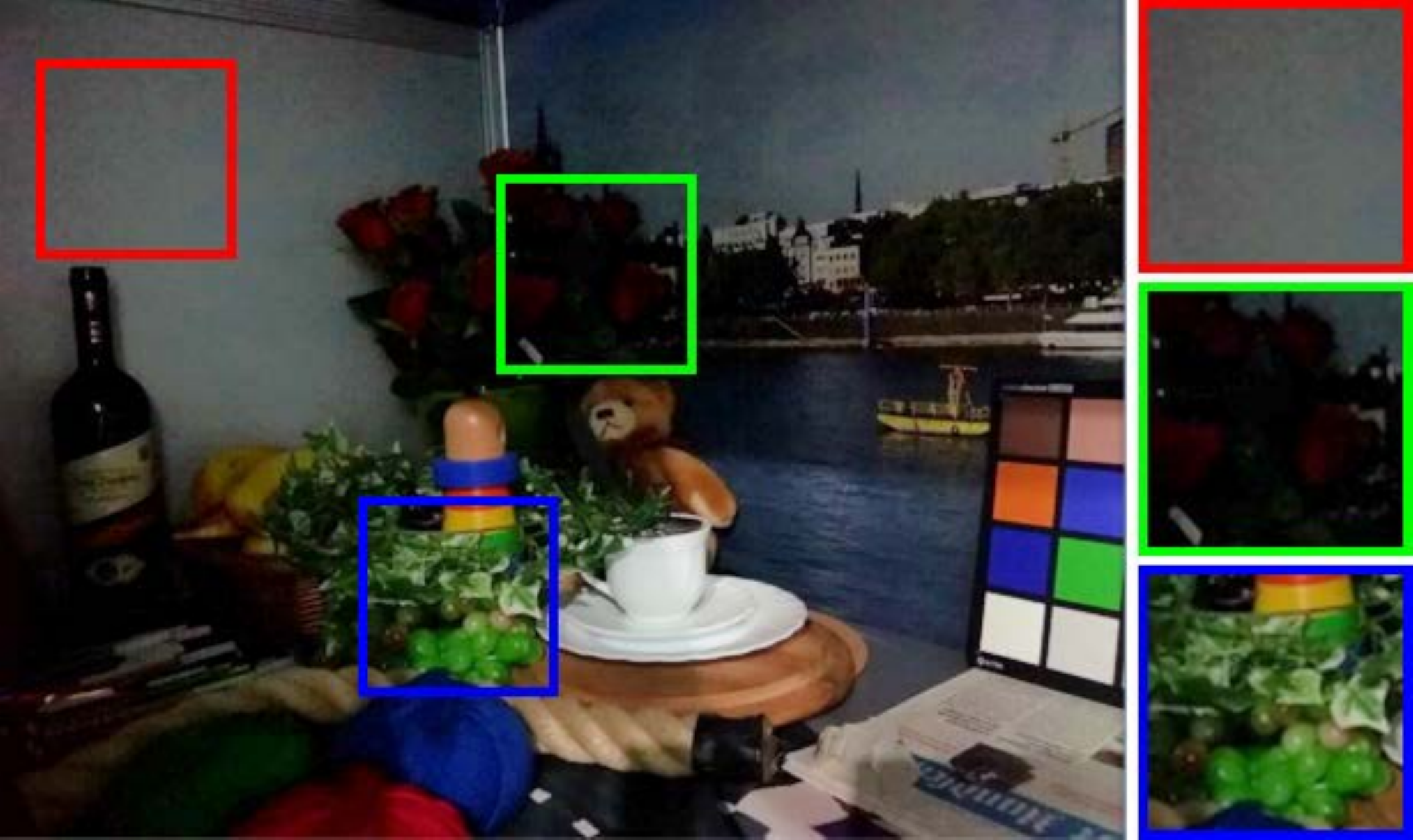}\\
		{ Input }&  HE (3.0200)&  {SRIE} (3.0925)\\
		\includegraphics[width=.32\linewidth]{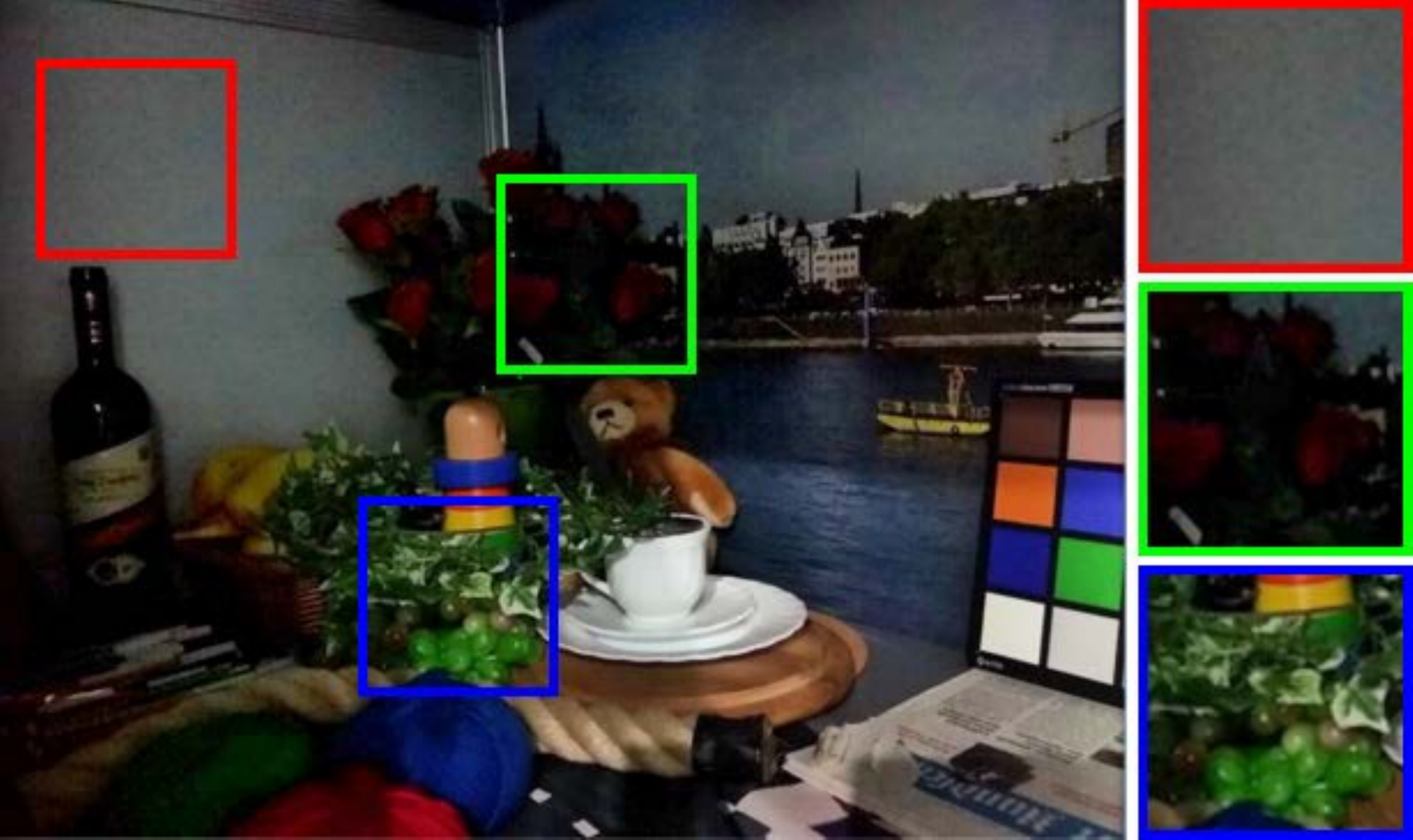}&
		\includegraphics[width=.32\linewidth]{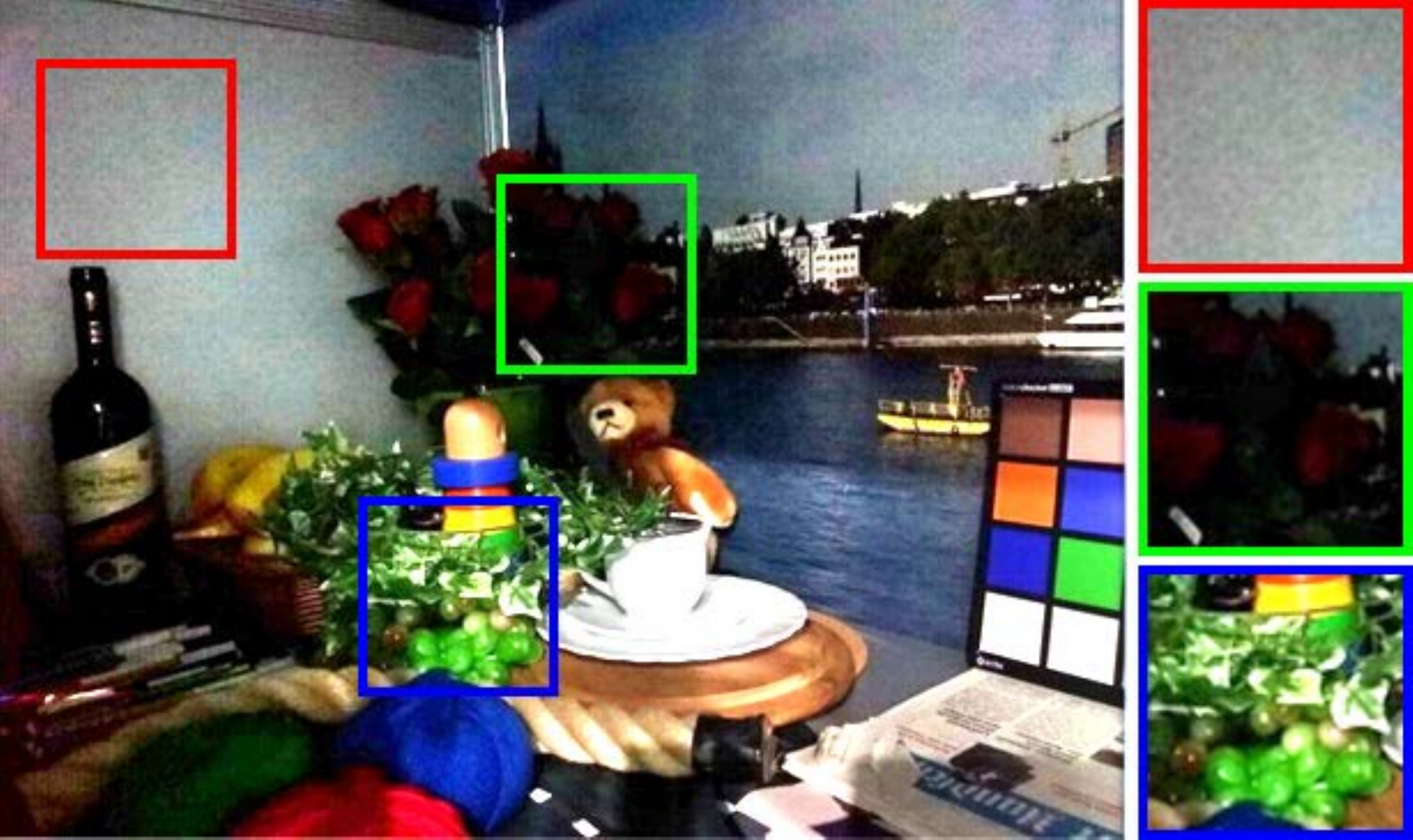}&
		\includegraphics[width=.32\linewidth]{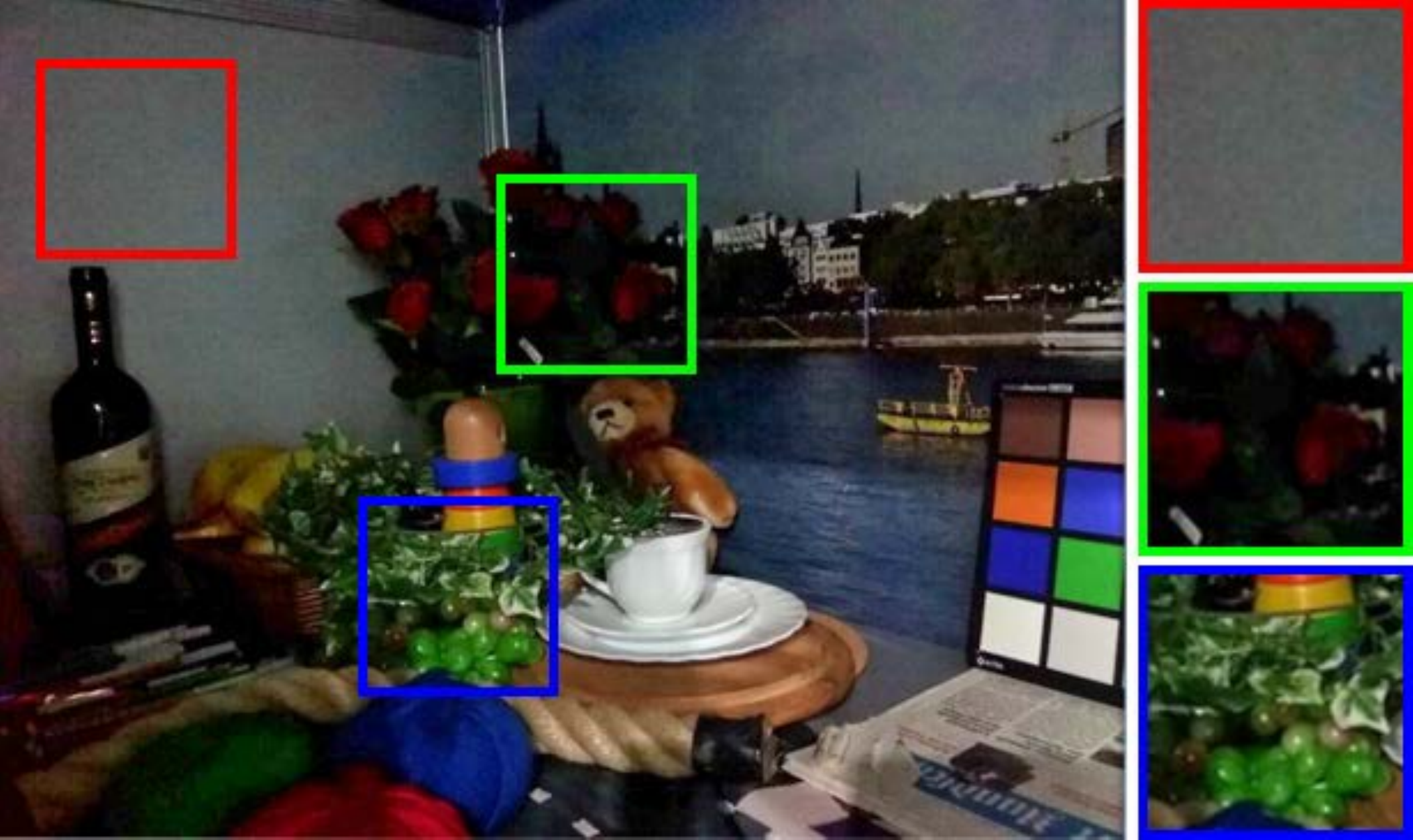}\\
		{ WVM (2.8731)}&  LIME ({2.9322})&  {JIEP} ({2.8731})\\
		\includegraphics[width=.32\linewidth]{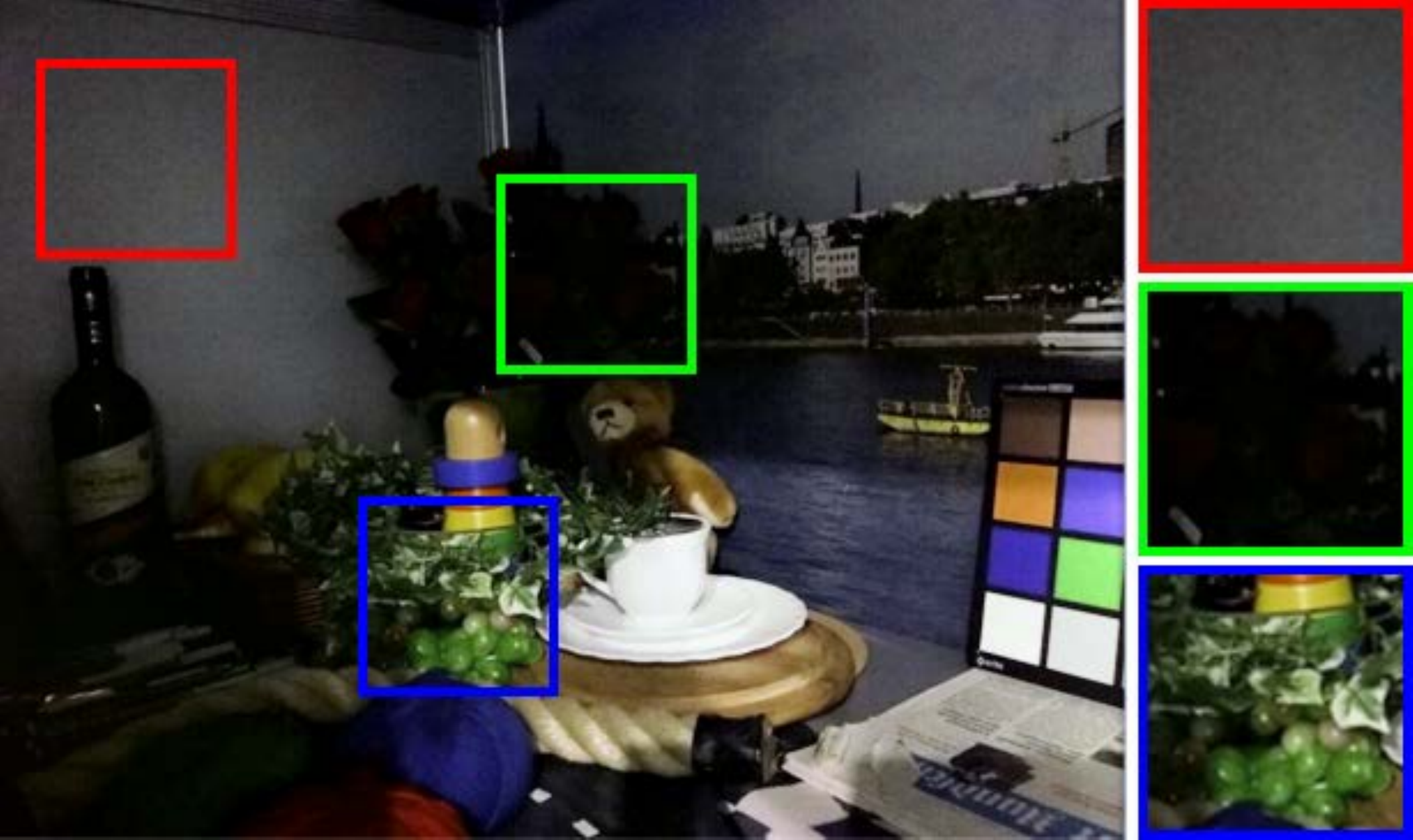}&
		\includegraphics[width=.32\linewidth]{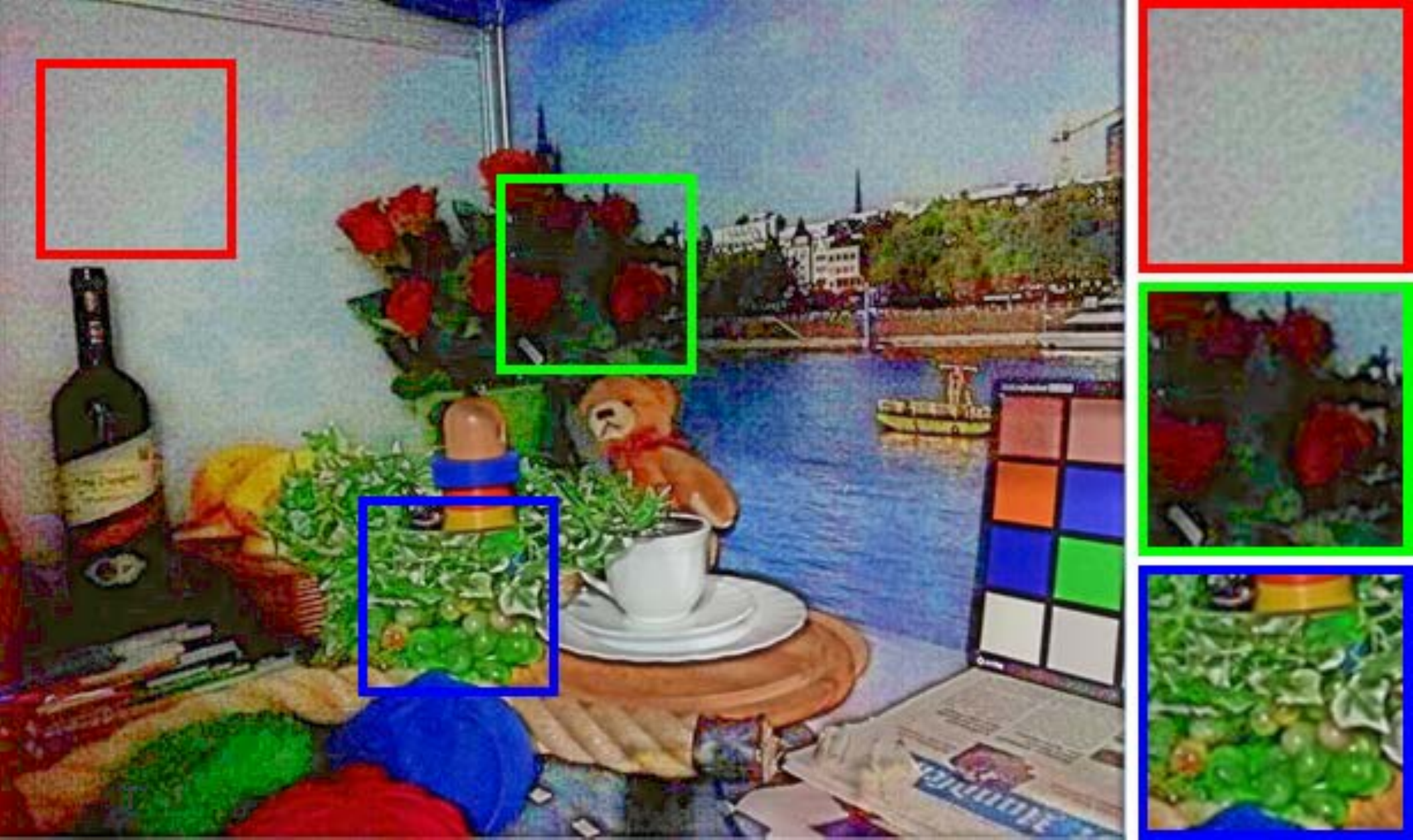}&
		\includegraphics[width=.32\linewidth]{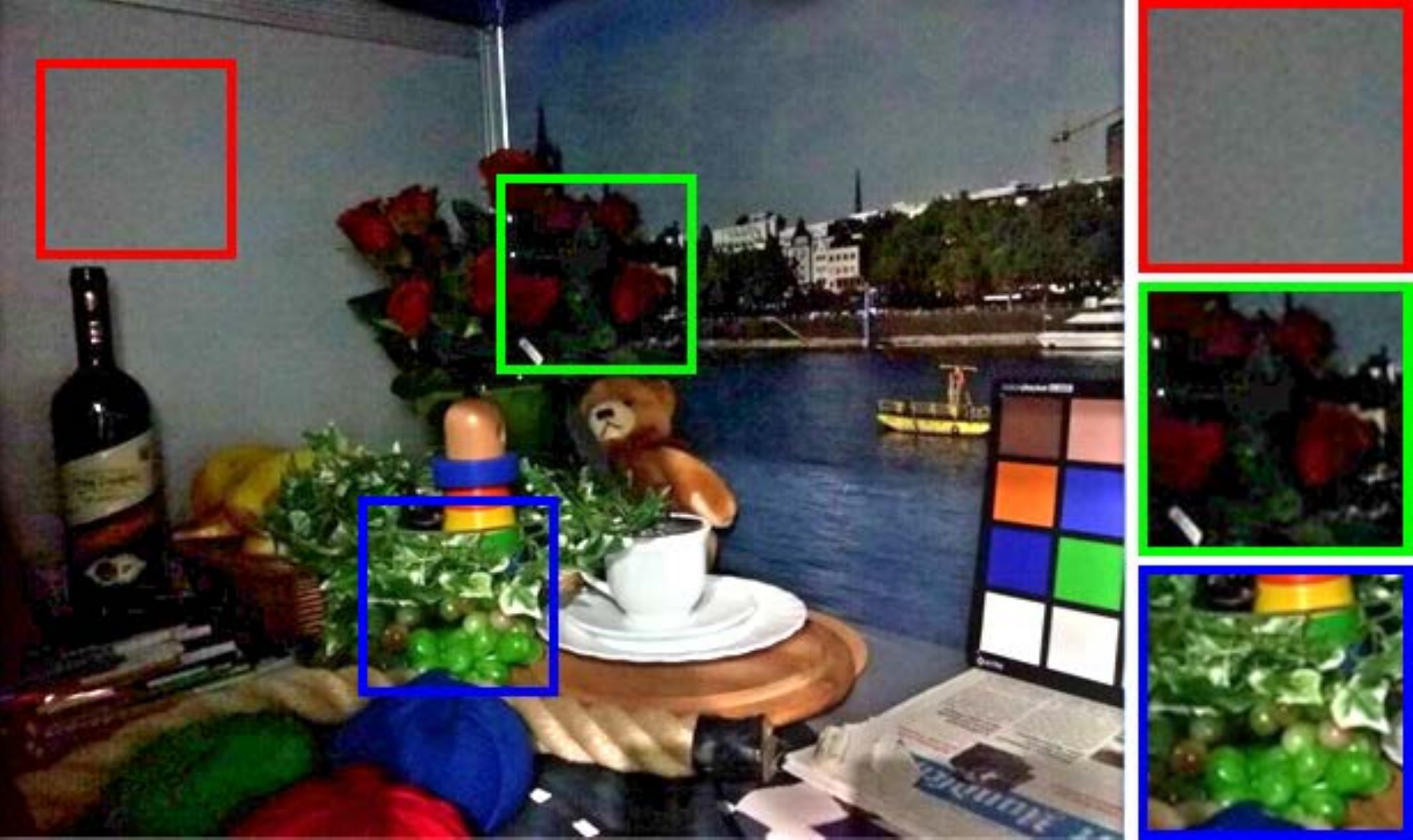}\\
		  HDRNet (3.0753) & {RetinexNet} ({2.8091})&  {Ours} ({\textbf{2.6186}})\\
	\end{tabular}
	\caption{Comparisons on an example in LIME dataset. The NIQE scores are reported in the brackets.}
	\label{fig:lime}
\end{figure*}
\begin{figure*}[!htb]
	\centering
	\begin{tabular}{c@{\extracolsep{0.5em}}c@{\extracolsep{0.5em}}c}
		\includegraphics[width=.32\linewidth]{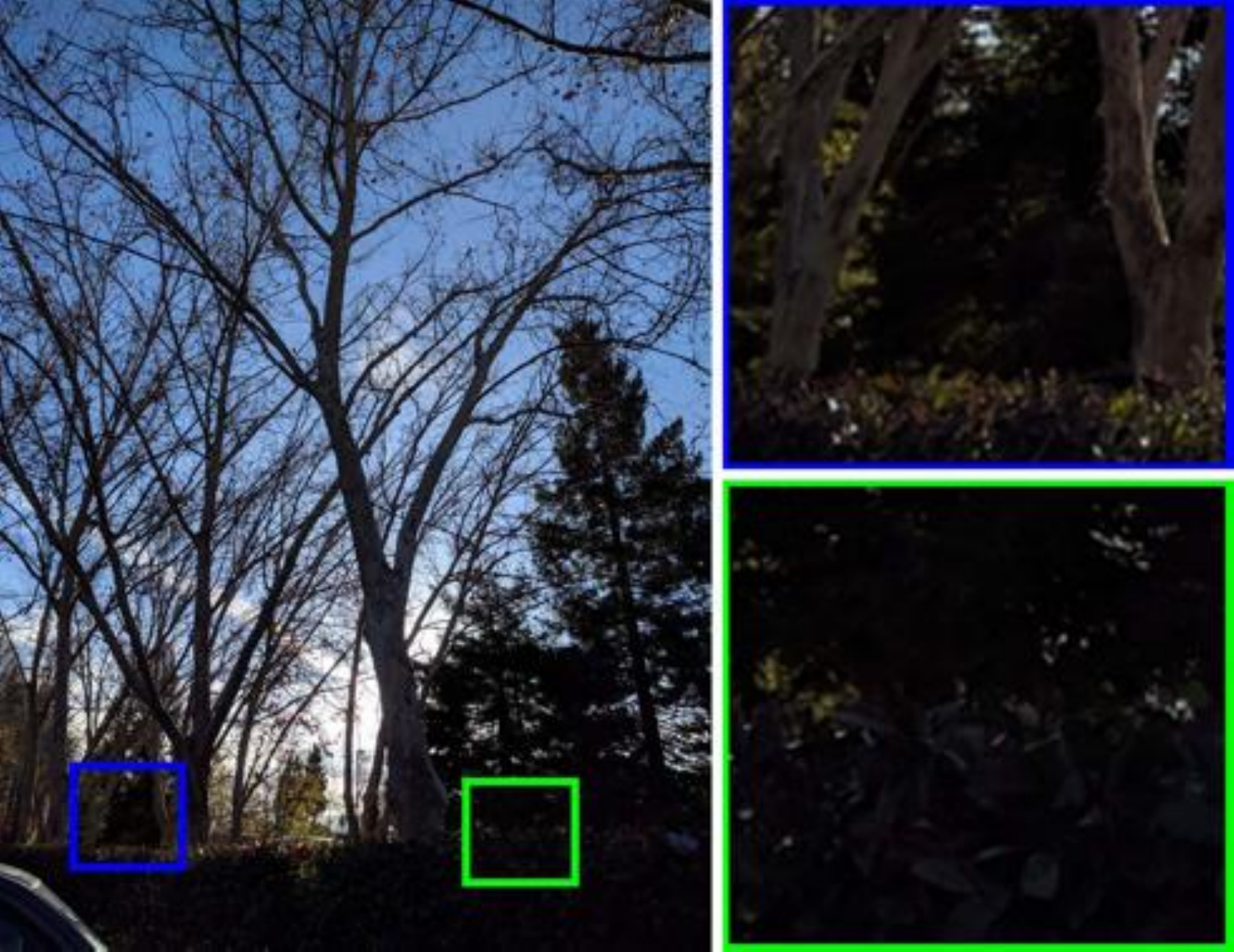}&
		\includegraphics[width=.32\linewidth]{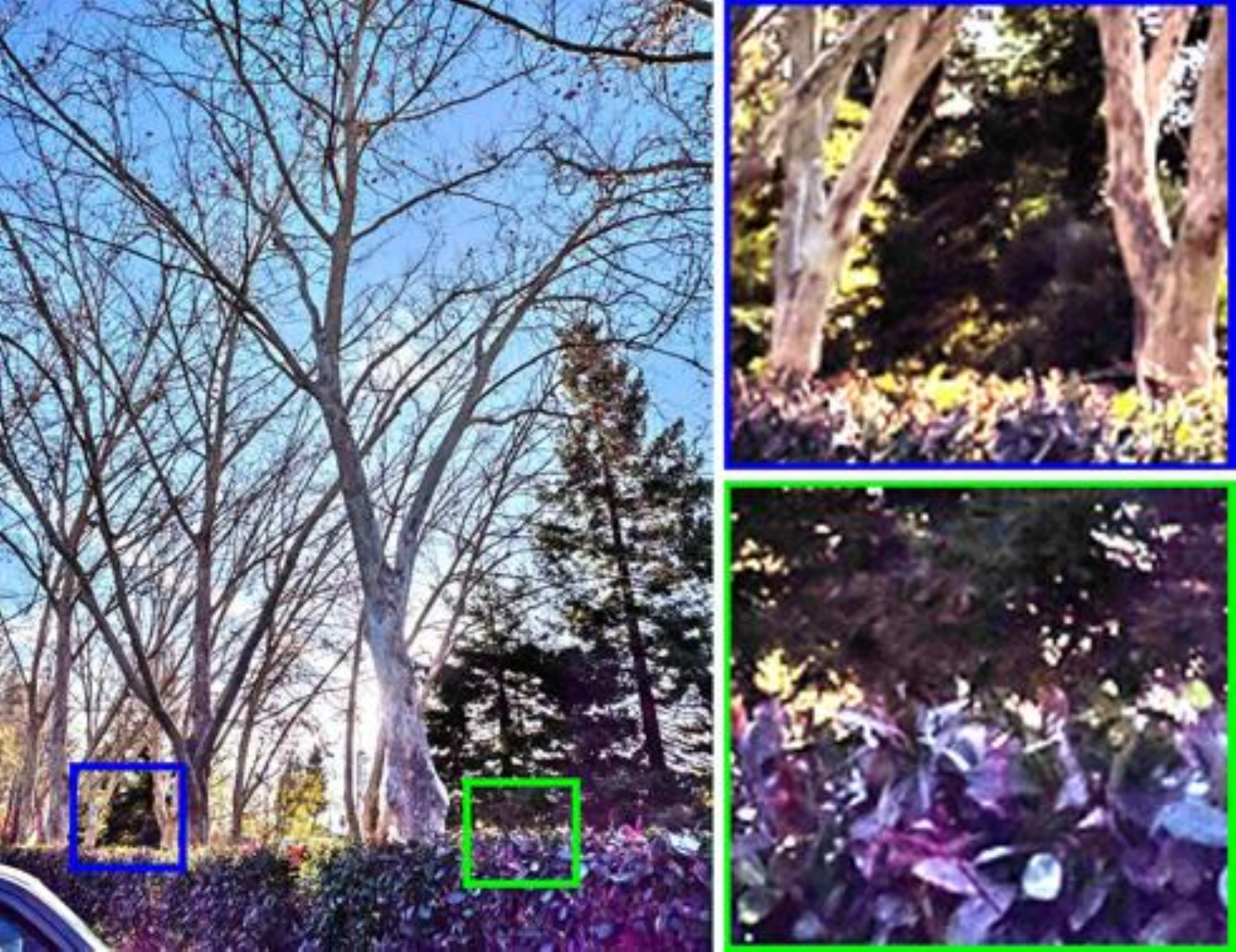}&	
		\includegraphics[width=.32\linewidth]{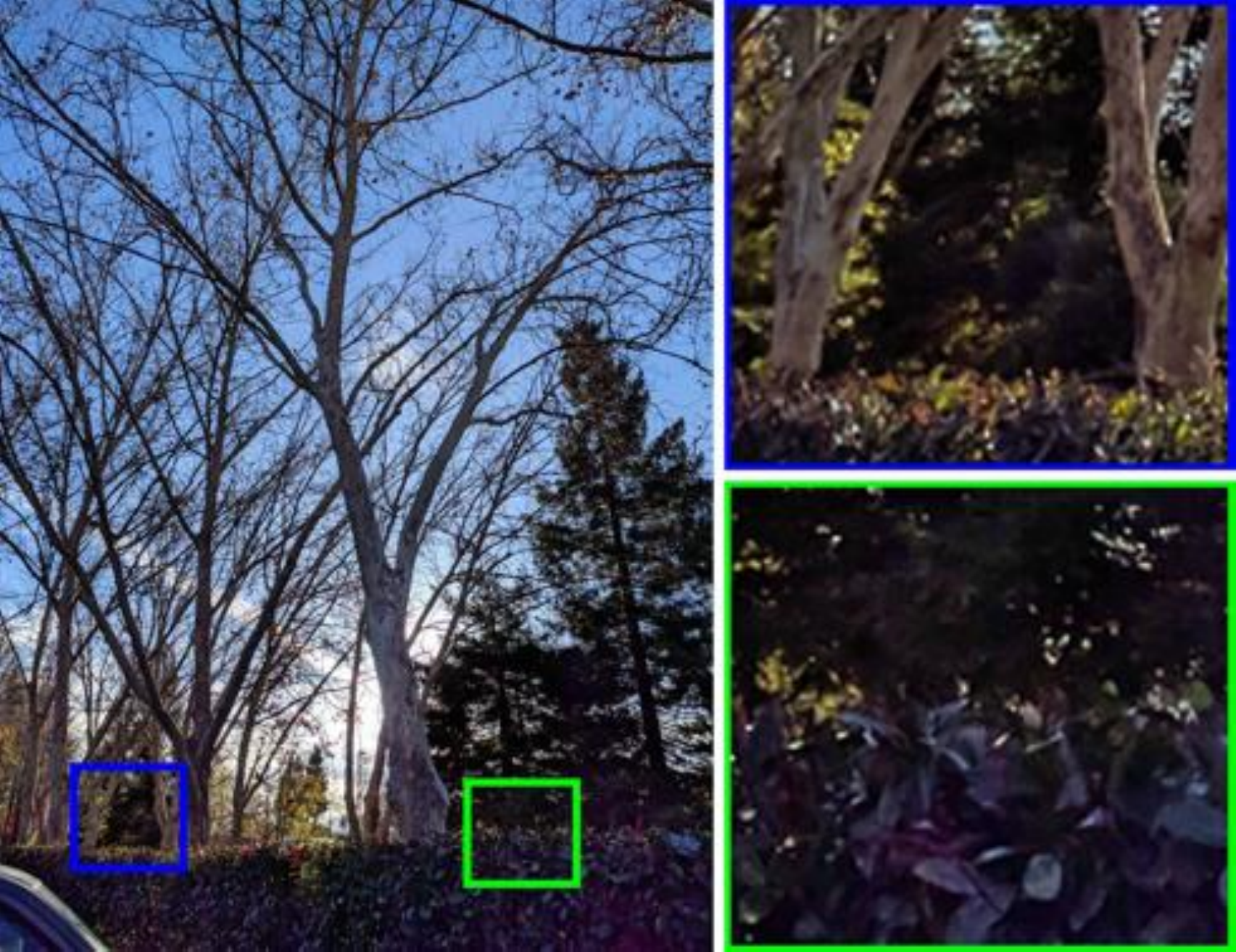}\\
		{ Input}& LIME &{ JIEP}\\
		\includegraphics[width=.32\linewidth]{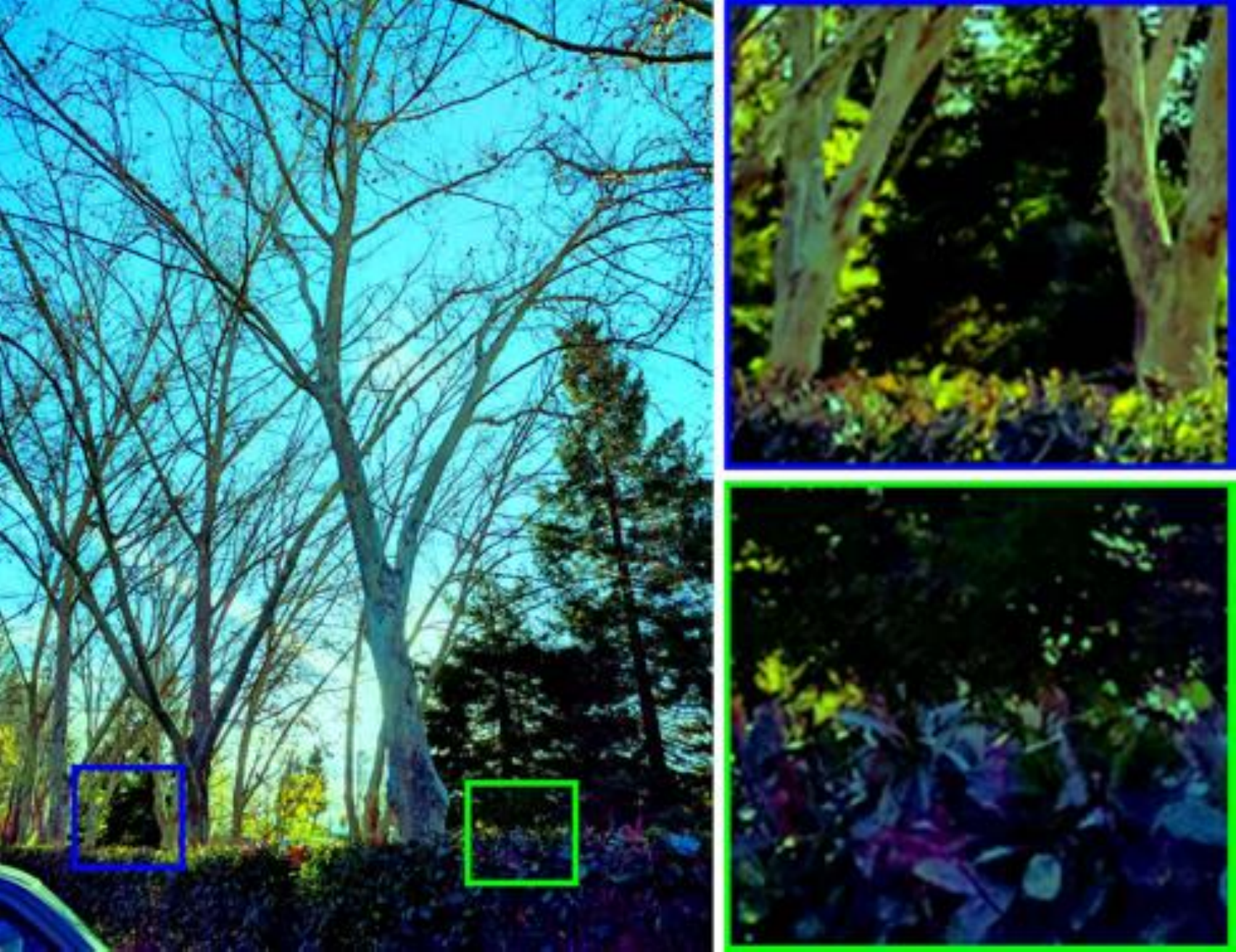}&
		\includegraphics[width=.32\linewidth]{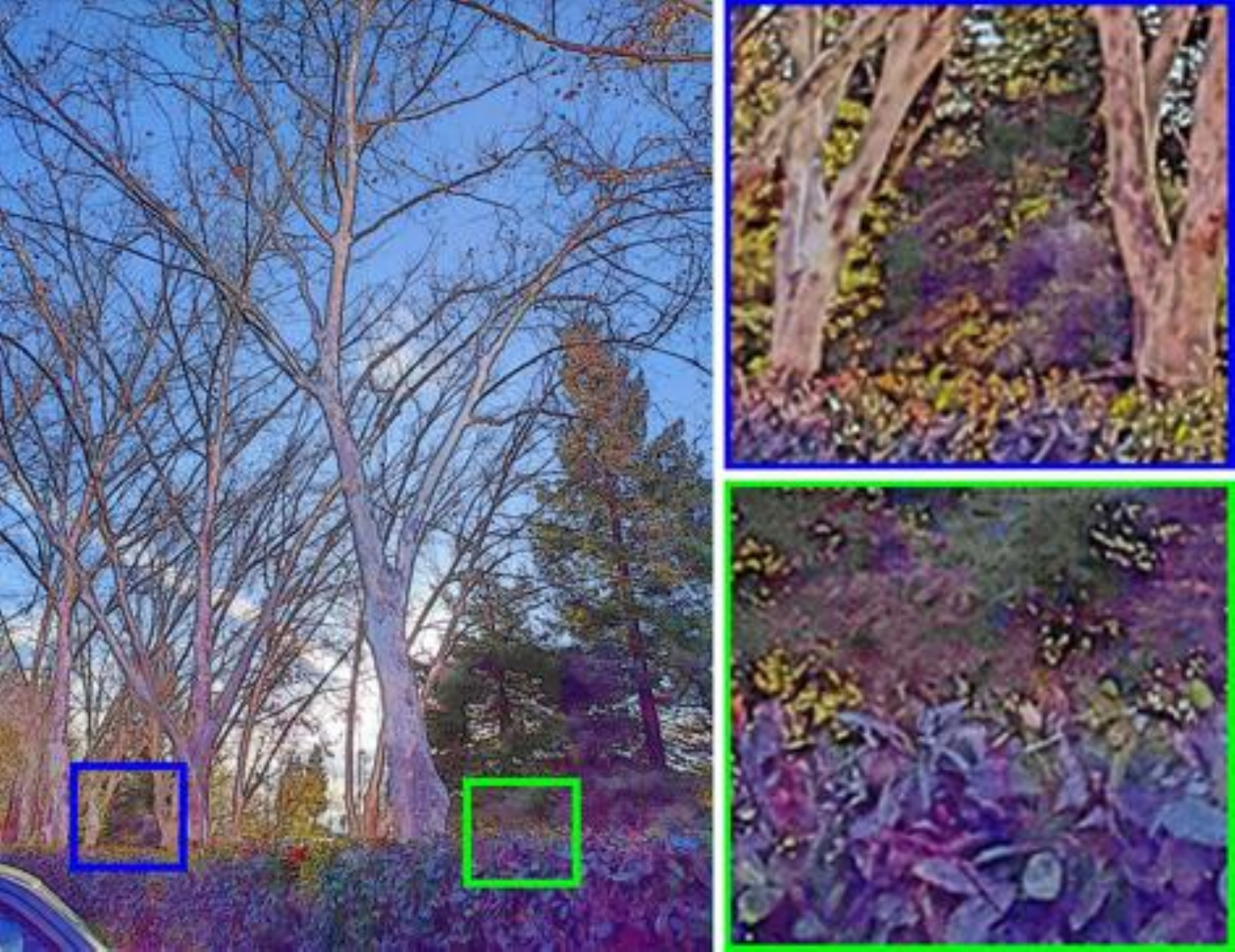}&
		\includegraphics[width=.32\linewidth]{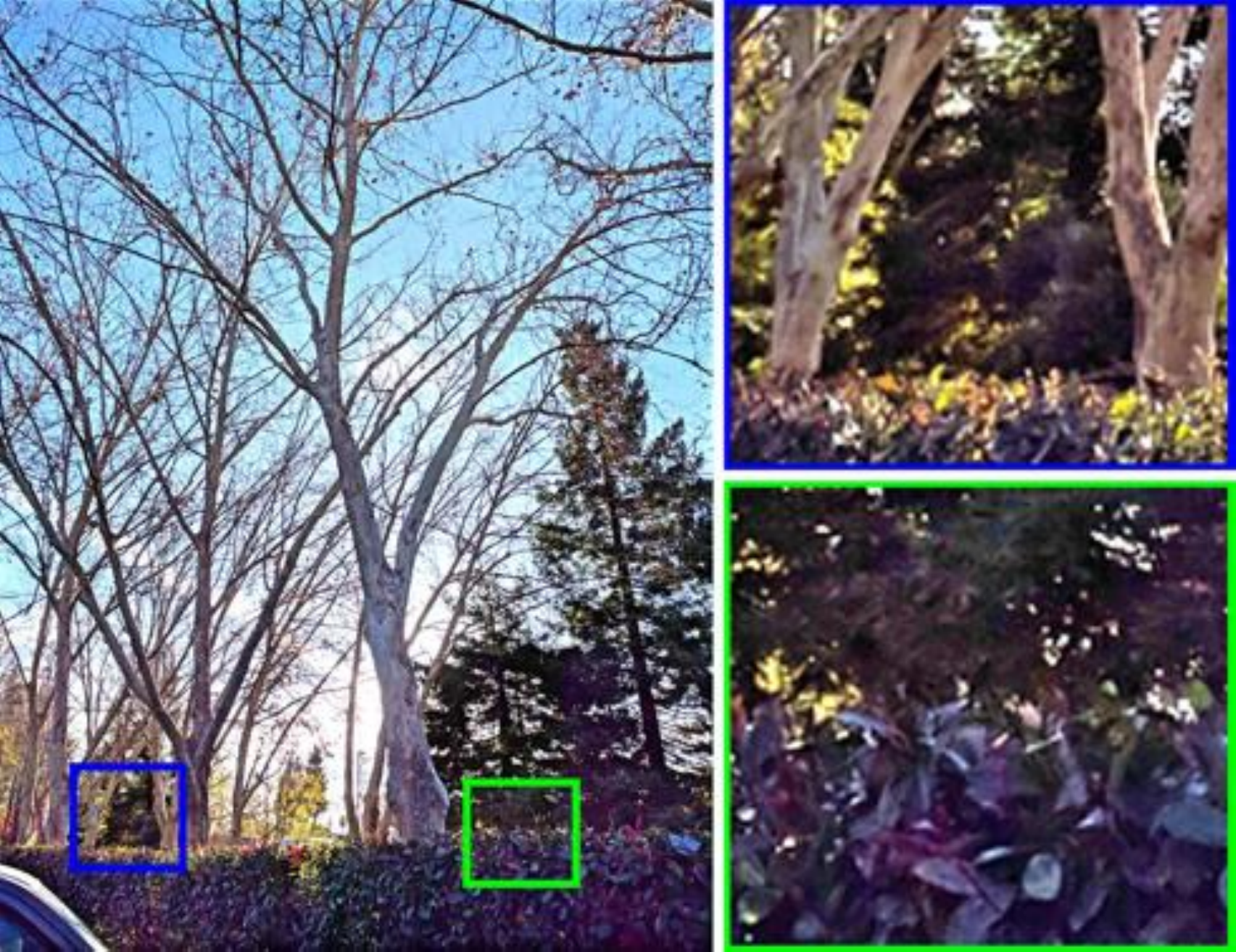}\\
		{ HDRNet} &{ RetinexNet} &{ Ours} \\ 
	\end{tabular}
	\caption{Visual comparisons of underexposed image correction on real-world scenario. }
	\label{fig:mobile}
\end{figure*}

\subsection{Methodology Comparisons} \label{analysis}
We provide a range of experiments to compare the performance of different methodologies on underexposed image correction. 
Then we illustrate the roles of data-driven and knowledge-based submodules in our deep model. 
In Fig.~\ref{fig:decomp}, we compare our proposed method with the state-of-the-art decomposition based enhancement approaches~\cite{fu2015probabilistic,fu2016weighted,cai2017joint} by illustrating the decomposed components and final enhanced results.
Obviously, the illumination of our method is smoother than other state-of-the-art approaches, so that the reflectance and enhanced result preserves most details and thus are clearer than the results of compared methods. All these results verify that our method can obtain more realistic constraints for the Retinex type intrinsic image decomposition and therefore is more suitable for underexposed image correction.

Furthermore, we conduct some experiments to compare with the recent discriminative deep learning approach (e.g., HDRNet~\cite{hasinoff2017Deep}, RetinexNet~\cite{Chen2018Retinex}). Notice that since no physical knowledges is considered in HDRNet, this method can only obtain the enhanced results by learning their network model (designed in heuristic manner) from synthesized training data. RetinexNet considers the Retinex decomposition, but due to the naive generation fashion of training data, i.e., changing exposure time, the predicted enhanced results usually contain too many details and lack the naturalness in real world scenarios.
As shown in Fig.~\ref{fig:VC0}, HDRNet fails to recover more details in the enhanced results, the result of RetinexNet generates more details, but it is extremely unnatural. Our method can successfully recover most of the details in the dark region and keep the naturalness to be most extent. Moreover, our result has a distinguishing promotion in terms of brightness, presenting much higher visibility.

In the third experiment, we explore the performance of our hybrid prior navigated deep propagation (based on the ensemble of two different methodologies). That is, in Fig.~\ref{fig:IlluminationParameters}, we plot visual performances and quantitative results of our method with varied algorithmic parameters $\mu_\mathbf{I}$ and $\lambda_\mathbf{I}$. 
As for $\mu_\mathbf{I}$, it is used to balance the principally designed and data-driven priors. We turn this parameter in the range $[1,20]$ and plot the visual performance and quantitative results in subfigure (a). It can be seen that the performance of our method is stable and the NIQE scores only slightly changed in a small interval (about $10^{-2}$).
While the parameter $\lambda_\mathbf{I}$ is to penalize the auxiliary variable (calculated based on the network propagation). We observe in subfigure (b) that the performances with $\mu_\mathbf{I}\in[1,20]$ are also stable. We argue that the stability of our hybrid prior is mainly because that the knowledge-based submodule actually provides a baseline performance guarantee and the data-driven submodule can successfully enrich more details to further improve the performance.

\subsection{Underexposed Image Correction}
In this part, we first make a series of quantitative and qualitative comparisons with a lot of state-of-the-art approaches for settling the underexposed image correction. Then the experiments in real scenarios are conducted to test the visual performance. Finally face detection based YOLOv3~\cite{redmon2018yolov3} is executed to further verify our naturalness.

\textbf{Challenging Benchmarks}$\;\;$
In this section, we evaluate the performance of our method against state-of-the-art methods, including HE~\cite{cheng2004simple}, MSRCR~\cite{rahman2004retinex}, GOLW~\cite{shan2010globally}, NPEA~\cite{wang2013naturalness}, LIME~\cite{guo2017lime}, SRIE~\cite{fu2015probabilistic}, WVM~\cite{fu2016weighted}, JIEP~\cite{cai2017joint}, HDRNet~\cite{hasinoff2017Deep}, RRM~\cite{li2018structure}, RetinexNet~\cite{Chen2018Retinex} on different benchmarks. Such as NASA~\footnote{{https://dragon.larc.nasa.gov/retinex/pao/news/}} (23 images in the indoor and outdoor scenes), NPE~\cite{wang2013naturalness} (130 images in different natural scenes ), LIME~\cite{guo2017lime} (10 images in different challenging scenes).

Fig.~\ref{fig:rescomp} reports the averaged NIQE values on different datasets. It is obvious that our method obtains better quantitative performance than other state-of-the-art methods. We also plot visual comparisons on example images in Figs.~\ref{fig:nonuniform}-\ref{fig:lime}. It can be seen that most methods can partially improve the visual quality of the given observations. However, the results of SRIE, WVM and JIEP still express low-visibility, especially on the most challenging example in Fig.~\ref{fig:lime}. Although with improved contrast, LIME tends to obtain images with severe over-exposure. Additionally, all these compared methods all fail to recover the detailed information in the dark, such as the rose flower in the second zoomed-in region of Fig.~\ref{fig:lime}.  Learning-based approaches (i.e., HDRNet, RetinexNet) generate unrealistic results with color distortion, especially the result of RetinexNet.
In contrast, our proposed method not only enhances the visibility, but also preserves most of the details, providing much better enhancement performance.

We also compare our method with four recently proposed methods to evaluate the computational cost. Table.~\ref{tab:times} shows the average running time in seconds among on different benchmarks. It can be seen that our method is the fastest among these compared methods except the NASA dataset. This indicates our method has the significant advantage in terms of time cost.
\begin{figure*}
	\centering
	\begin{tabular}{c@{\extracolsep{0.3em}}c@{\extracolsep{0.2em}}c@{\extracolsep{0.2em}}c@{\extracolsep{0.2em}}c@{\extracolsep{0.2em}}c@{\extracolsep{0.2em}}c}
		\includegraphics[width=0.135\textwidth]{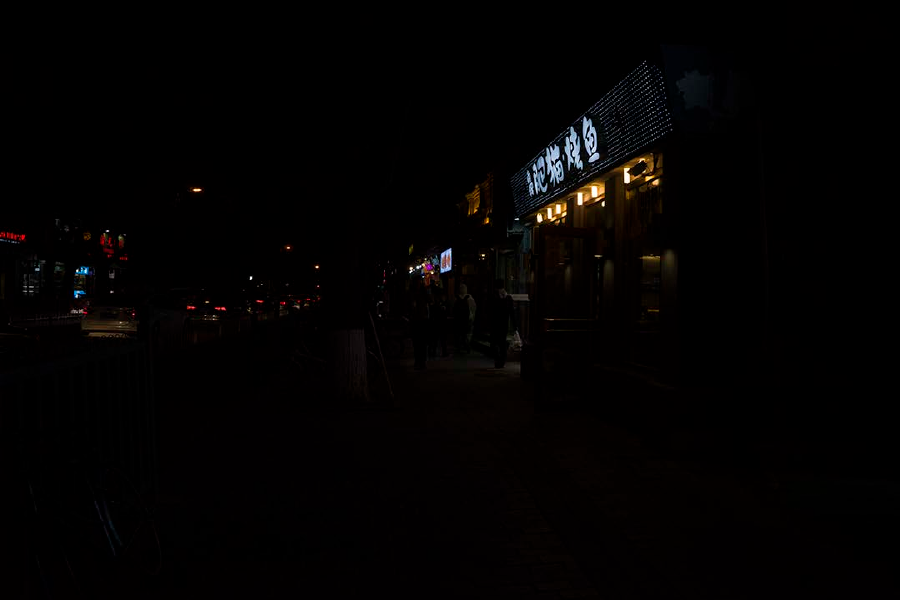}&
		\includegraphics[width=0.135\textwidth]{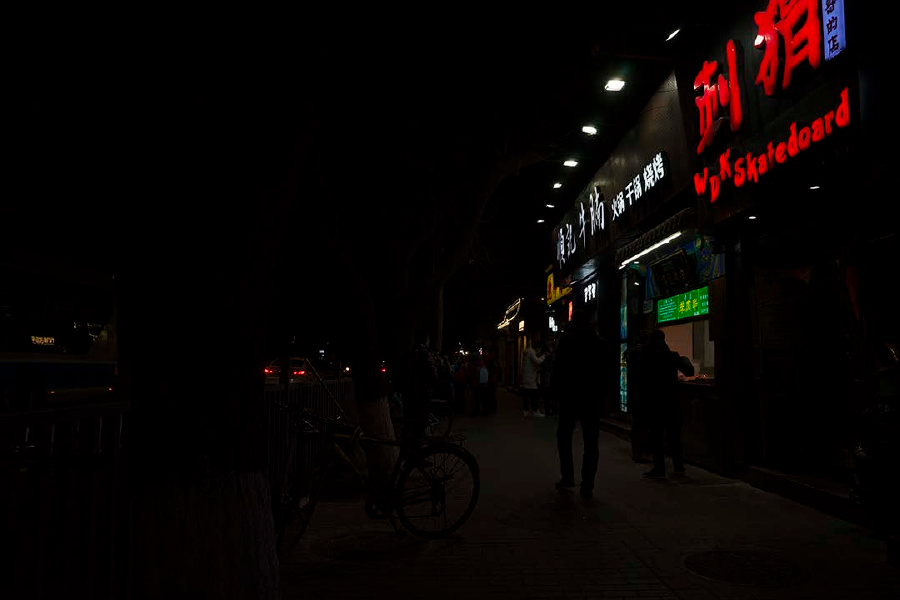}&
		\includegraphics[width=0.135\textwidth]{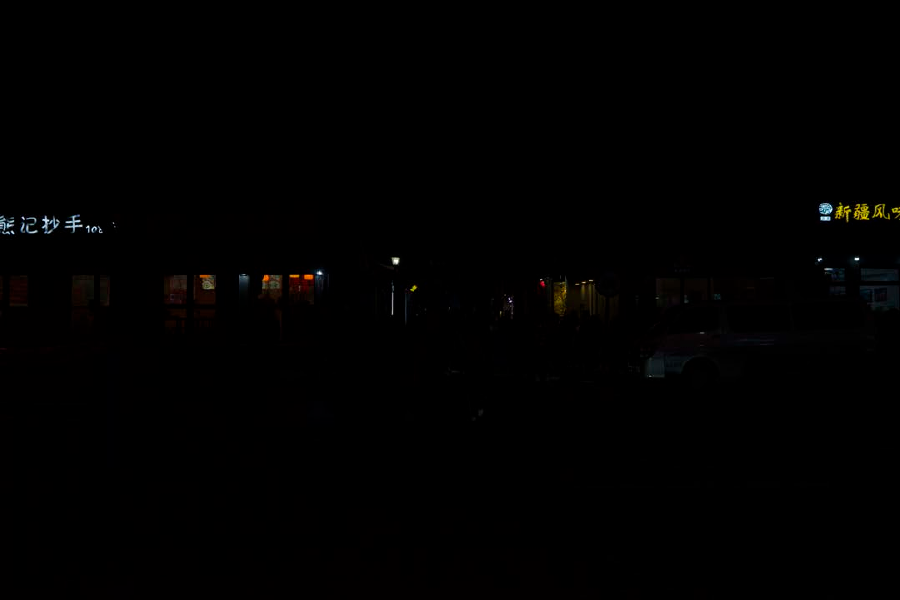}&
		\includegraphics[width=0.135\textwidth]{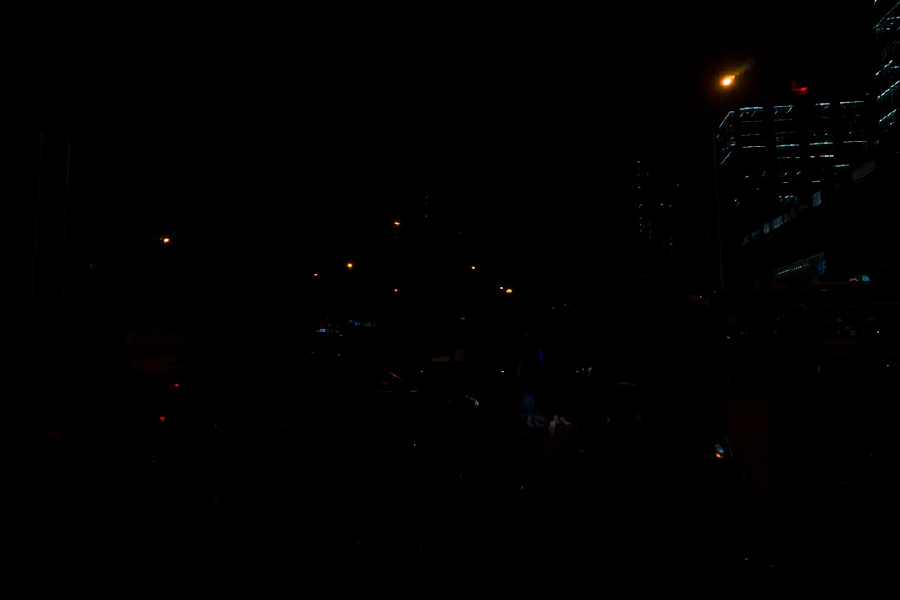}&
		\includegraphics[width=0.135\textwidth]{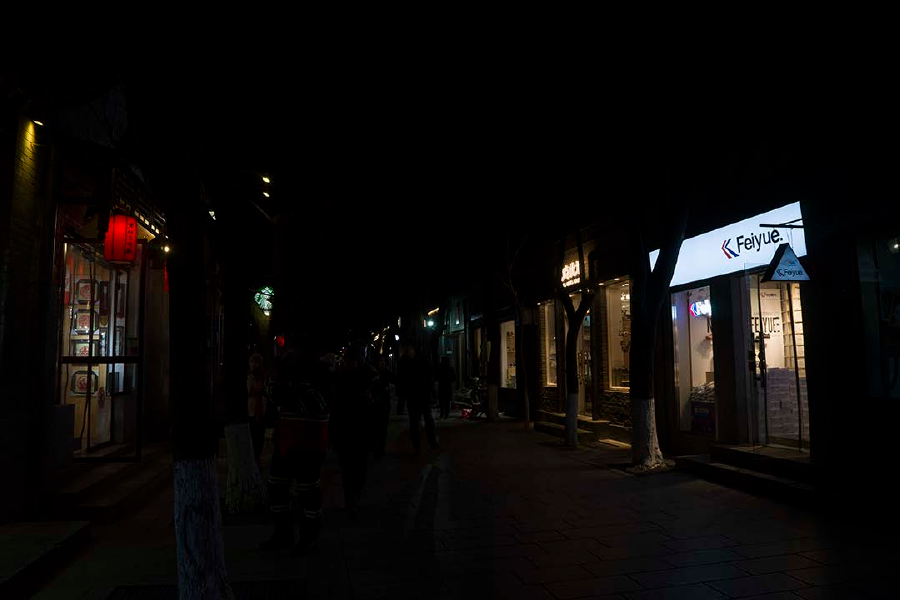}&
		\includegraphics[width=0.135\textwidth]{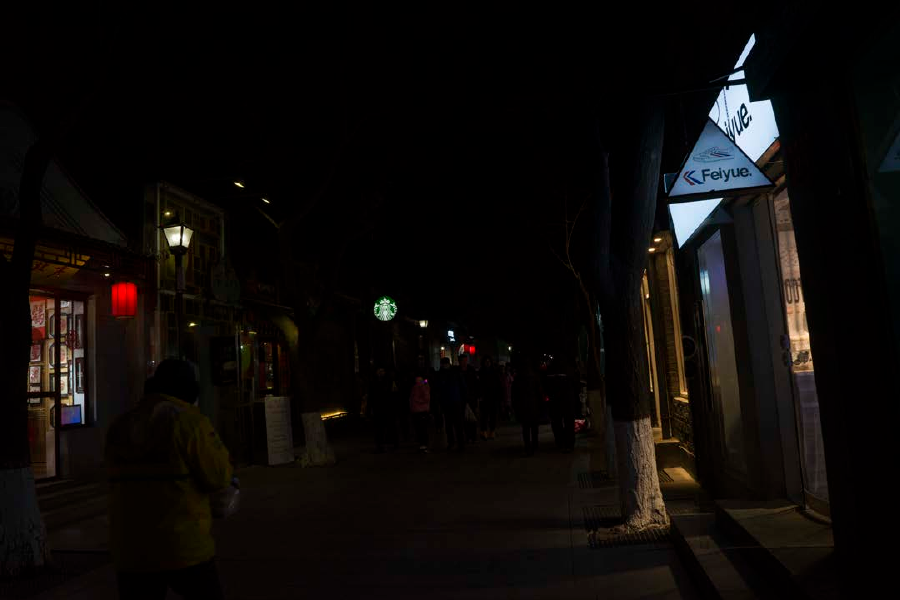}&
		\includegraphics[width=0.135\textwidth]{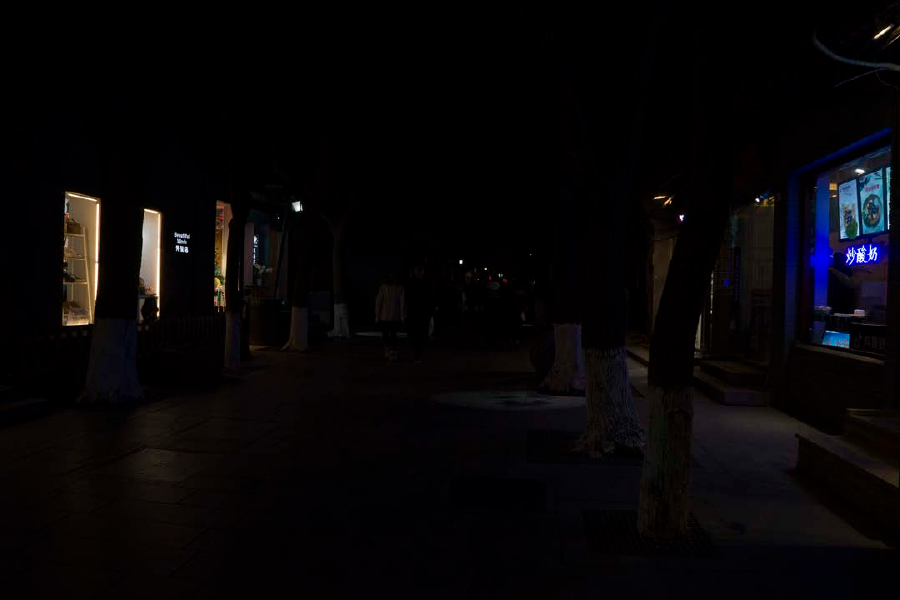}\\
		\includegraphics[width=0.135\textwidth]{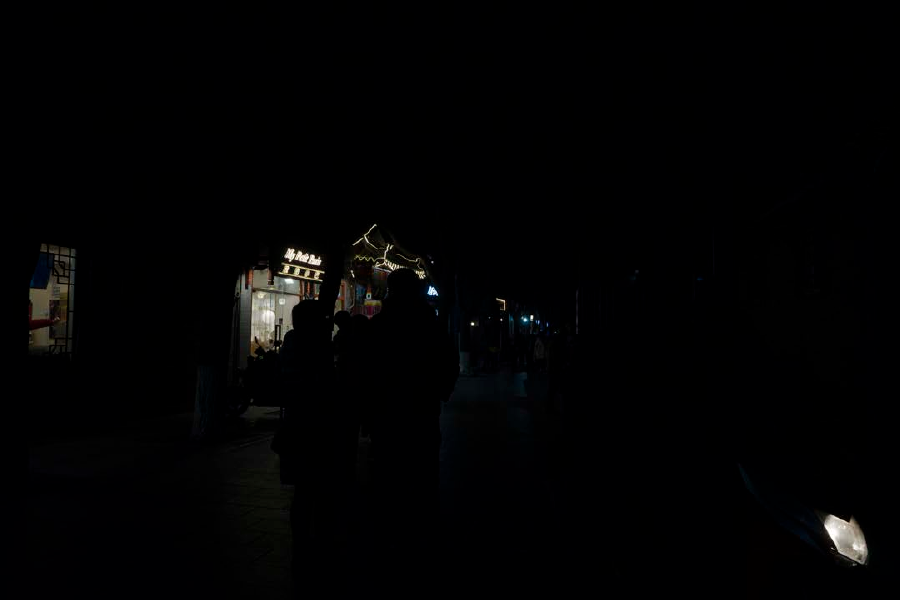}&
		\includegraphics[width=0.135\textwidth]{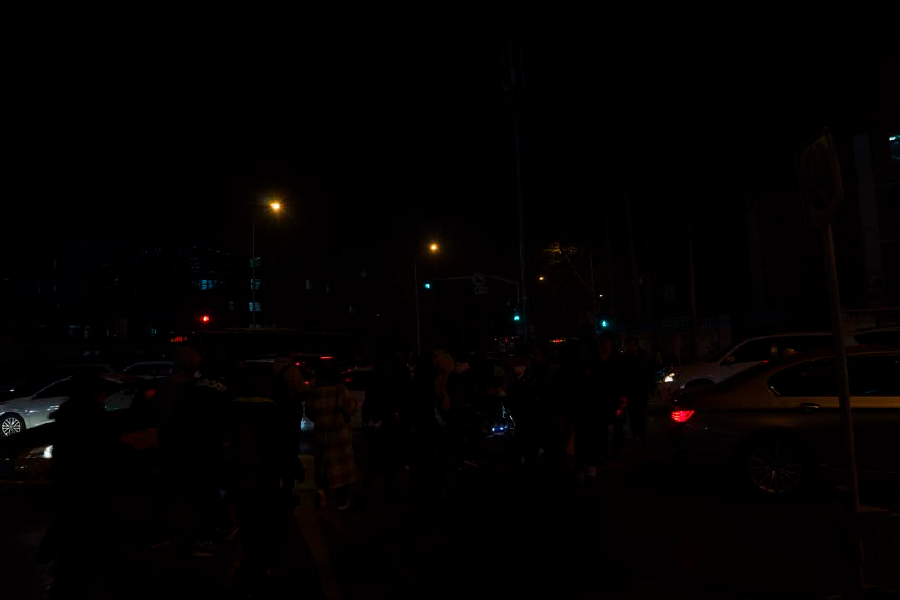}&
		\includegraphics[width=0.135\textwidth]{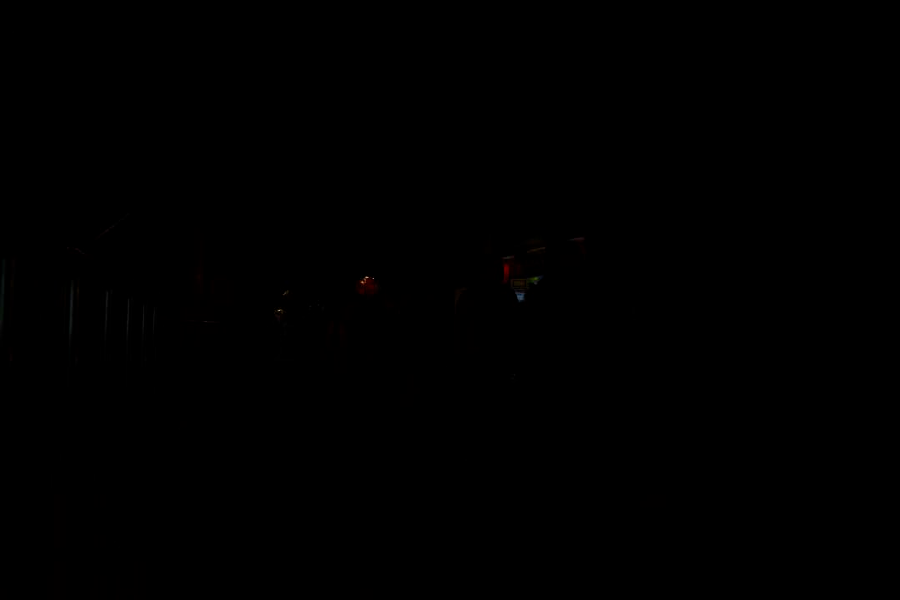}&
		\includegraphics[width=0.135\textwidth]{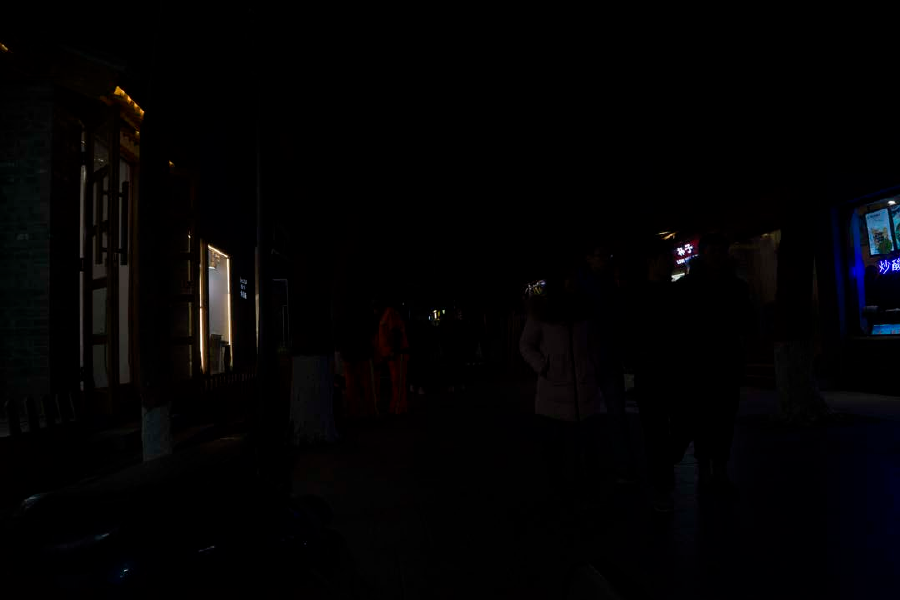}&
		\includegraphics[width=0.135\textwidth]{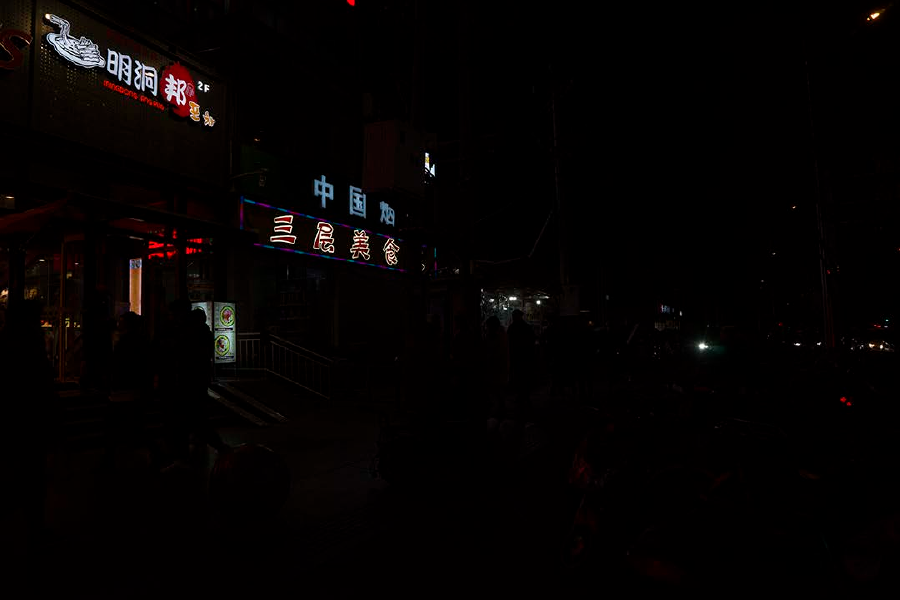}&
		\includegraphics[width=0.135\textwidth]{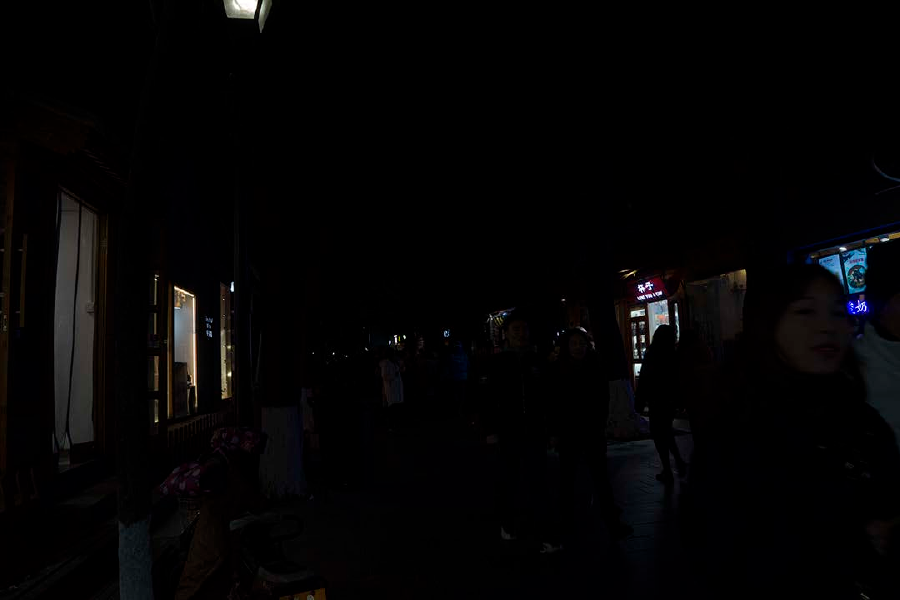}&
		\includegraphics[width=0.135\textwidth]{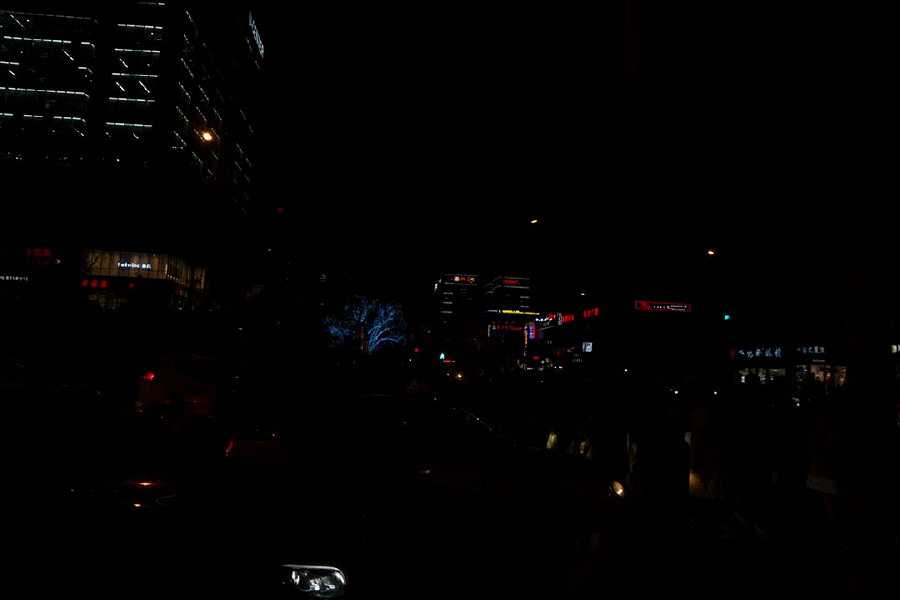}\\
	\end{tabular}
	\caption{Some sample images selected from DARK FACE dataset.}
	\label{fig:DarkFace}
\end{figure*}

\begin{figure*}
	\centering
	\begin{tabular}{c@{\extracolsep{0.5em}}c@{\extracolsep{0.5em}}c}
		\includegraphics[width=0.32\textwidth]{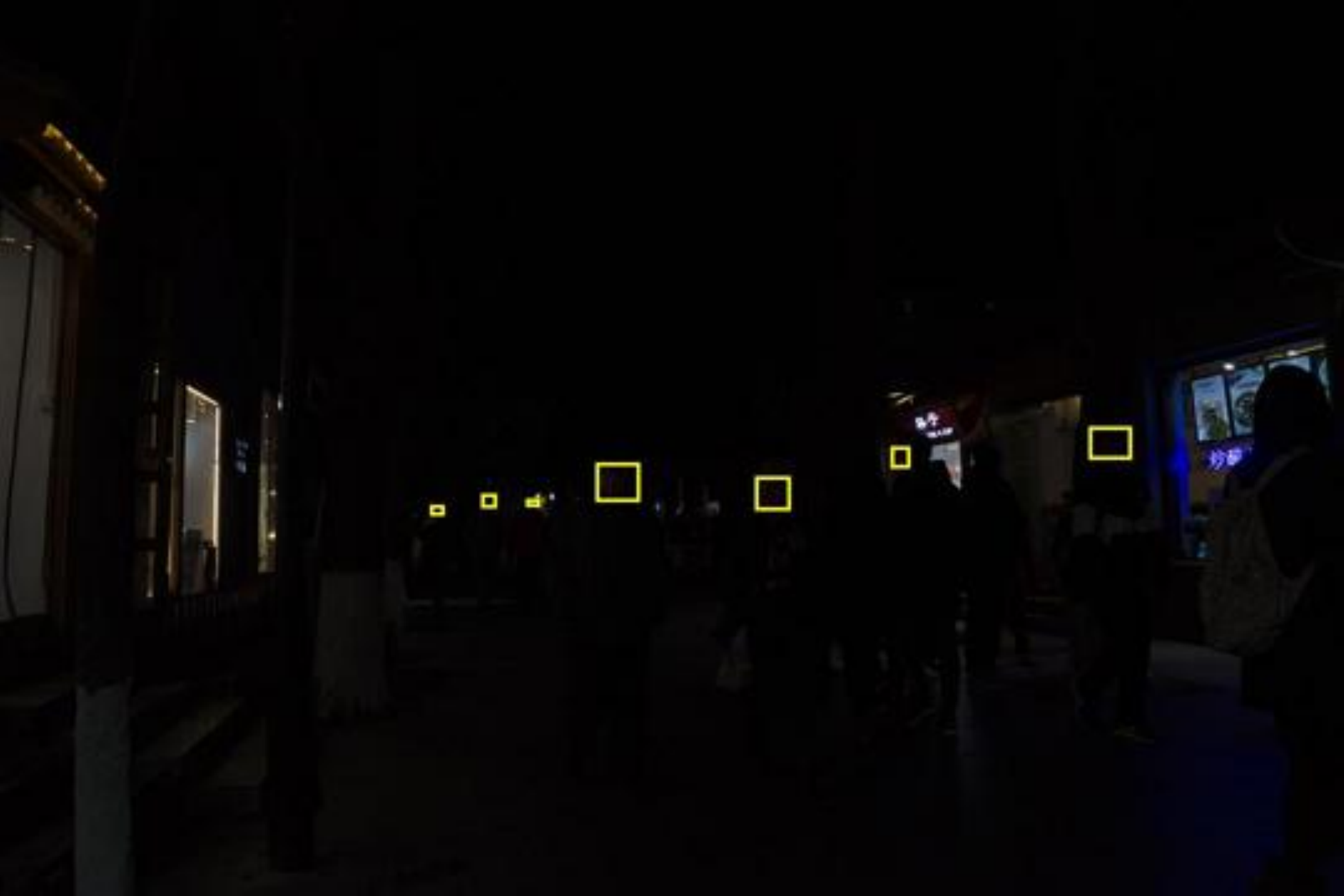}&
		\includegraphics[width=0.32\textwidth]{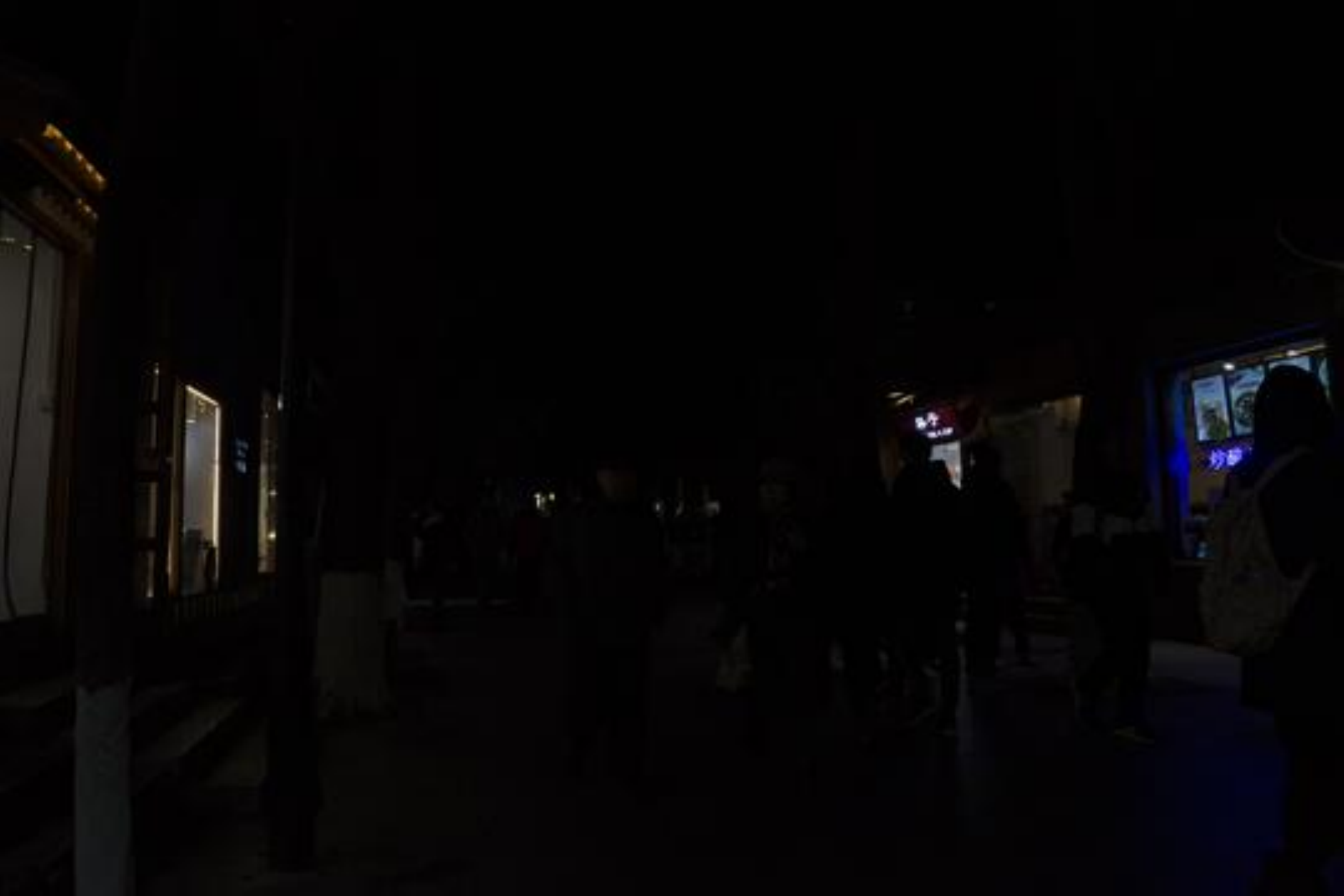}&
		\includegraphics[width=0.32\textwidth]{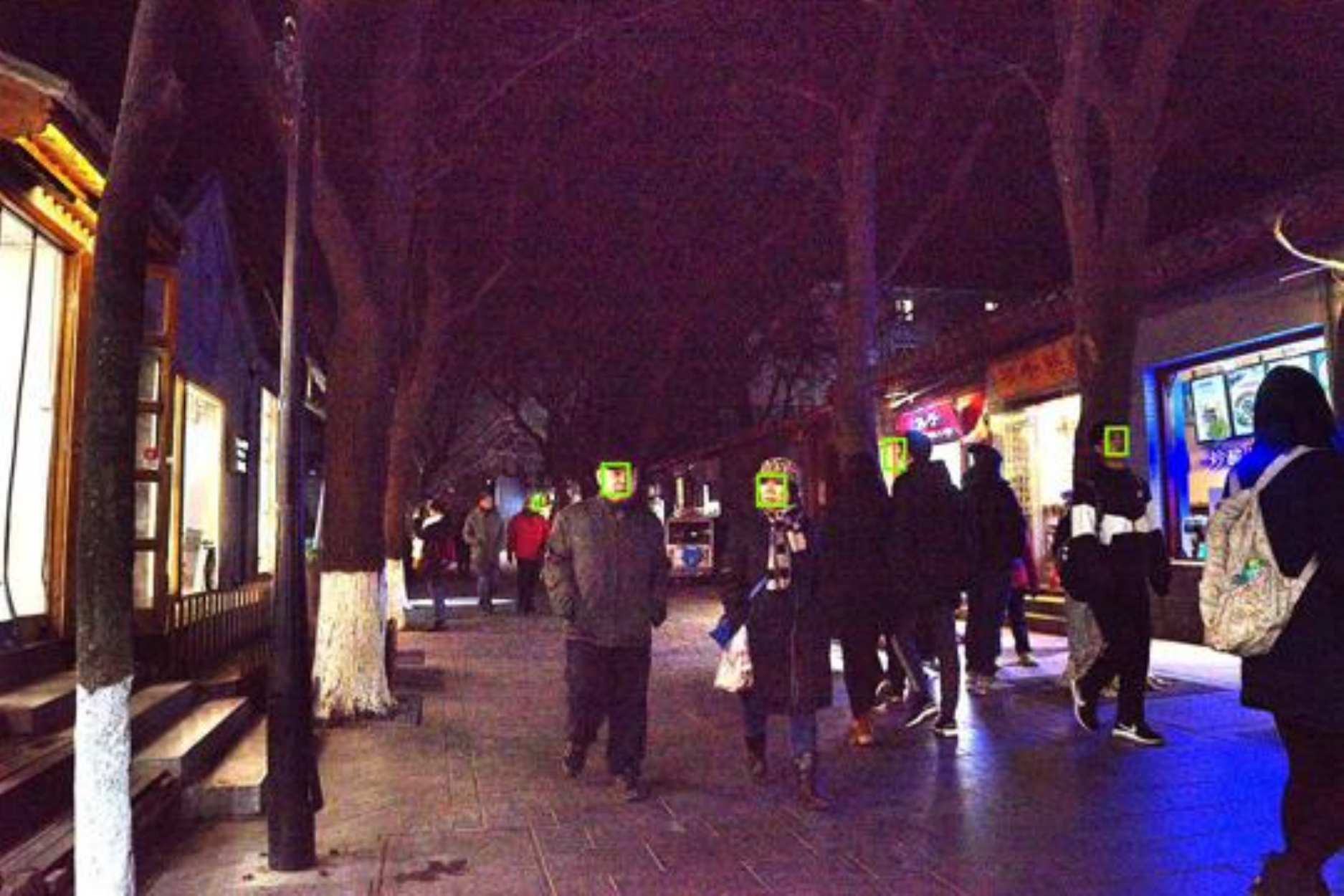}\\
		Underexposed image with label&Underexposed input (0 / 0)&LIME~\cite{guo2017lime} (60.00 / 42.86)\\
		\includegraphics[width=0.32\textwidth]{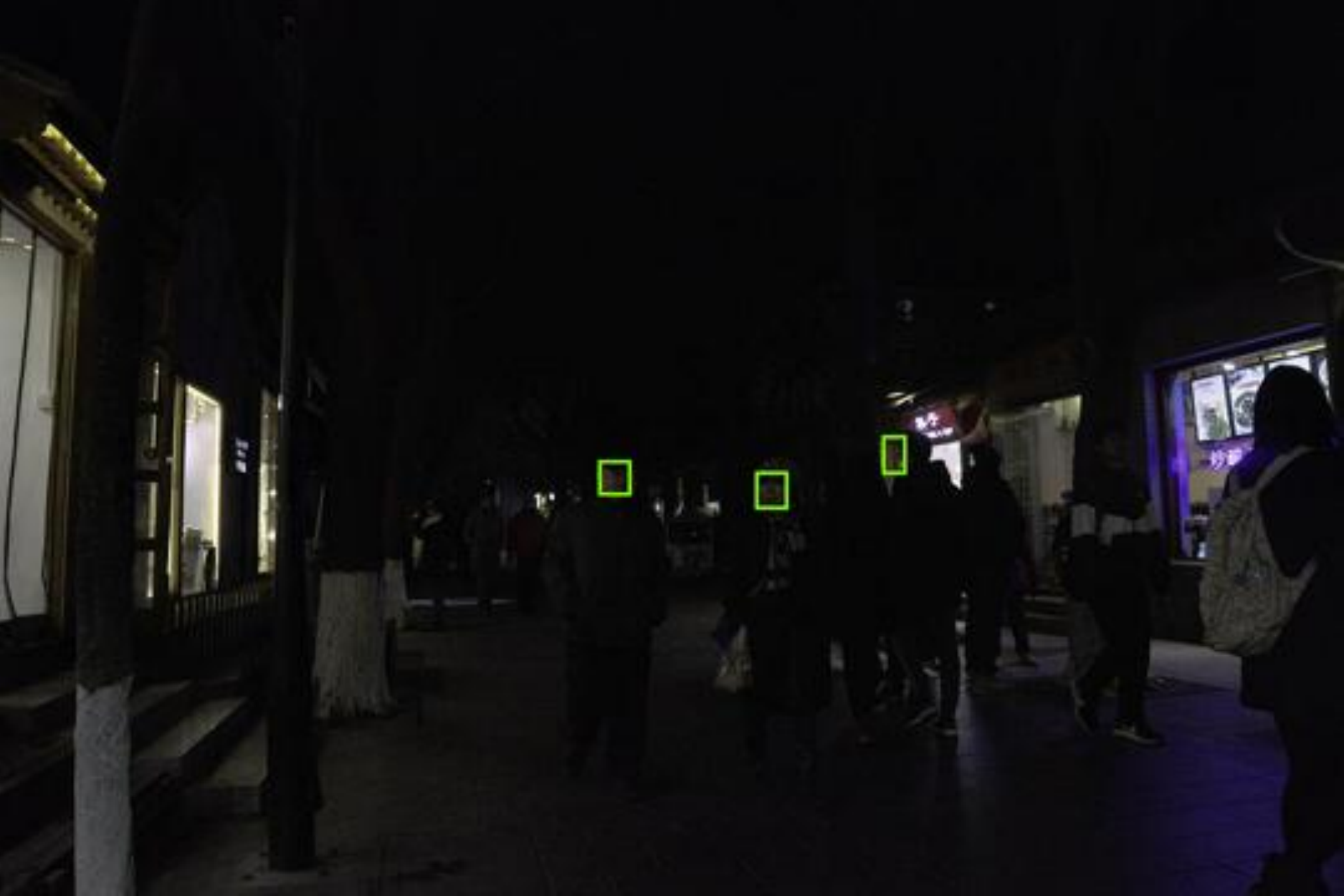}&
		\includegraphics[width=0.32\textwidth]{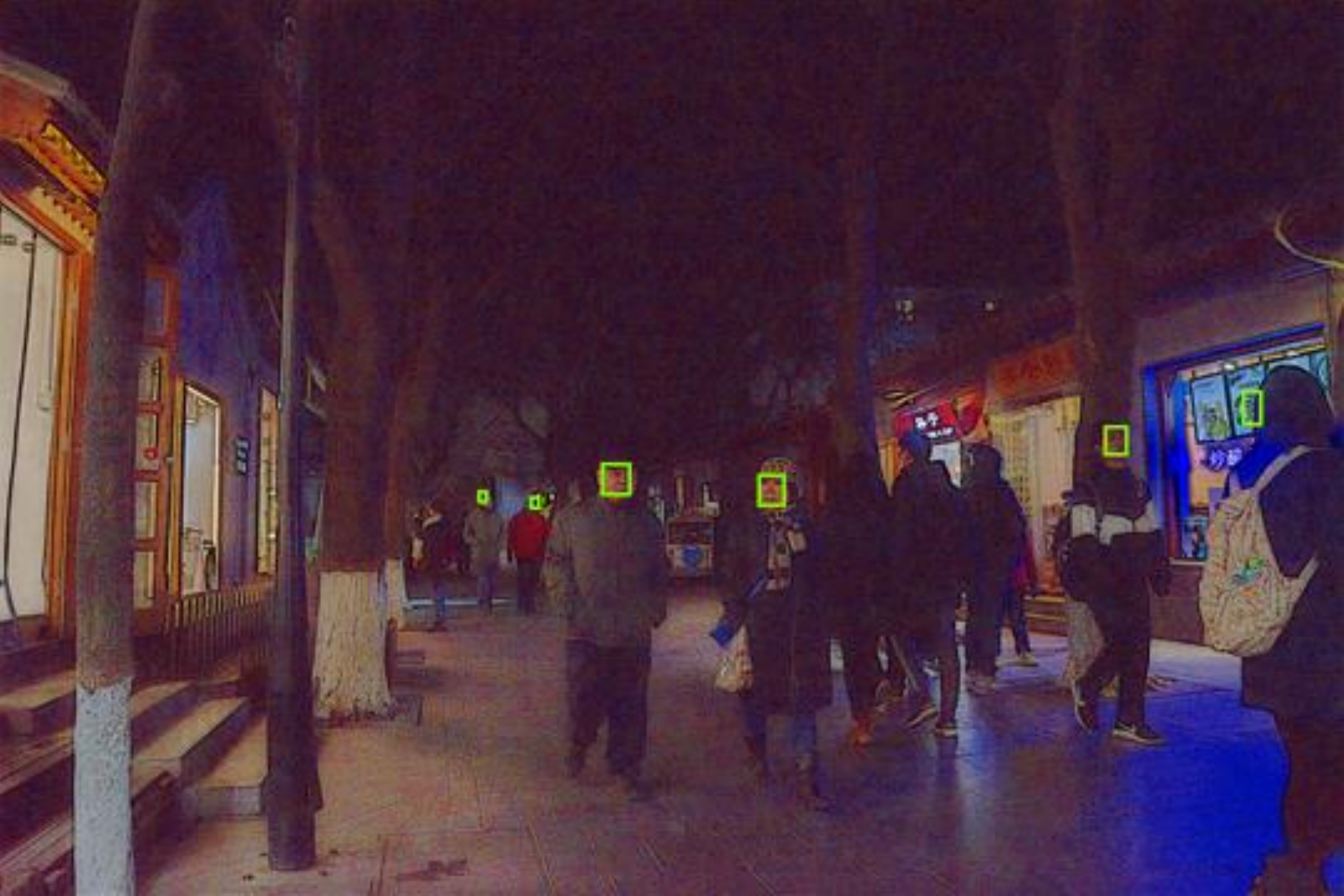}&
		\includegraphics[width=0.32\textwidth]{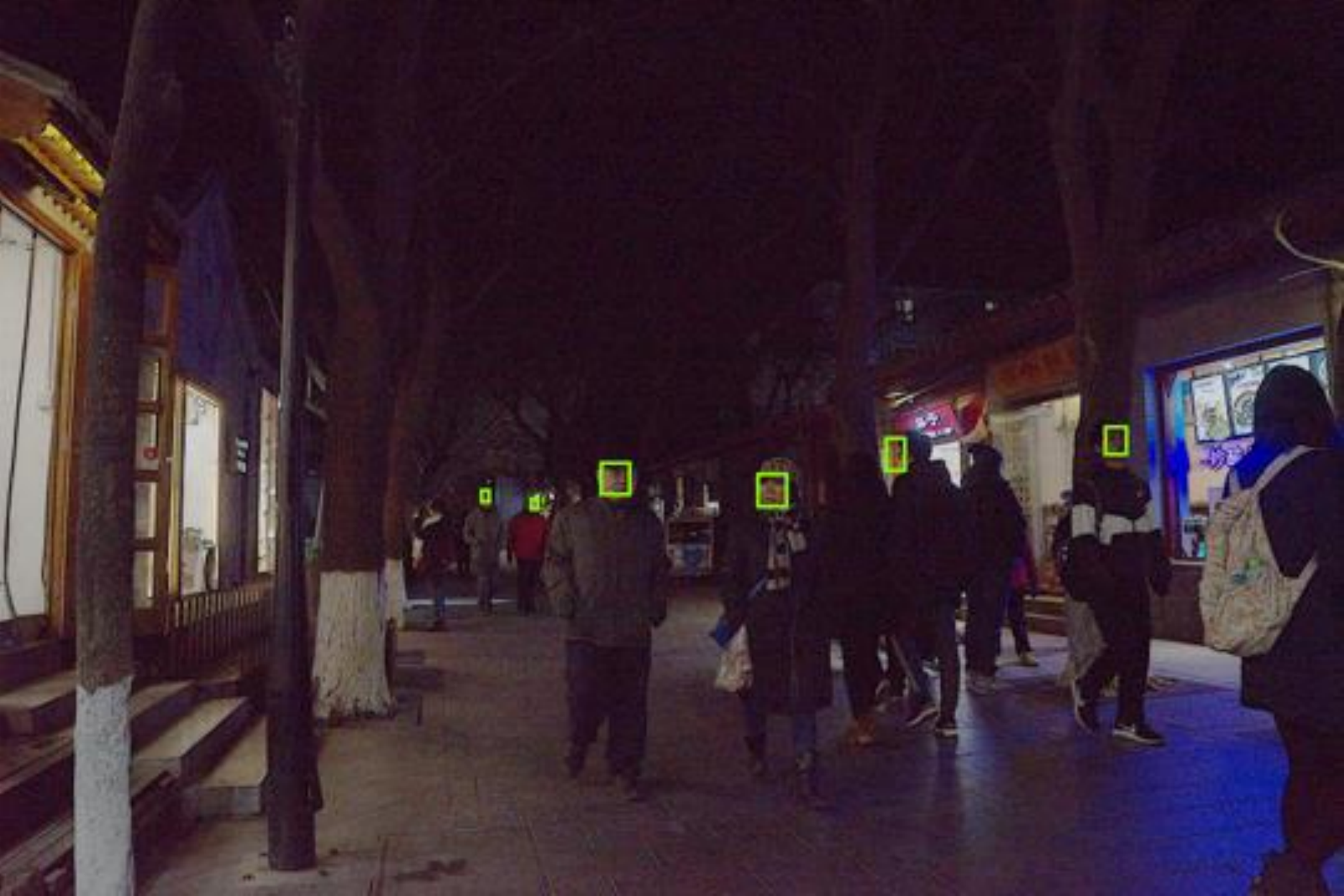}\\
		HDRNet~\cite{hasinoff2017Deep} (1.00 / 42.86)&RetinexNet~\cite{Chen2018Retinex} (66.67 / 57.14)&Ours ({83.33 / \textbf{71.43}})\\
	\end{tabular}
	\caption{Comparison of face detection based on YOLOv3~\cite{redmon2018yolov3} among LIME~\cite{guo2017lime} (representative Retinex-based method), HDRNet~\cite{hasinoff2017Deep} (end-to-end network), RetinexNet~\cite{Chen2018Retinex} (Retinex-based end-to-end network)  and our method. The Precision / Recall scores are reported below each image.
		Clearly, our method can detect almost all objects with the better quantitative performance. This experiment not only manifest the naturalness of our result, but also reflect the prospects of our proposed method in the real vision application.}
	\label{fig:detection}
\end{figure*}

\begin{table}[t]
	\centering
	\caption{Average running time on different benchmarks. The best and second are highlighted in \textcolor{red}{red} and \textcolor{blue}{blue} color, respectively.}
	\begin{tabular}{cccccc}
		\toprule
		Dataset& LIME& JIEP& HDRNet& RetinexNet& Ours\\ 
		\midrule
		NASA& \textcolor{red}{0.0336}& 0.9858& 4.8329& 0.1618& \textcolor{blue}{0.0477}\\
		NPE& \textcolor{blue}{0.1902}& 5.0584& 13.6522& 0.2948& \textcolor{red}{0.1112}\\
		LIME& \textcolor{blue}{0.2149}& 2.5536& 14.9201& 0.3356& \textcolor{red}{0.1758}\\
		\bottomrule
	\end{tabular}
	\label{tab:times}
\end{table}

\textbf{Real-world Scenarios}$\;\;$
We also evaluate our method on real-world underexposed scenarios. We select an example image from HDR+ Burst Photography Dataset~\cite{hasinoff2016burst}, which is captured by the Android mobile cameras using the public Android Camera2 API. As Fig.~\ref{fig:mobile} shows, 
it can be seen that our method, LIME, and RetinexNet all have better performance than other state-of-the-art methods in the dark regions. However, the zoomed-in regions of LIME are over-exposed and contain color distortion, RetinexNet generates the unnatural enhanced result which looks like style migration. In contrast, our proposed method obtains more natural visual quality with clear details on the test image.

\textbf{Face Detection Based on YOLOv3}$\;\;$
We know that the naturalness of enhanced results is not precise enough to illustrate by NIQE value derived from the statistic regular. Indeed, Visual expression of enhanced results further supports the naturalness, but it is over-subjective because of personal preference. To address these problems, we consider evaluating the naturalness property from the perspective of the performance of the face detection task.

\begin{table}[t]
	\centering
	\caption{Quantitative comparison of face detection.}
	\begin{tabular}{cccccc}
		\toprule
		Metric& Input&LIME& HDRNet& RetinexNet& Ours\\ 
		\midrule
		mAP (\%) &16.77&53.49&34.25&37.10&\textbf{54.28}\\
		Avg. Recall (\%) &12.67&74.17&35.67&42.99&\textbf{78.43}\\
		\bottomrule
	\end{tabular}
	\label{tab:Facedetection}
\end{table}

To be specific, we adopt a well-known object detection framework, i.e., YOLOv3~\cite{redmon2018yolov3} to present the task of face detection. Following most existing face detection works, we use WIDER Face dataset~\cite{yang2016wider} as our training data. It needs to be noticed that the illumination is also considered in this dataset, but those images are easy to recognize the objects by our eyes. To fully verify the capability of underexposed image correction algorithms, we select 100 challenging images from DARK FACE dataset~\footnote{\url{https://flyywh.github.io/CVPRW2019LowLight/}} which comes from the sub-challenge of UG2$+$ PRIZE CHALLENGE held at CVPR 2019. We select some images from our built dataset to present the difficulty of recognition and detection as Fig.~\ref{fig:DarkFace} shows.

As Table.~\ref{tab:Facedetection} shows, our method achieves the best quantitative performance in all metrics (i.e., mAP, Average Recall) against other state-of-the-art methods. Specifically, end-to-end network based methods (i.e., HDRNet, RetinexNet) harvest the worst quantitative performance. It is worth noting that RetinexNet is superior to HDRNet, the reason may be that RetinexNet adopts the Retinex decomposition to achieve a more favorable performance for detection. The representative Retinex-based method (i.e., LIME) indeed presents the fine numerical results, which only consider the designed prior driven by knowledges. 
In contrast, the average recall of our method is higher than the LIME about four percentage points, which reflects our method can detect more objects. Actually, the detection network trained by lots of natural images needs the input which satisfies the distribution of natural images, to achieve more excellent performance. In this view, our method indeed performs more effective naturalness. Actually, our numerical results are not objectively prominent for face detection task, whose cause may be that many noises and artifacts are produced in the enhanced procedure to influence the detection (as Fig.~\ref{fig:detection} shows). We will consider the procedure of noises removal to further improve the enhanced performance in our future work.

\begin{figure*}[!htb]
	\centering
	\begin{tabular}{c@{\extracolsep{0.28em}}c@{\extracolsep{0.28em}}c@{\extracolsep{0.28em}}c@{\extracolsep{0.28em}}c}
		\includegraphics[width=0.192\linewidth]{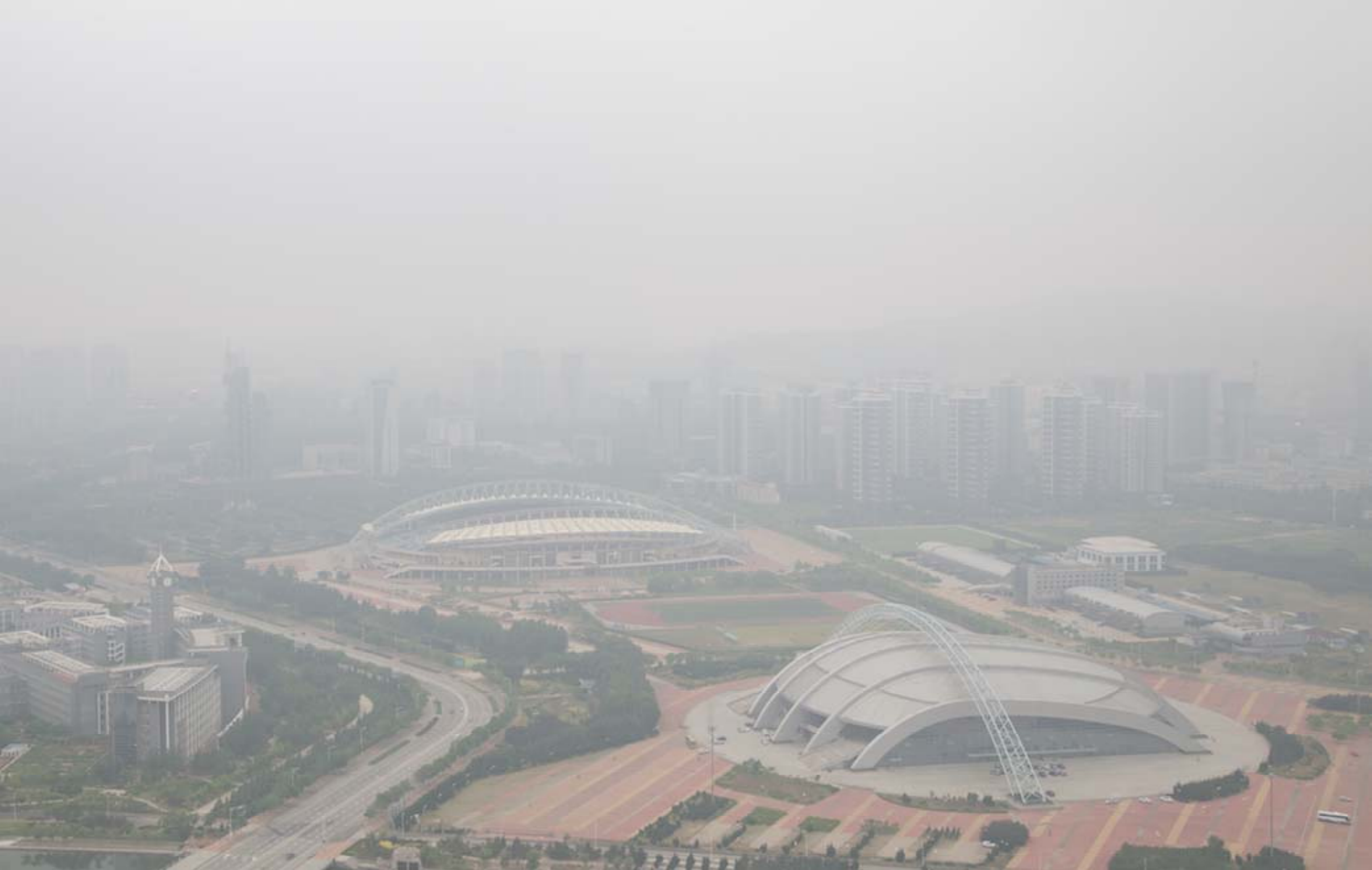}&
		\includegraphics[width=0.192\linewidth]{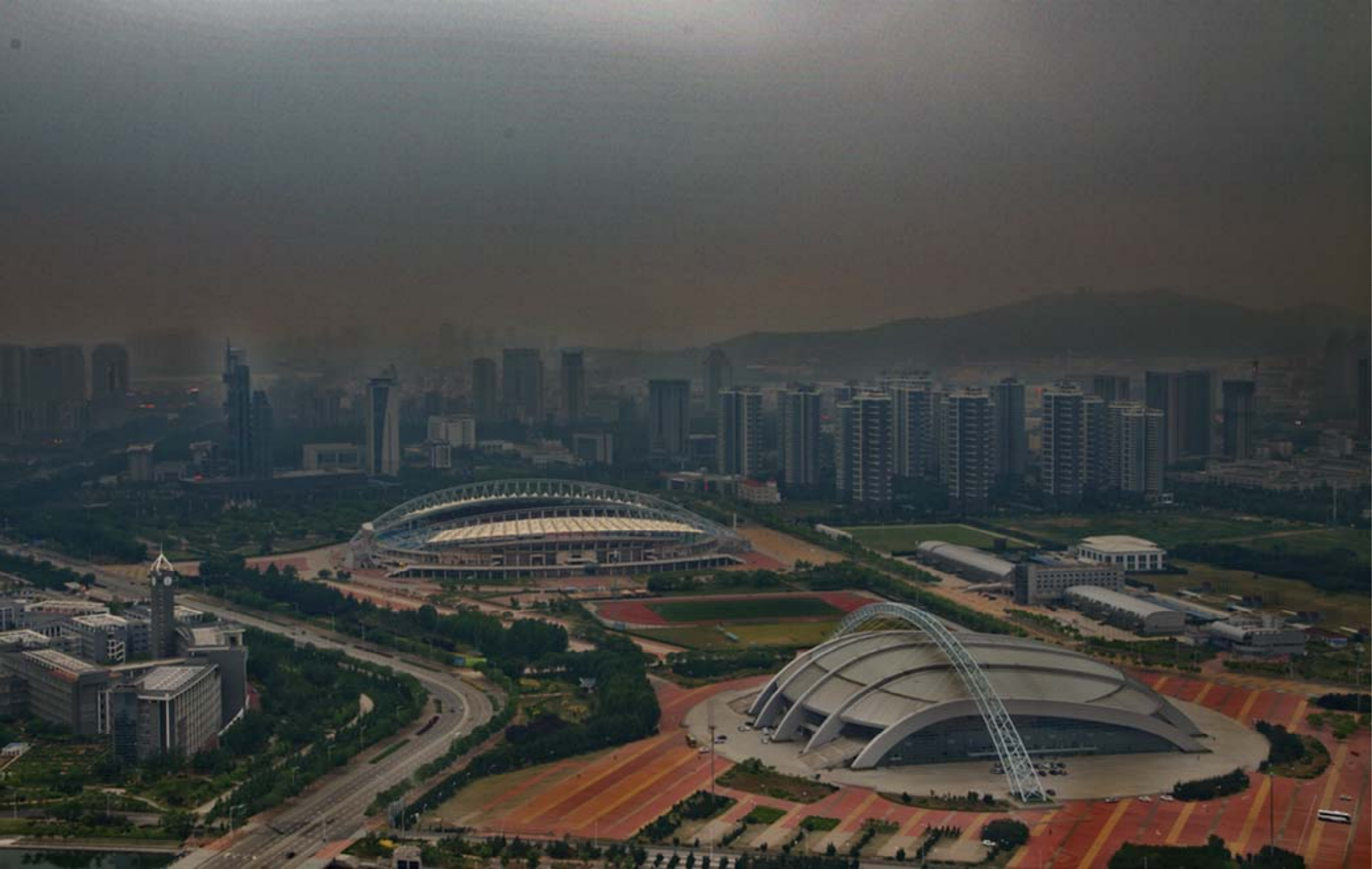}&
		\includegraphics[width=0.192\linewidth]{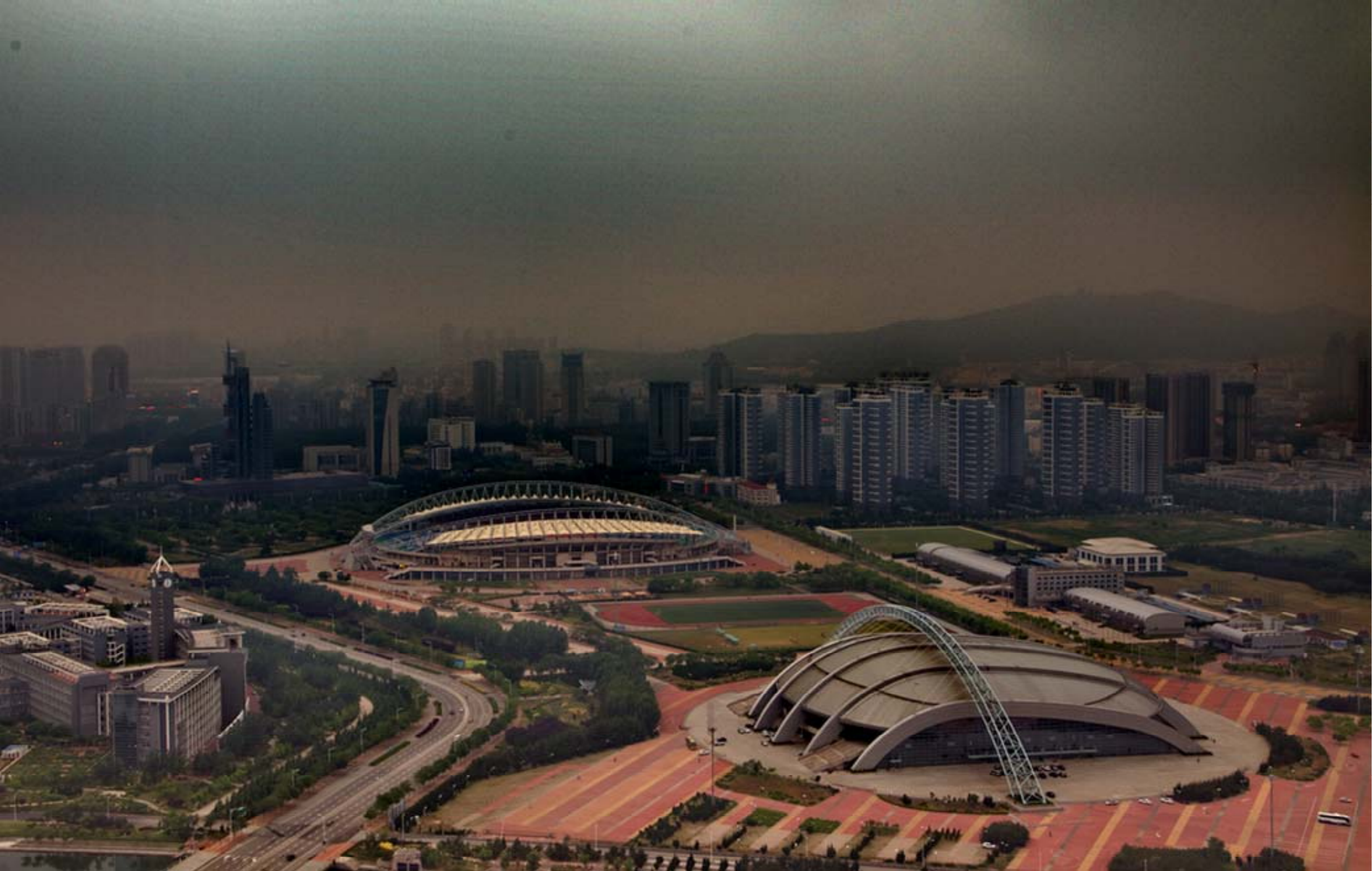}&
		\includegraphics[width=0.192\linewidth]{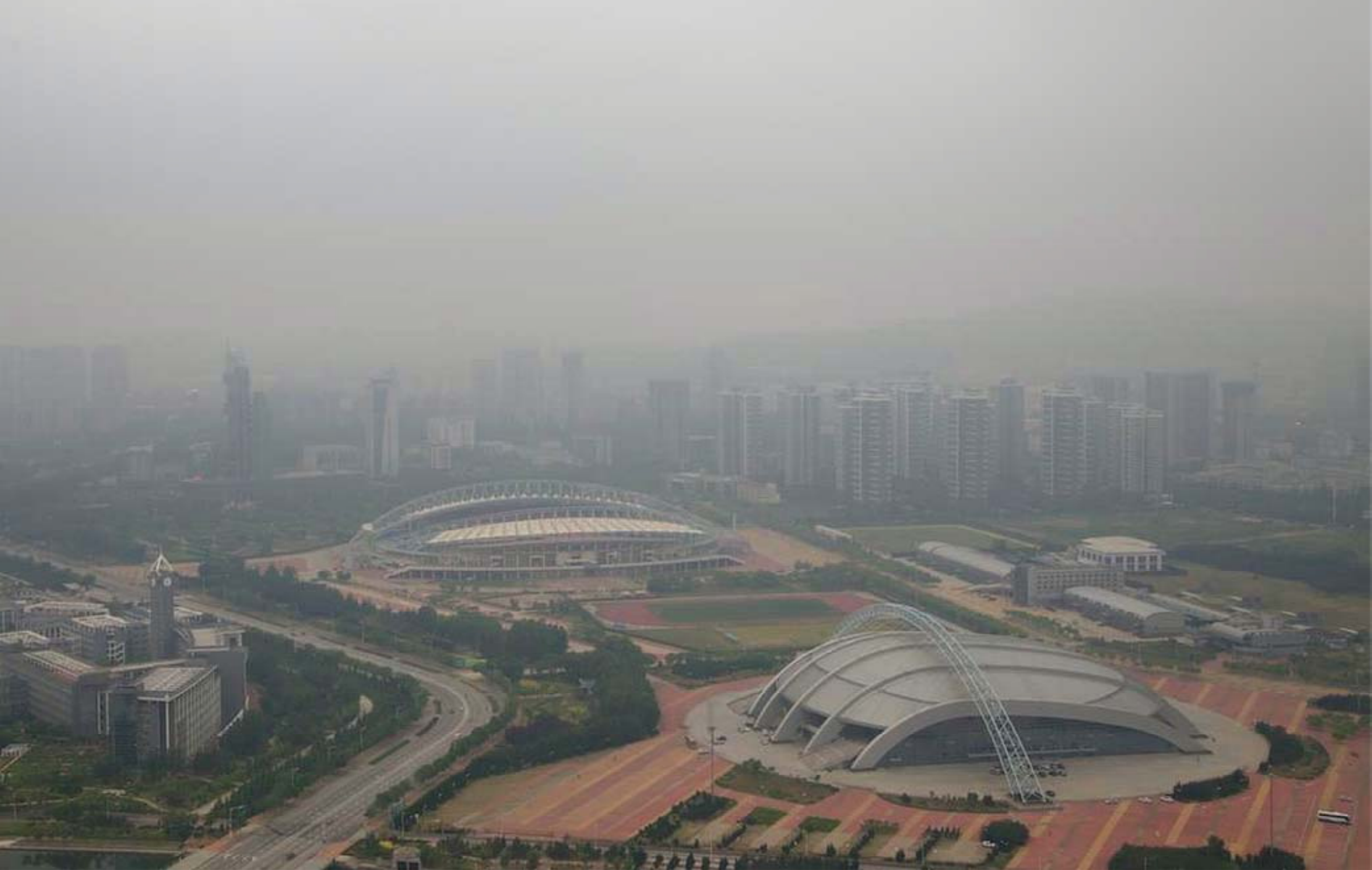}&
		\includegraphics[width=0.192\linewidth]{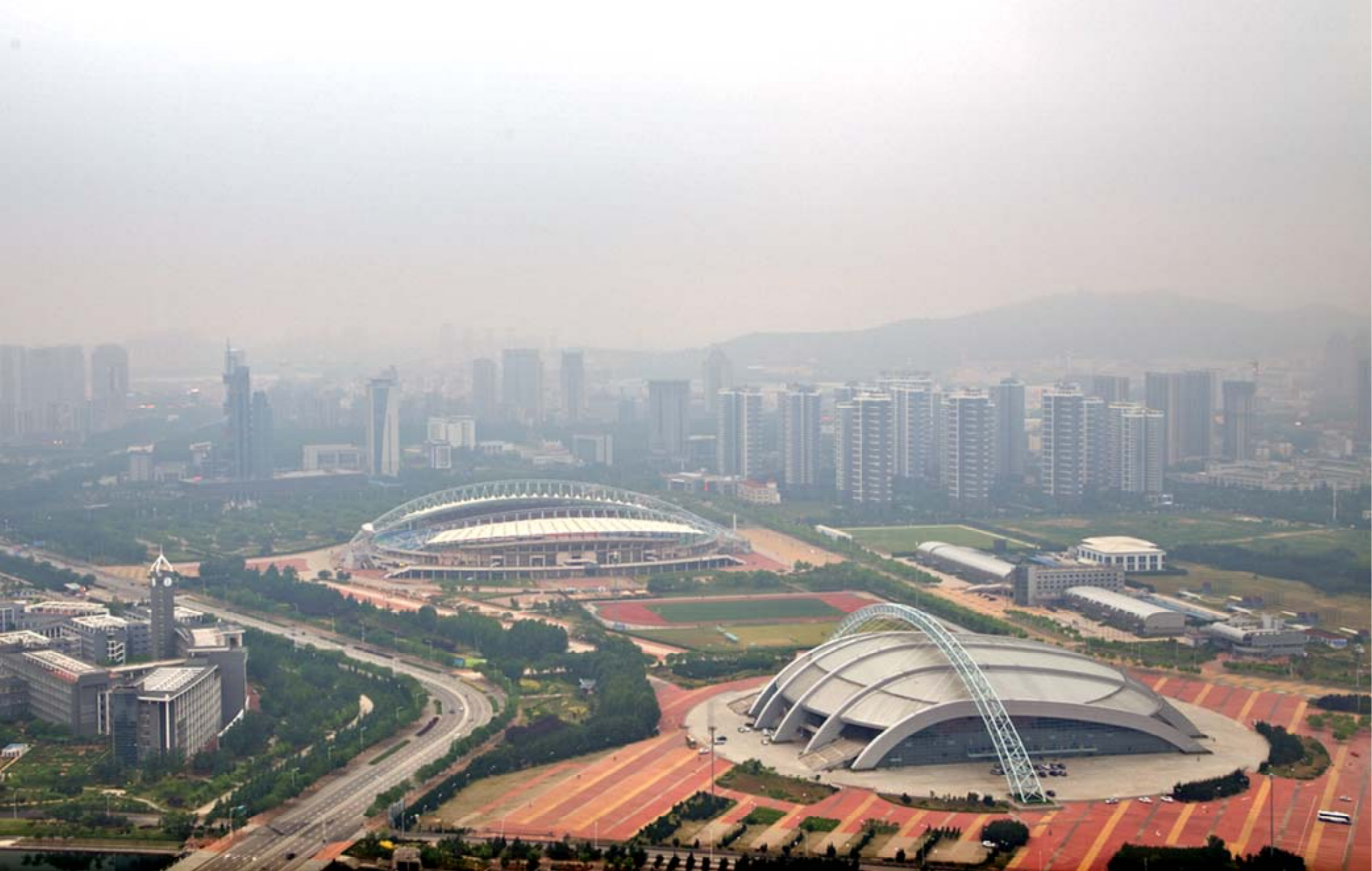}\\
		\includegraphics[width=0.192\linewidth]{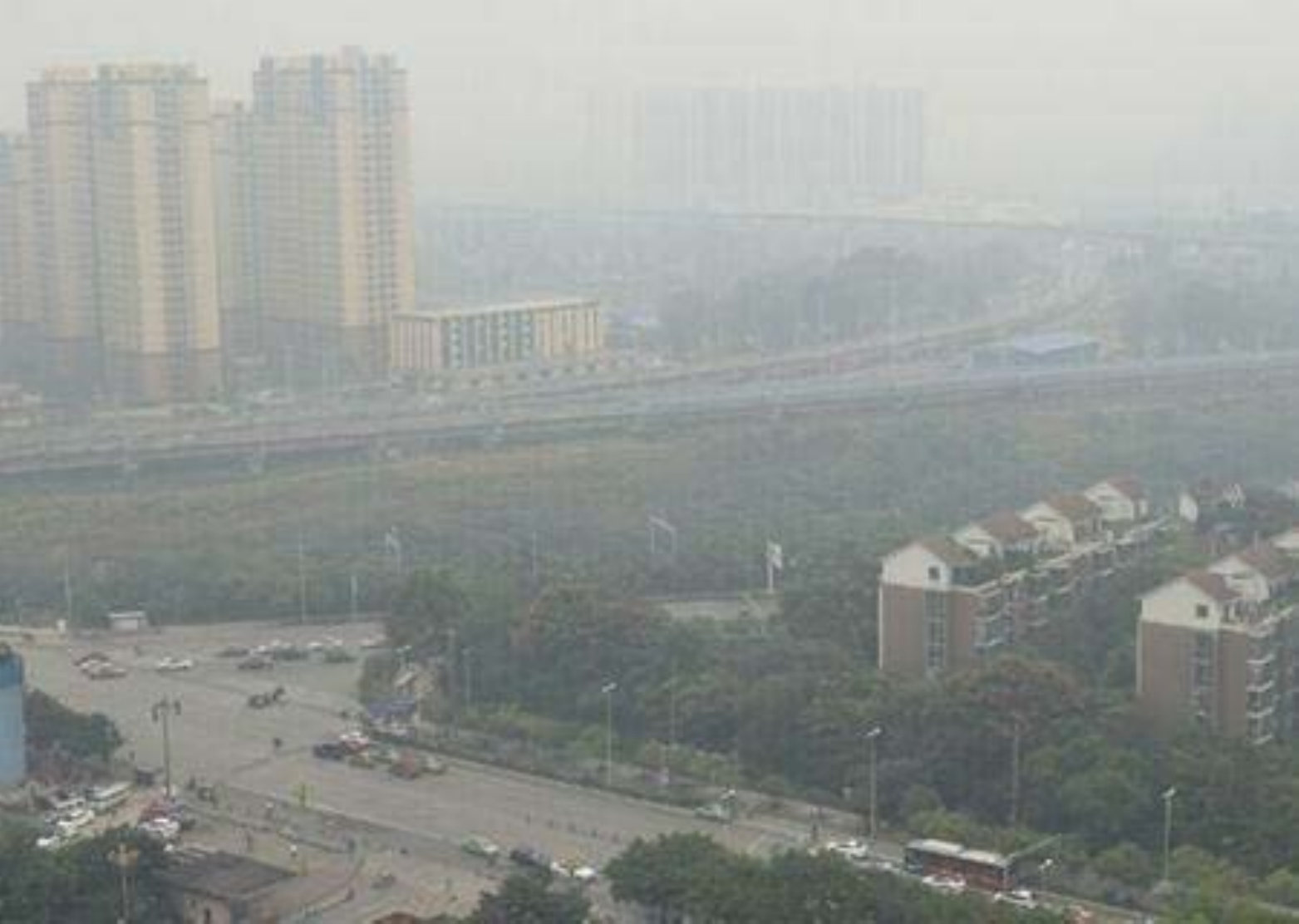}&
		\includegraphics[width=0.192\linewidth]{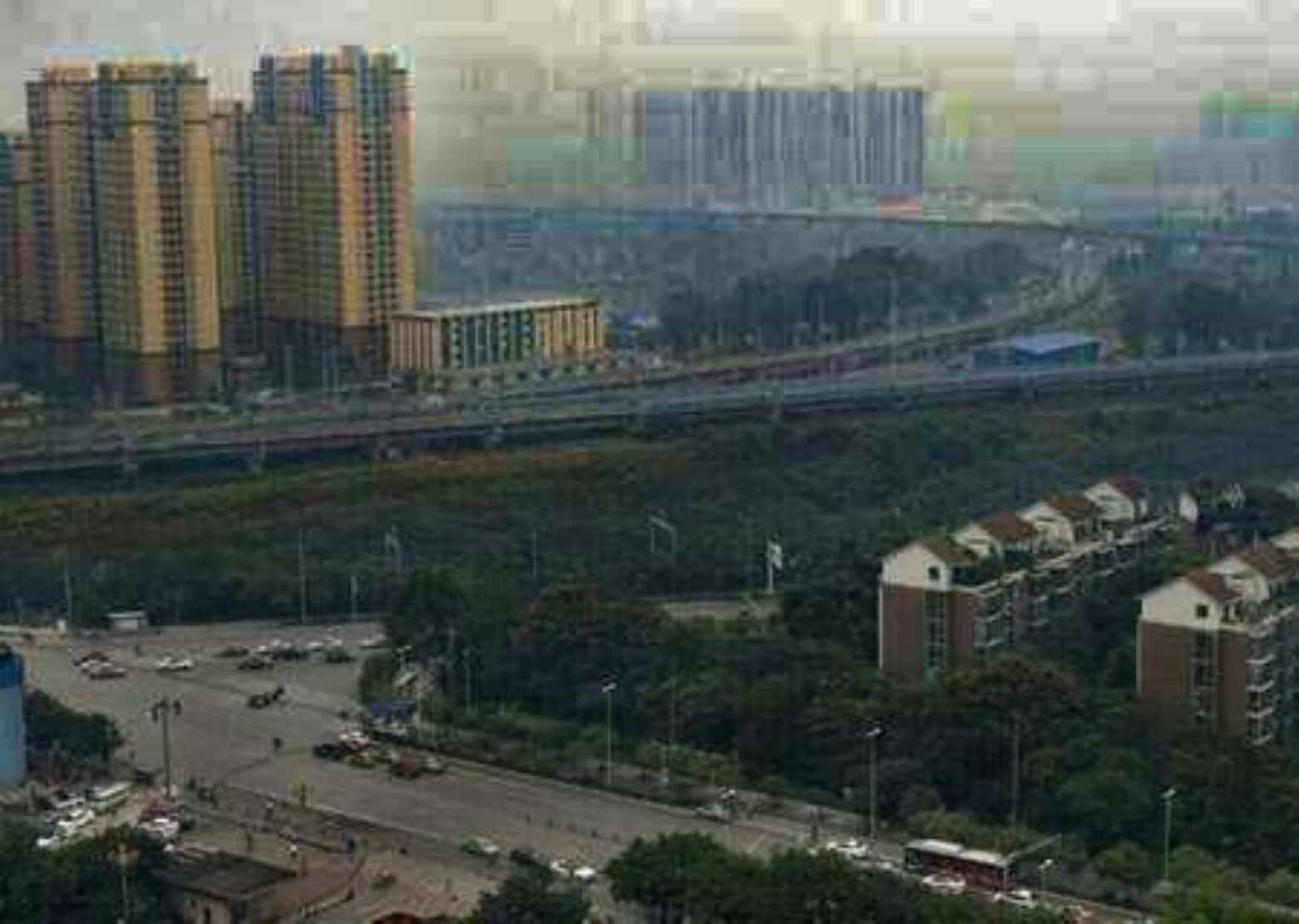}&
		\includegraphics[width=0.192\linewidth]{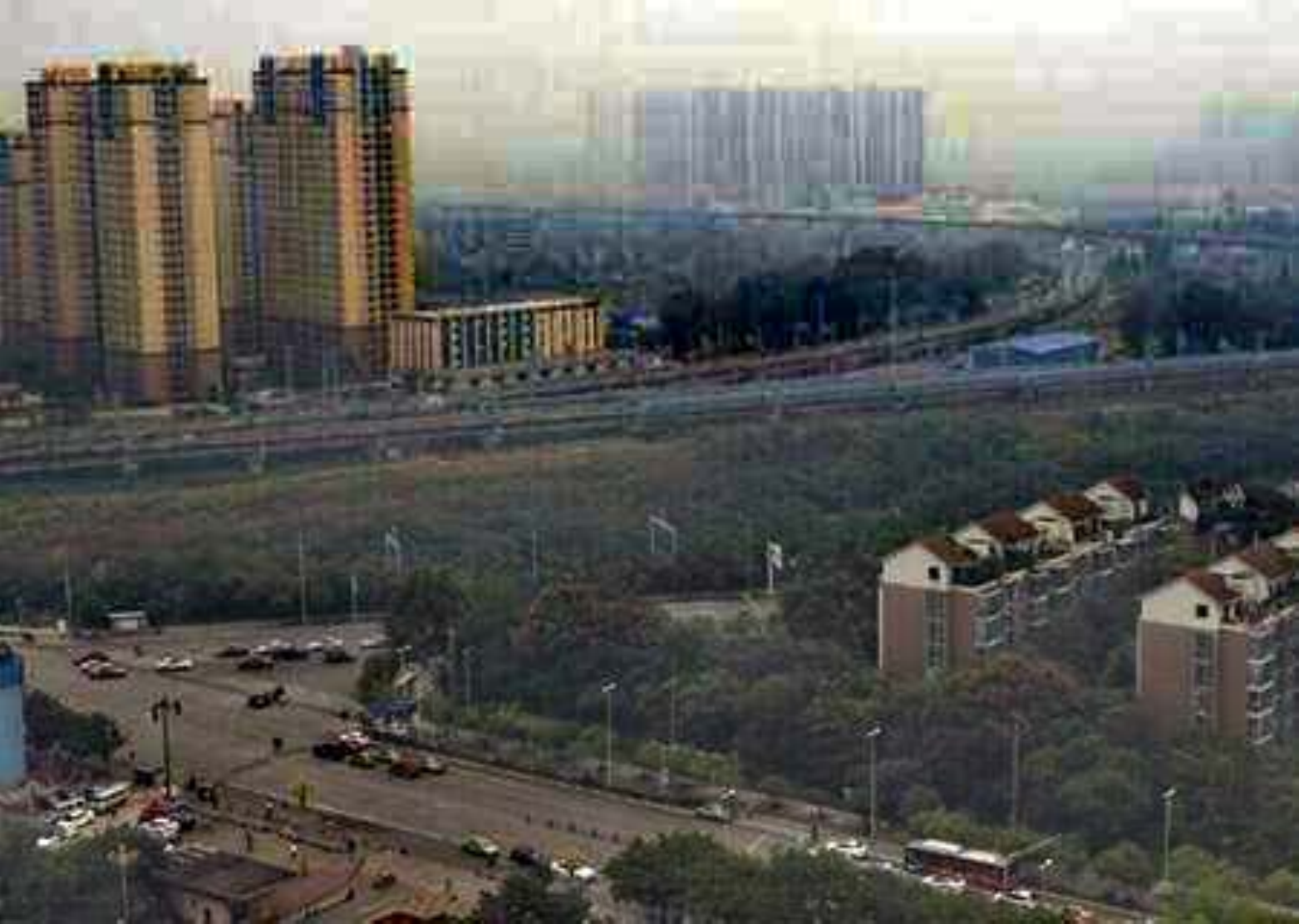}&
		\includegraphics[width=0.192\linewidth]{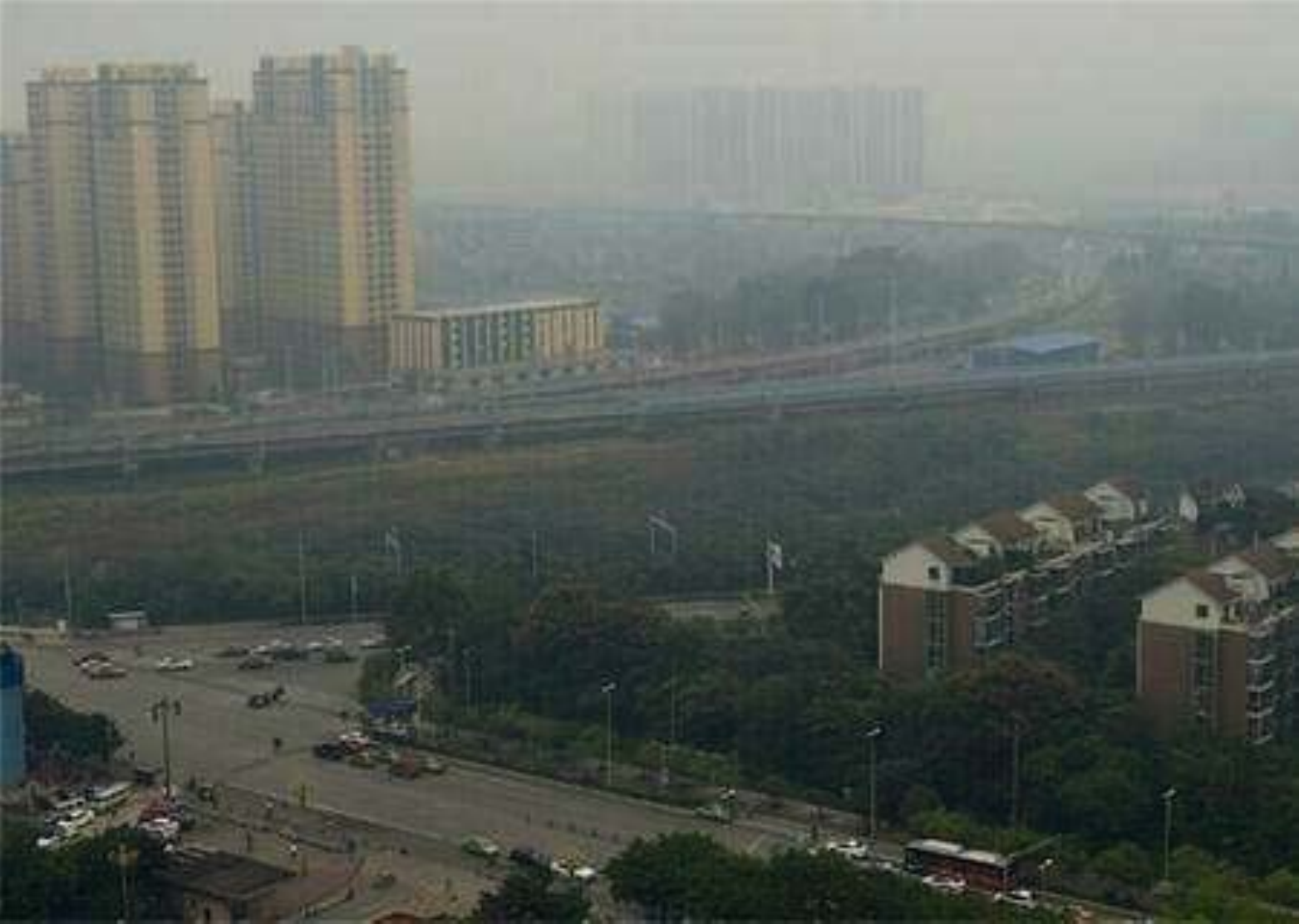}&
		\includegraphics[width=0.192\linewidth]{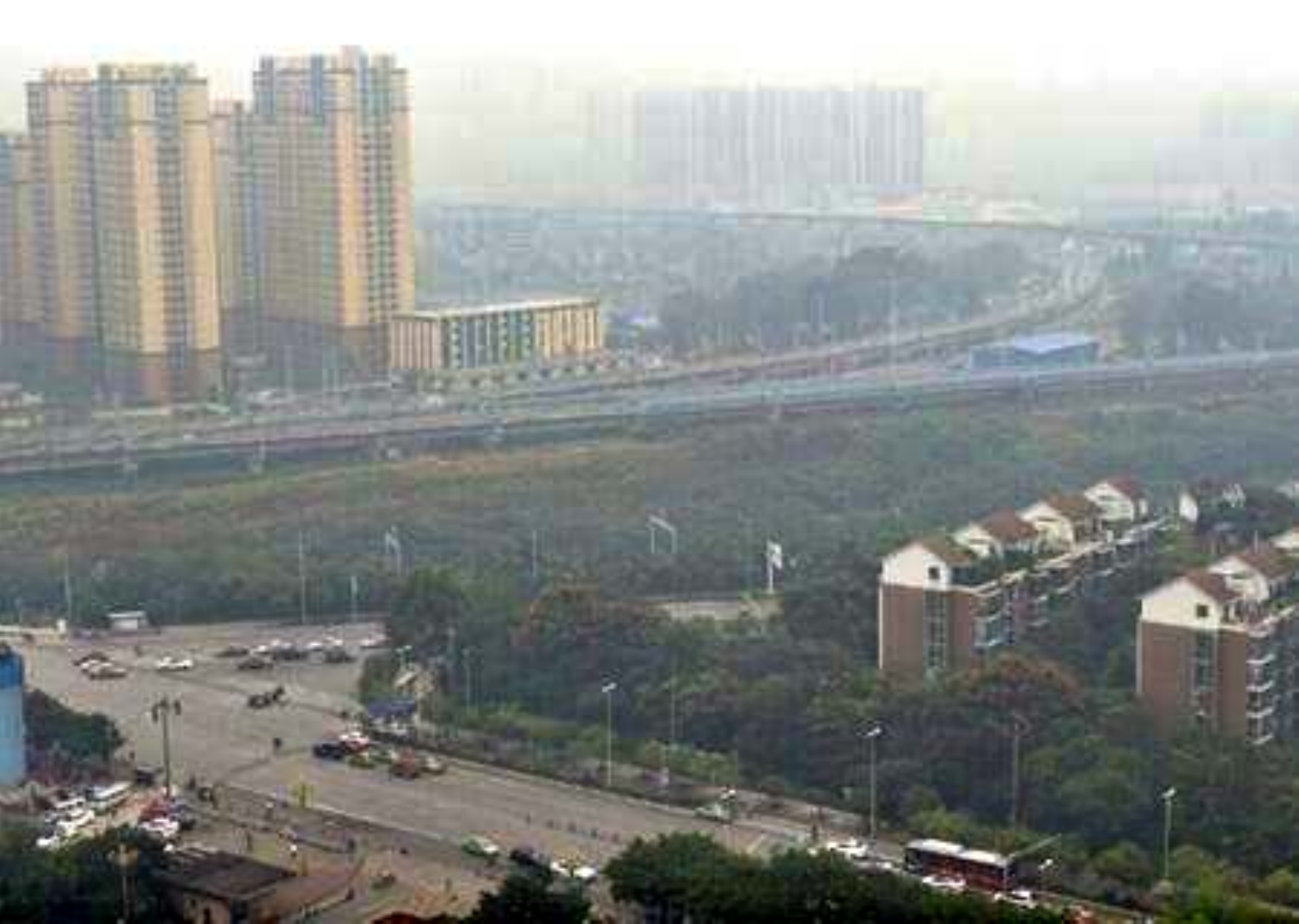}\\
		\includegraphics[width=0.192\linewidth]{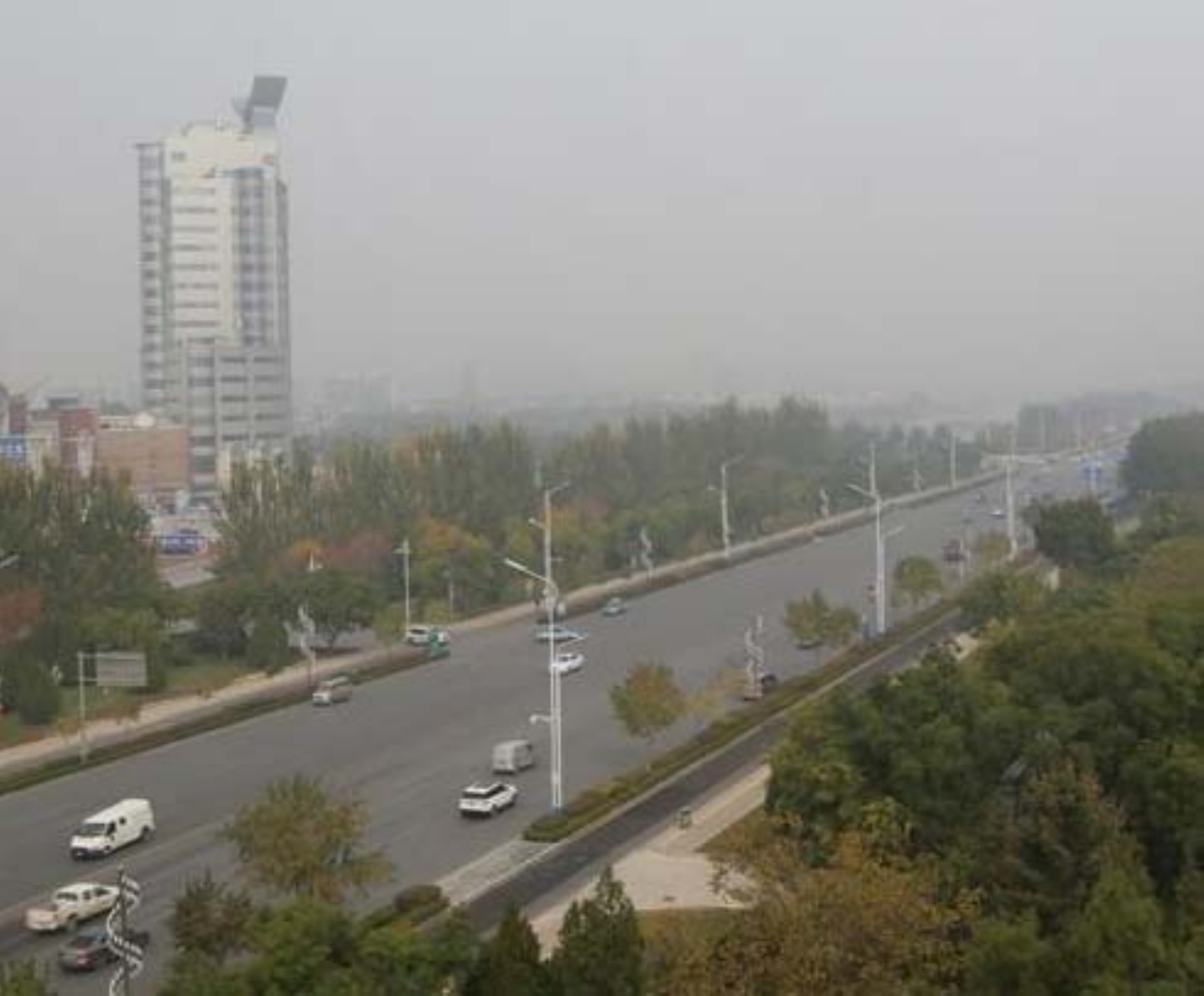}&
		\includegraphics[width=0.192\linewidth]{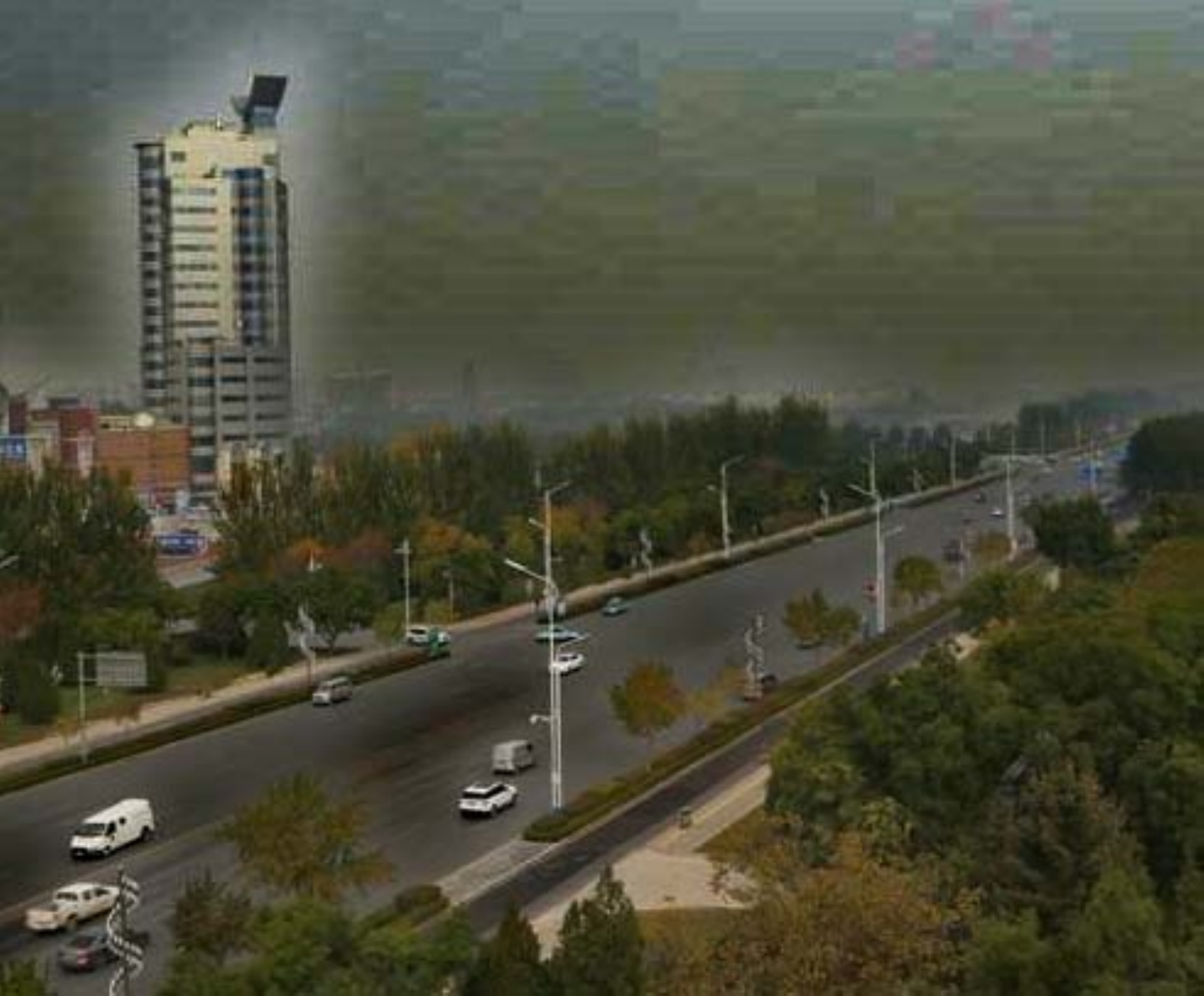}&
		\includegraphics[width=0.192\linewidth]{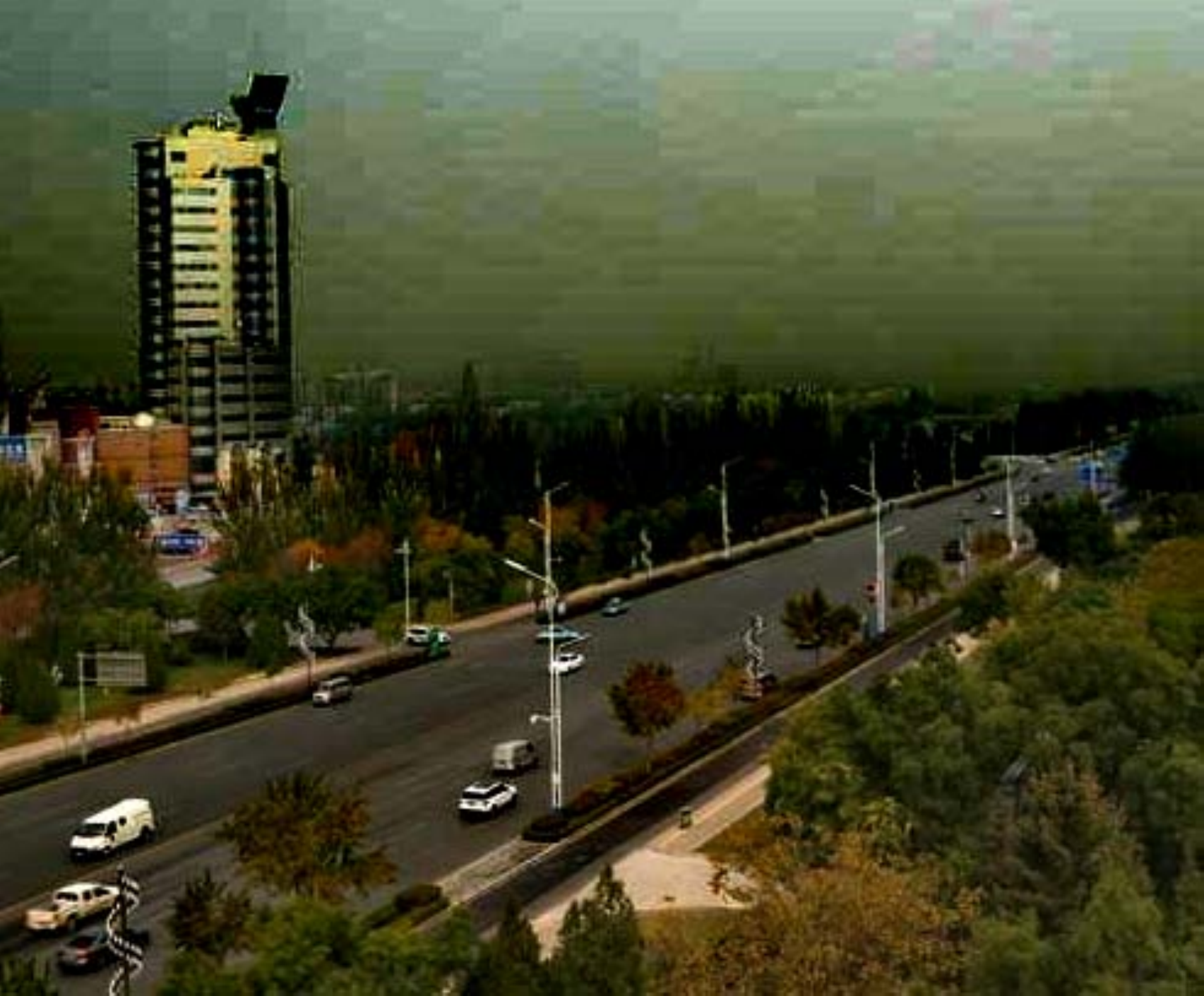}&
		\includegraphics[width=0.192\linewidth]{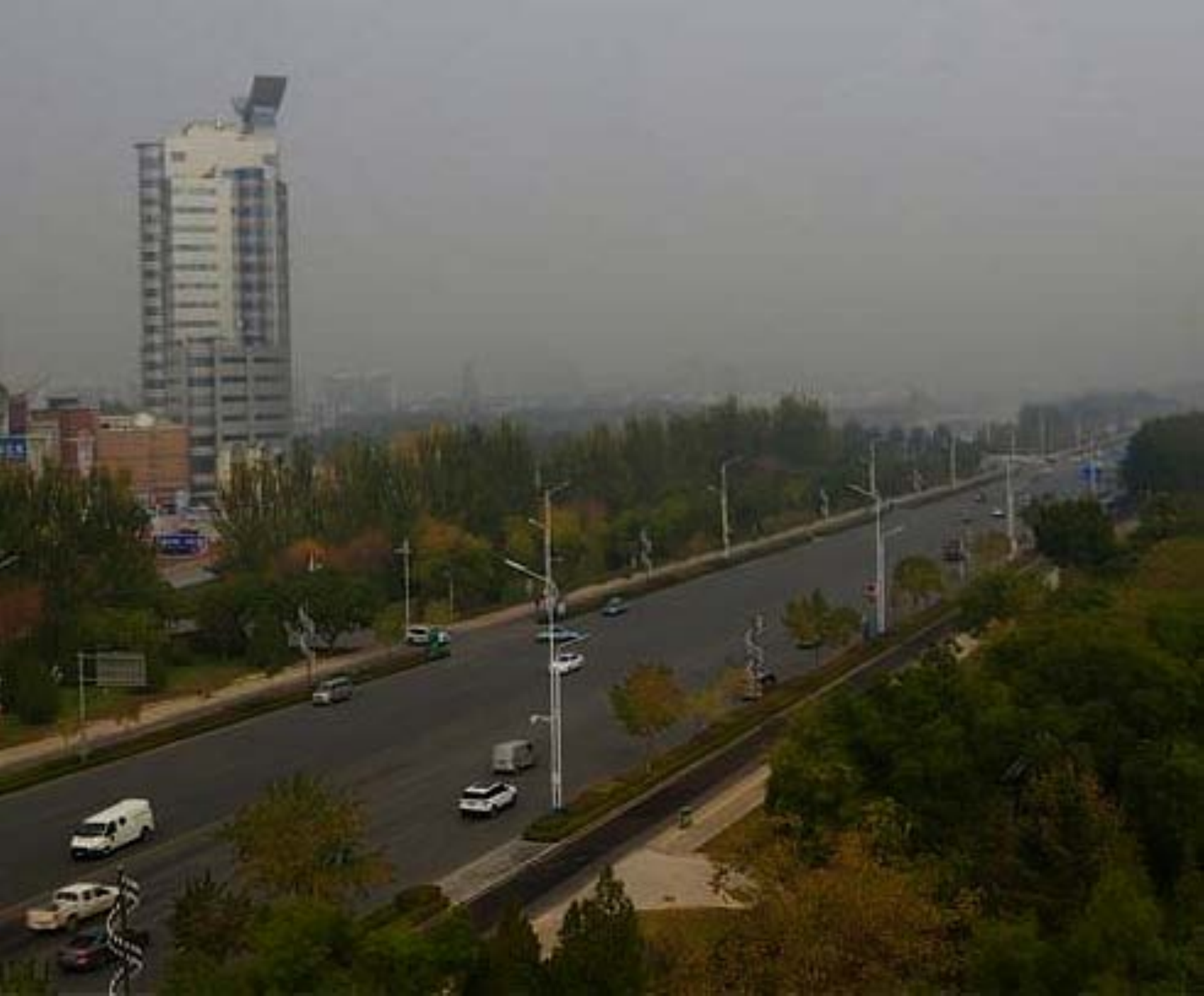}&
		\includegraphics[width=0.192\linewidth]{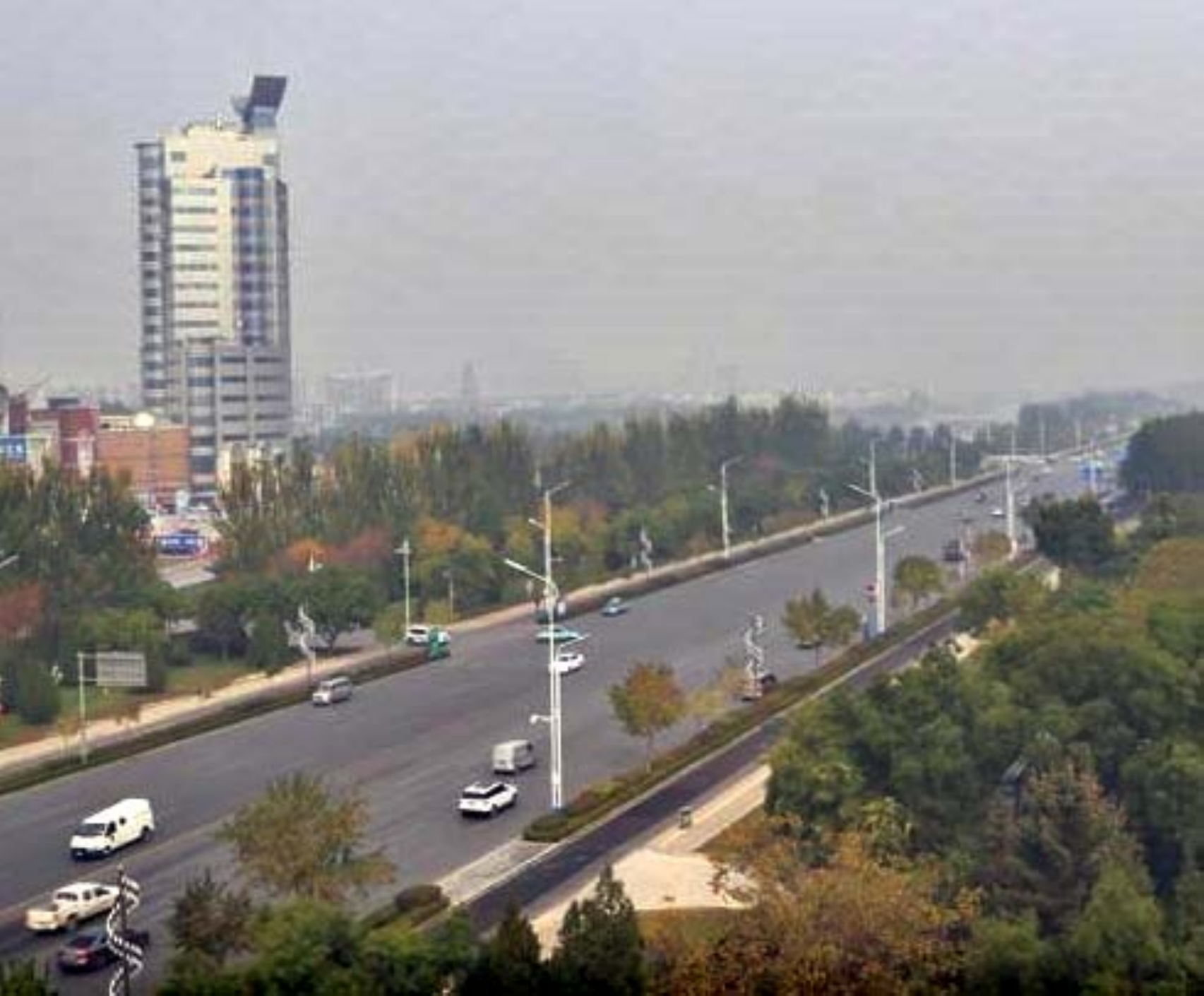}\\
		\includegraphics[width=0.192\linewidth]{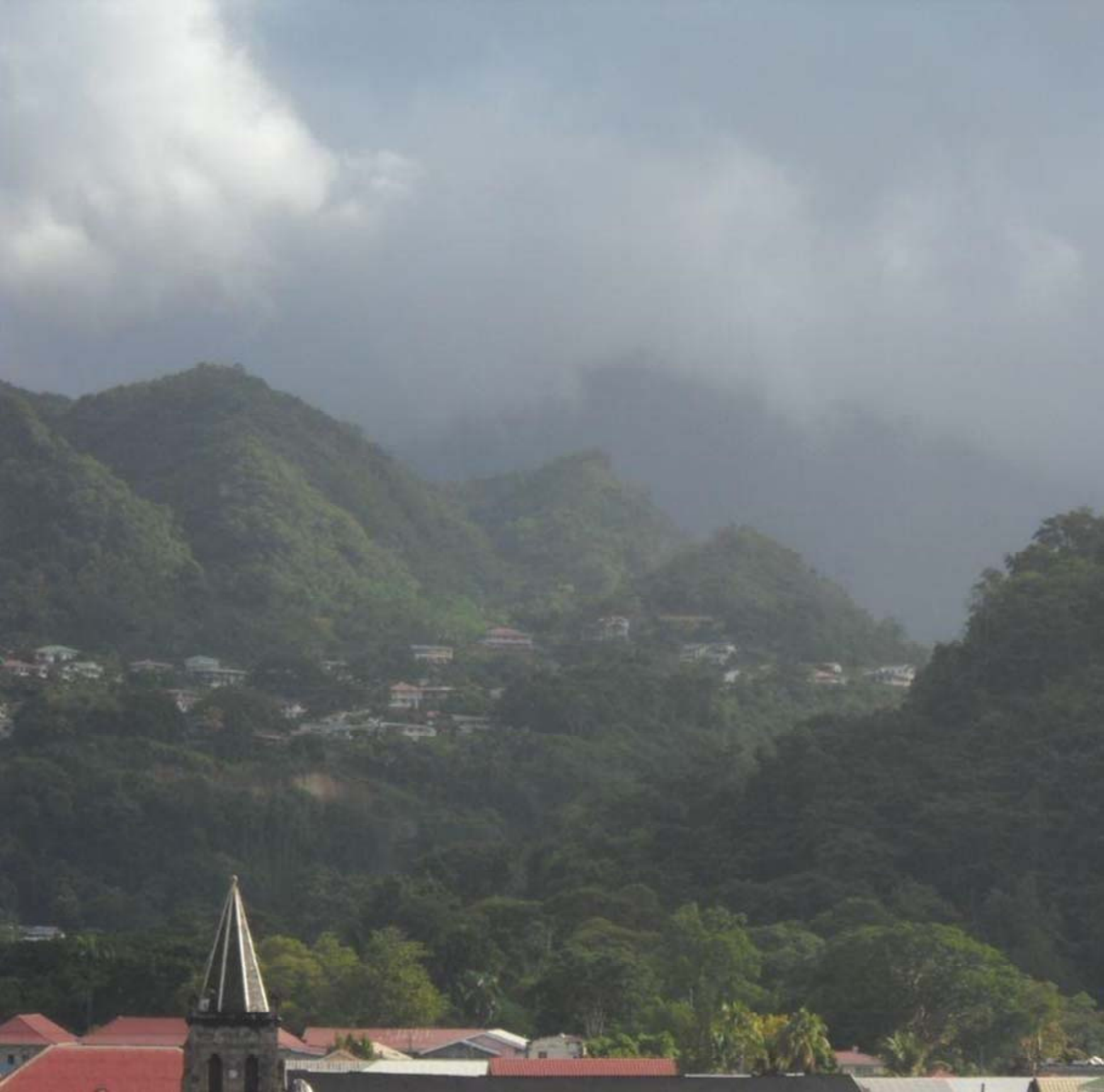}&
		\includegraphics[width=0.192\linewidth]{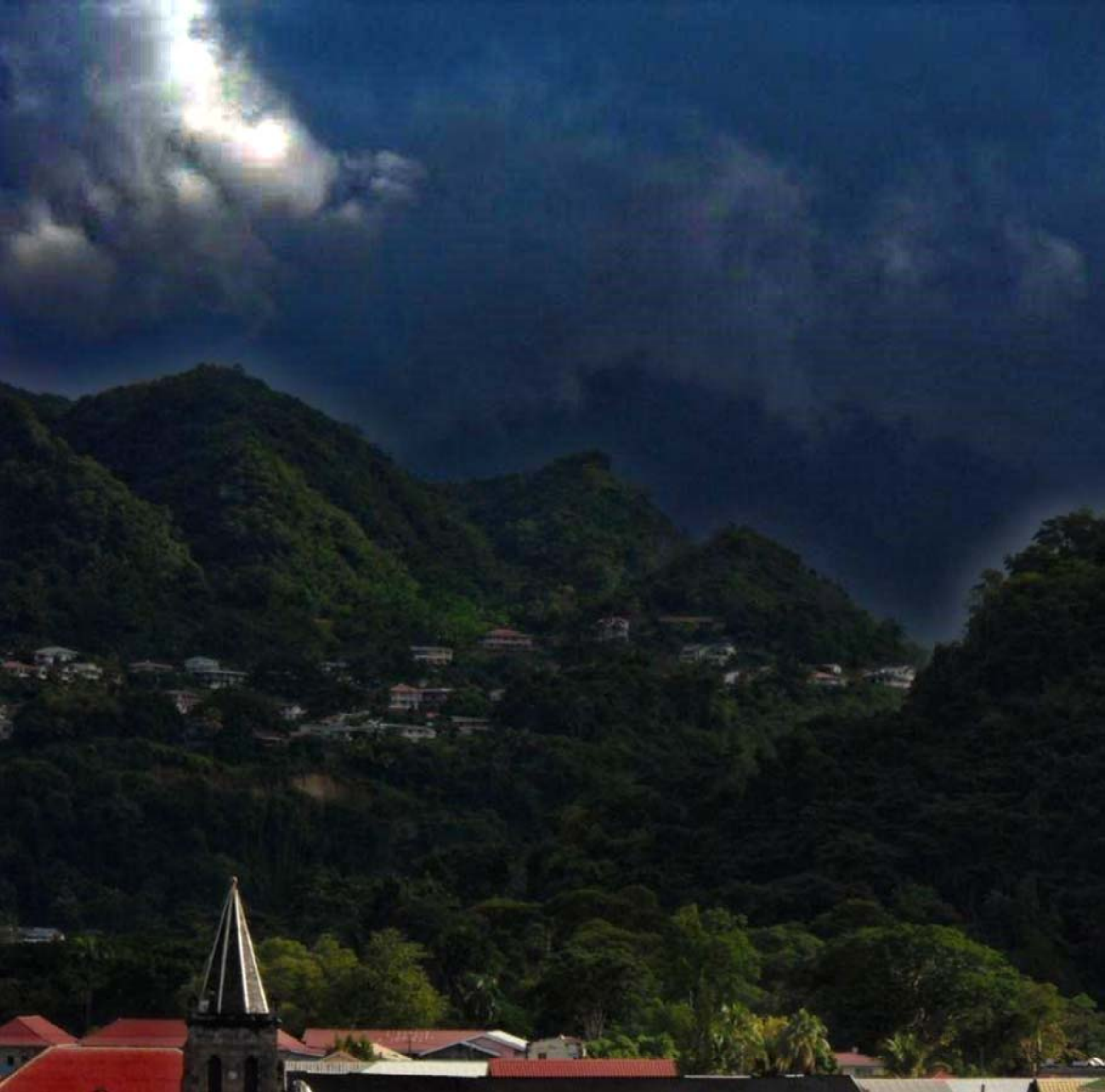}&
		\includegraphics[width=0.192\linewidth]{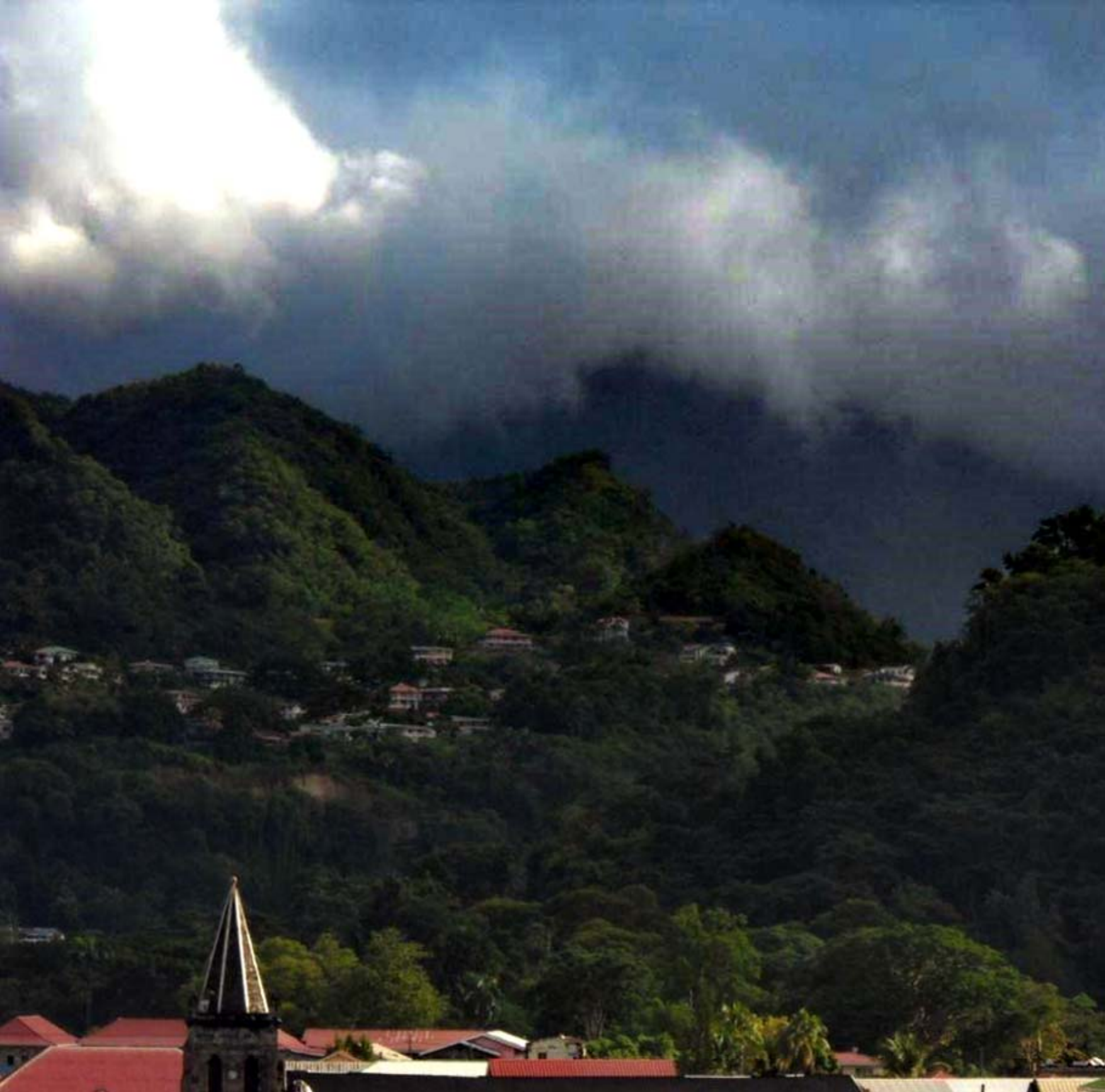}&
		\includegraphics[width=0.192\linewidth]{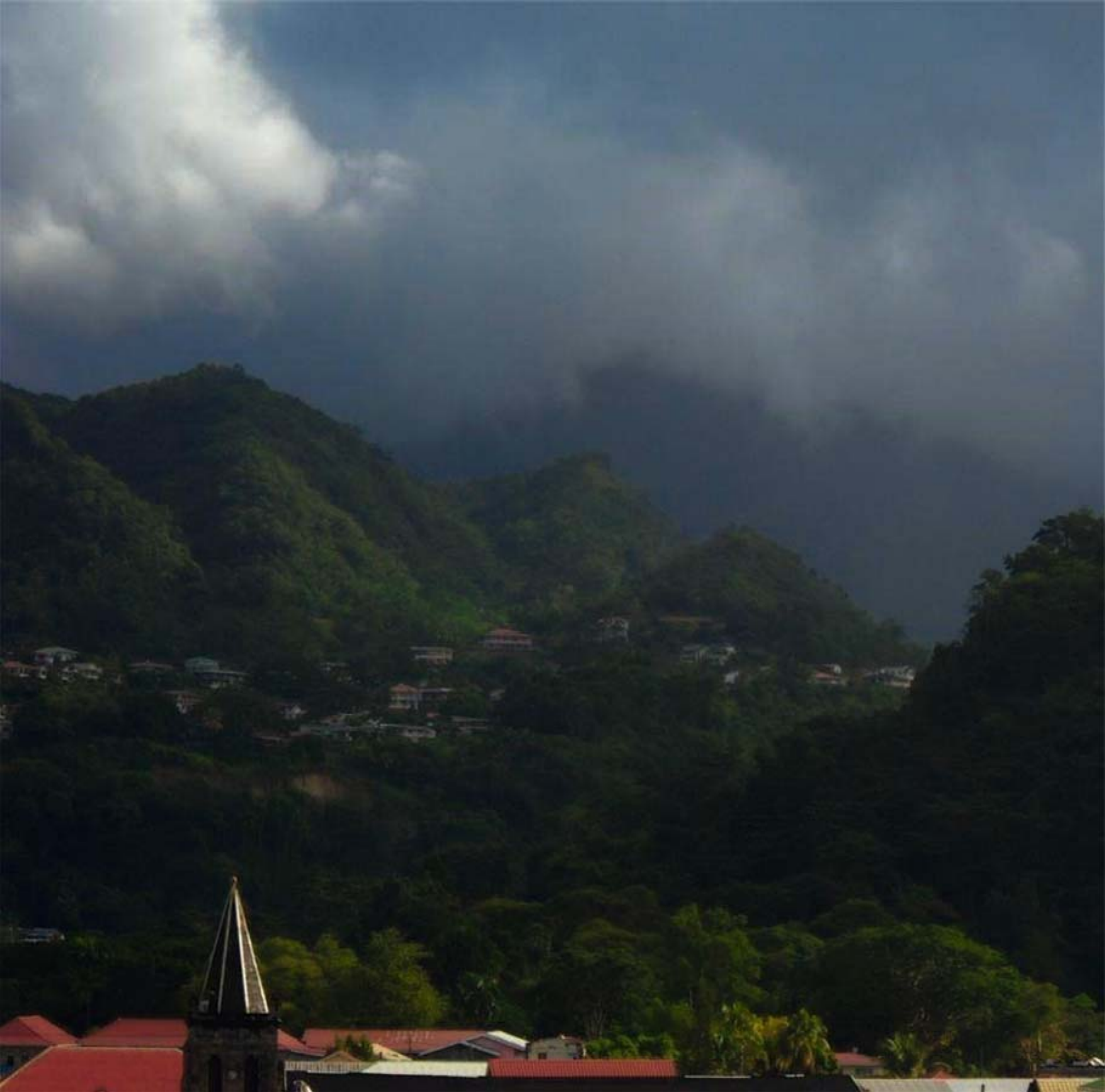}&
		\includegraphics[width=0.192\linewidth]{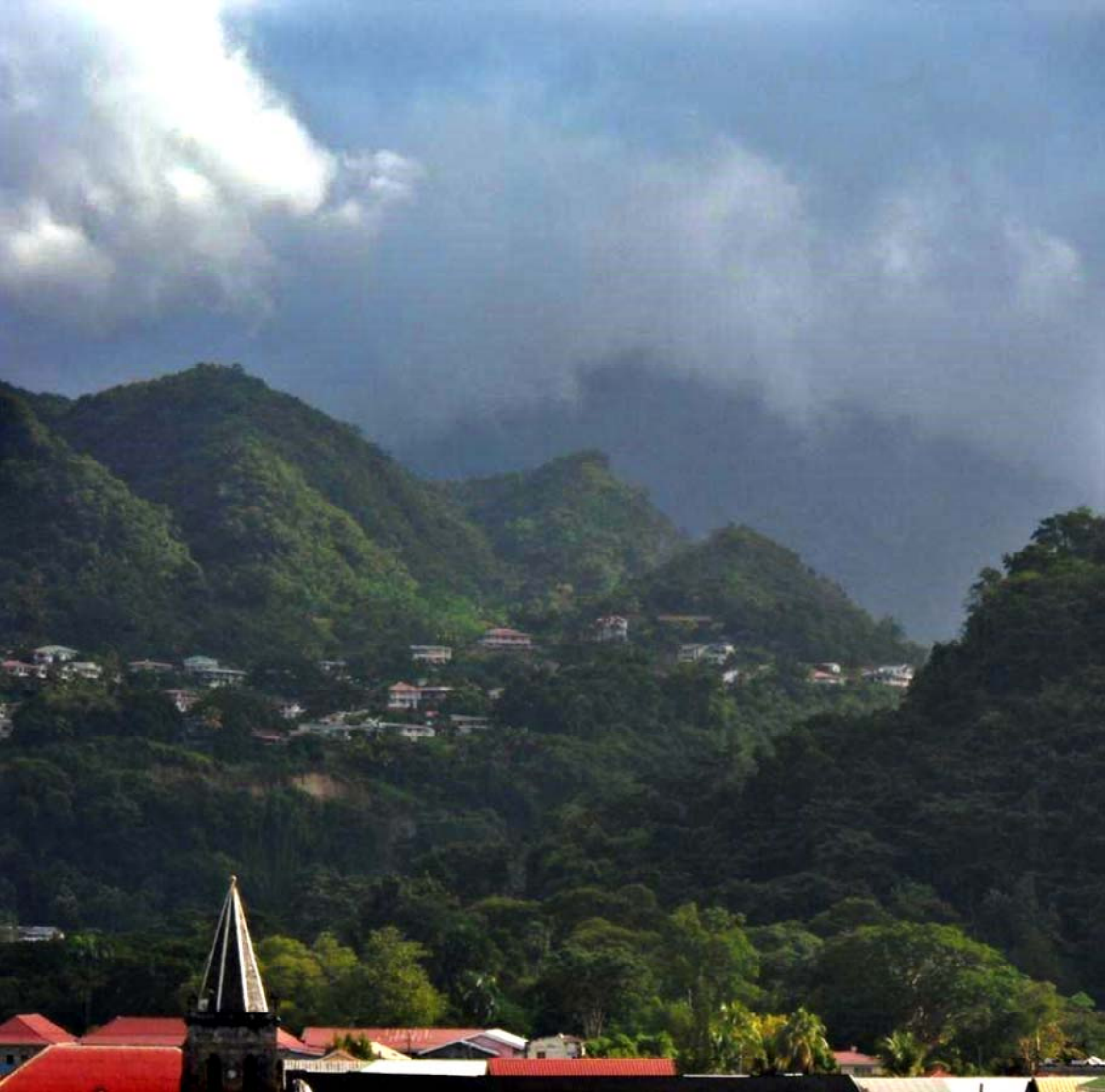}\\
		Input   &DCP& NLD&AODNet& Ours\\
	\end{tabular}
	\caption{Visual comparisons of single image haze removal on real-world scenario.}
	\label{fig:dehazing}
\end{figure*}

\subsection{Single Image Haze Removal}
Finally, we further conduct an experiment of single image haze removal to verify the effectiveness of our proposed framework. We follow the duality of Retinex and image dehazing~\cite{galdran2017duality}, to execute this task. The visual comparisons with three representative methods (i.e., the classical dehazing method, DCP~\cite{he2011single}, the traditional optimization based method, NLD~\cite{berman2016non} and the end-to-end network, AODNet~\cite{li2017aod}) are presented in Fig.~\ref{fig:dehazing}. Obviously, traditional methods (i.e., DCP, NLD) generate the dehazing results with many artifacts and color distortion, which indicates that only depending on prior regularization is extremely hard to achieve the desired results in real complex scenarios. While AODNet (network-based method) presents unclear details depict and underexposure results. By comparison, it can be easily seen that our results have the prominent visual expression and more natural than other state-of-the-art approaches.


\section{Conclusions}
In this paper, we provided a new perspective to formulate the problem of underexposed image correction within hybrid priors (i.e., knowledge and learnable architectures) energy-inspired model. We designed a propagation procedure navigated by the hybrid priors to simultaneously propagate the reflectance and illumination. Extensive experiments of underexposed image correction validated that the effectiveness and superiority of our method against other state-of-the-art approaches both in the enhanced effects and running time. Subsequently, face detection task is executed to further verify the naturalness and practical values of our proposed hybrid priors navigated deep propagation. Finally, the single image haze removal task is also performed to illustrate our excellence again other state-of-the-art approaches.


%

%
%
%
\section*{Acknowledgment}
This work is partially supported by the National Natural Science Foundation of China (Nos. 61672125, 61300086, and 61632019), the Fundamental Research Funds for the Central Universities	.
%
%
%



\bibliographystyle{IEEEtran}
\bibliography{reference}

\begin{thebibliography}{10}
\providecommand{\url}[1]{#1}
\csname url@samestyle\endcsname
\providecommand{\newblock}{\relax}
\providecommand{\bibinfo}[2]{#2}
\providecommand{\BIBentrySTDinterwordspacing}{\spaceskip=0pt\relax}
\providecommand{\BIBentryALTinterwordstretchfactor}{4}
\providecommand{\BIBentryALTinterwordspacing}{\spaceskip=\fontdimen2\font plus
\BIBentryALTinterwordstretchfactor\fontdimen3\font minus
  \fontdimen4\font\relax}
\providecommand{\BIBforeignlanguage}[2]{{%
\expandafter\ifx\csname l@#1\endcsname\relax
\typeout{** WARNING: IEEEtran.bst: No hyphenation pattern has been}%
\typeout{** loaded for the language `#1'. Using the pattern for}%
\typeout{** the default language instead.}%
\else
\language=\csname l@#1\endcsname
\fi
#2}}
\providecommand{\BIBdecl}{\relax}
\BIBdecl

\bibitem{wang2005Brightness}
C.~Wang and Z.~Ye, ``Brightness preserving histogram equalization with maximum
  entropy: a variational perspective,'' \emph{IEEE Transactions on Consumer
  Electronics}, vol.~51, no.~4, pp. 1326--1334, 2005.

\bibitem{sheet2010brightness}
D.~Sheet, H.~Garud, A.~Suveer, M.~Mahadevappa, and J.~Chatterjee, ``Brightness
  preserving dynamic fuzzy histogram equalization,'' \emph{IEEE Transactions on
  Consumer Electronics}, vol.~56, no.~4, 2010.

\bibitem{guo2017lime}
X.~Guo, Y.~Li, and H.~Ling, ``Lime: Low-light image enhancement via
  illumination map estimation,'' \emph{IEEE Transactions on Image Processing},
  vol.~26, no.~2, pp. 982--993, 2017.

\bibitem{zhang2018high}
Q.~Zhang, W.-S. Zheng, G.~Yuan, C.~Xiao, and L.~Zhu, ``High-quality exposure
  correction of underexposed photos,'' in \emph{ACM Multimedia}, 2018.

\bibitem{li2018structure}
M.~Li, J.~Liu, W.~Yang, X.~Sun, and Z.~Guo, ``Structure-revealing low-light
  image enhancement via robust retinex model,'' \emph{IEEE Transactions on
  Image Processing}, vol.~27, no.~6, pp. 2828--2841, 2018.

\bibitem{hasinoff2017Deep}
M.~Gharbi, J.~Chen, J.~T. Barron, S.~W. Hasinoff, and F.~Durand, ``Deep
  bilateral learning for real-time image enhancement,'' \emph{ACM Transactions
  on Graphics}, vol.~36, no.~4, p. 118, 2017.

\bibitem{Chen2018Retinex}
C.~Wei, W.~Wang, W.~Yang, and J.~Liu, ``Deep retinex decomposition for
  low-light enhancement,'' in \emph{British Machine Vision Conference}, 2018.

\bibitem{cheng2004simple}
H.~Cheng and X.~Shi, ``A simple and effective histogram equalization approach
  to image enhancement,'' \emph{Digital Signal Processing}, vol.~14, no.~2, pp.
  158--170, 2004.

\bibitem{power2011Multi}
M.~J. Power, C.~Whitlock, P.~Bartlein, and L.~R. Stevens,
  ``Multi-histransactions on graphicsram equalization methods for contrast
  enhancement and brightness preserving,'' \emph{IEEE Transactions on Consumer
  Electronics}, vol.~53, no.~3, pp. 1186--1194, 2011.

\bibitem{yun2011Contrast}
S.~H. Yun, H.~K. Jin, and S.~Kim, ``Contrast enhancement using a weighted
  histransactions on graphicsram equalization,'' in \emph{IEEE International
  Conference on Consumer Electronics}, vol.~10, no.~11, 2011, pp. 203--204.

\bibitem{McCann2016}
J.~McCann, \emph{Retinex Theory}.\hskip 1em plus 0.5em minus 0.4em\relax
  Springer New York, 2016.

\bibitem{cai2017joint}
B.~Cai, X.~Xu, K.~Guo, K.~Jia, B.~Hu, and D.~Tao, ``A joint intrinsic-extrinsic
  prior model for retinex,'' in \emph{Proceeding of the IEEE Conference on
  Computer Vision and Pattern Recognition}, 2017.

\bibitem{land1971lightness}
E.~H. Land and J.~J. McCann, ``Lightness and retinex theory,'' \emph{Journal of
  the Optical Society of America}, vol.~61, no.~1, pp. 1--11, 1971.

\bibitem{kimmel2003variational}
R.~Kimmel, M.~Elad, D.~Shaked, R.~Keshet, and I.~Sobel, ``A variational
  framework for retinex,'' \emph{International Journal of Computer Vision},
  vol.~52, no.~1, pp. 7--23, 2003.

\bibitem{ng2011total}
M.~K. Ng and W.~Wang, ``A total variation model for retinex,'' \emph{SIAM
  Journal on Imaging Sciences}, vol.~4, no.~1, pp. 345--365, 2011.

\bibitem{fu2015probabilistic}
X.~Fu, Y.~Liao, D.~Zeng, Y.~Huang, X.-P. Zhang, and X.~Ding, ``A probabilistic
  method for image enhancement with simultaneous illumination and reflectance
  estimation,'' \emph{IEEE Transactions on Image Processing}, vol.~24, no.~12,
  pp. 4965--4977, 2015.

\bibitem{fu2016weighted}
X.~Fu, D.~Zeng, Y.~Huang, X.-P. Zhang, and X.~Ding, ``A weighted variational
  model for simultaneous reflectance and illumination estimation,'' in
  \emph{Proceeding of the IEEE Conference on Computer Vision and Pattern
  Recognition}, 2016.

\bibitem{dmitry2018Deep}
D.~Ulyanov, A.~Vedaldi, and V.~S. Lempitsky, ``Deep image prior,'' in
  \emph{Proceeding of the IEEE Conference on Computer Vision and Pattern
  Recognition}, 2018.

\bibitem{kim2015Accurate}
J.~Kim, J.~K. Lee, and K.~M. Lee, ``Accurate image super-resolution using very
  deep convolutional networks,'' in \emph{Proceeding of the IEEE Conference on
  Computer Vision and Pattern Recognition}, 2016, pp. 1646--1654.

\bibitem{zhang2018residual}
Y.~Zhang, Y.~Tian, Y.~Kong, B.~Zhong, and Y.~Fu, ``Residual dense network for
  image super-resolution,'' in \emph{Proceedings of the IEEE Conference on
  Computer Vision and Pattern Recognition}, 2018, pp. 2472--2481.

\bibitem{liu2018Proximal}
R.~Liu, X.~Fan, S.~Cheng, X.~Wang, and Z.~Luo, ``Proximal alternating direction
  network: A globally converged deep unrolling framework,'' in
  \emph{Association for the Advancement of Artificial Intelligence}, 2018.

\bibitem{liu2018GCM}
R.~Liu, Y.~He, S.~Cheng, X.~Fan, and Z.~Luo, ``Learning collaborative
  generation correction modules for blind image deblurring and beyond,'' in
  \emph{ACM Multimedia}, 2018.

\bibitem{li2017aod}
B.~Li, X.~Peng, Z.~Wang, J.~Xu, and D.~Feng, ``Aod-net: All-in-one dehazing
  network,'' in \emph{Proceedings of the IEEE International Conference on
  Computer Vision}, 2017.

\bibitem{liu2018learning}
R.~Liu, L.~Ma, Y.~Wang, and L.~Zhang, ``Learning converged propagations with
  deep prior ensemble for image enhancement,'' \emph{IEEE Transactions on Image
  Processing}, vol.~28, no.~3, pp. 1528--1543, 2018.

\bibitem{liu2019deep}
R.~Liu, S.~Cheng, L.~Ma, X.~Fan, and Z.~Luo, ``Deep proximal unrolling:
  Algorithmic framework, convergence analysis and applications,'' \emph{IEEE
  Transactions on Image Processing}, 2019.

\bibitem{liu2019convergence}
R.~Liu, S.~Cheng, Y.~He, X.~Fan, Z.~Lin, and Z.~Luo, ``On the convergence of
  learning-based iterative methods for nonconvex inverse problems,'' \emph{IEEE
  transactions on pattern analysis and machine intelligence}, 2019.

\bibitem{lore2017llnet}
K.~G. Lore, A.~Akintayo, and S.~Sarkar, ``Llnet: A deep autoencoder approach to
  natural low-light image enhancement,'' \emph{Pattern Recognition}, vol.~61,
  pp. 650--662, 2017.

\bibitem{cai2018Learning}
J.~Cai, S.~Gu, and L.~Zhang, ``Learning a deep single image contrast enhancer
  from multi-exposure images,'' \emph{IEEE Transactions on Image Processing},
  vol.~27, no.~4, pp. 2049--2062, 2018.

\bibitem{Zhang2019MM}
Y.~Zhang, J.~Zhang, and X.~Guo, ``Kindling the darkness: A practical low-light
  image enhancer,'' in \emph{ACM Multimedia}, 2019.

\bibitem{Roth2009Fields}
S.~Roth and M.~J. Black, ``Fields of experts,'' \emph{International Journal of
  Computer Vision}, vol.~82, no.~2, pp. 205--229, 2009.

\bibitem{poynton2012digital}
C.~Poynton, \emph{Digital video and HD: Algorithms and Interfaces}, 2012.

\bibitem{krizhevsky2012imagenet}
A.~Krizhevsky, I.~Sutskever, and G.~E. Hinton, ``Imagenet classification with
  deep convolutional neural networks,'' in \emph{Neural Information Processing
  Systems}, 2012.

\bibitem{mittal2013making}
A.~Mittal, R.~Soundararajan, and A.~C. Bovik, ``Making a ¡°completely blind¡±
  image quality analyzer,'' \emph{IEEE Signal Processing Letters}, vol.~20,
  no.~3, pp. 209--212, 2013.

\bibitem{redmon2018yolov3}
J.~Redmon and A.~Farhadi, ``Yolov3: An incremental improvement,'' \emph{arXiv
  preprint arXiv:1804.02767}, 2018.

\bibitem{rahman2004retinex}
Z.-u. Rahman, D.~J. Jobson, and G.~A. Woodell, ``Retinex processing for
  automatic image enhancement,'' \emph{Journal of Electronic Imaging}, vol.~13,
  no.~1, pp. 100--111, 2004.

\bibitem{shan2010globally}
Q.~Shan, J.~Jia, and M.~S. Brown, ``Globally optimized linear windowed tone
  mapping,'' \emph{IEEE Transactions on Visualization and Computer Graphics},
  vol.~16, no.~4, pp. 663--675, 2010.

\bibitem{wang2013naturalness}
S.~Wang, J.~Zheng, H.-M. Hu, and B.~Li, ``Naturalness preserved enhancement
  algorithm for non-uniform illumination images,'' \emph{IEEE Transactions on
  Image Processing}, vol.~22, no.~9, pp. 3538--3548, 2013.

\bibitem{hasinoff2016burst}
S.~W. Hasinoff, D.~Sharlet, R.~Geiss, A.~Adams, J.~T. Barron, F.~Kainz,
  J.~Chen, and M.~Levoy, ``Burst photography for high dynamic range and
  low-light imaging on mobile cameras,'' \emph{ACM Transactions on Graphics},
  vol.~35, no.~6, 2016.

\bibitem{yang2016wider}
S.~Yang, P.~Luo, C.~C. Loy, and X.~Tang, ``Wider face: A face detection
  benchmark,'' in \emph{Proceedings of the IEEE Conference on Computer Vision
  and Pattern Recognition}, 2016.

\bibitem{galdran2017duality}
A.~Galdran, A.~Alvarez-Gila, A.~Bria, J.~Vazquez-Corral, and M.~Bertalm{\i}o,
  ``On the duality between retinex and image dehazing,'' in \emph{Proceeding of
  the IEEE Conference on Computer Vision and Pattern Recognition}, 2018.

\bibitem{he2011single}
K.~He, J.~Sun, and X.~Tang, ``Single image haze removal using dark channel
  prior,'' \emph{IEEE Transactions on Pattern Analysis and Machine
  Intelligence}, vol.~33, no.~12, pp. 2341--2353, 2011.

\bibitem{berman2016non}
D.~Berman, S.~Avidan \emph{et~al.}, ``Non-local image dehazing,'' in
  \emph{Proceeding of the IEEE Conference on Computer Vision and Pattern
  Recognition}, 2016, pp. 1674--1682.

\end{thebibliography}
%
%
%

%
%
	\begin{IEEEbiography}[{\includegraphics[width=1in,height=1.25in,clip,keepaspectratio]{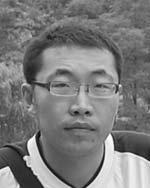}}]{Risheng Liu} (M'12-) received the BSc and PhD degrees both in mathematics from the Dalian University of Technology in 2007 and 2012, respectively. He was a visiting scholar in the Robotic Institute of Carnegie Mellon University from 2010 to 2012. He served as Hong Kong Scholar Research Fellow at
	the Hong Kong Polytechnic University from 2016 to 2017. He is currently an associate professor with the Key Laboratory for Ubiquitous Network and Service Software of Liaoning Province, Internal School of Information and Software Technology, Dalian University of Technology. His research interests include machine learning, optimization, computer vision and multimedia. He was a co-recipient of the IEEE ICME Best Student Paper Award in both 2014 and 2015. Two papers were also selected as Finalist of the Best Paper Award in ICME 2017. He is a member of the IEEE and ACM.		
\end{IEEEbiography}

\begin{IEEEbiography}[{\includegraphics[width=1in,height=1.25in,clip,keepaspectratio]{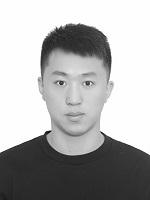}}]{Long Ma} received the B.E. degree in Information and Computing Science from Northeast Agricultural University, Harbin, China, in 2016. He received the M.S. degree in software engineering at Dalian University of Technology, Dalian, China, in 2019. He is currently pursuing the PhD degree in software engineering at Dalian University of Technology, Dalian, China. His research interests include computer vision, image enhancement and machine learning. 
\end{IEEEbiography}

\begin{IEEEbiography}[{\includegraphics[width=1in,height=1.25in,clip,keepaspectratio]{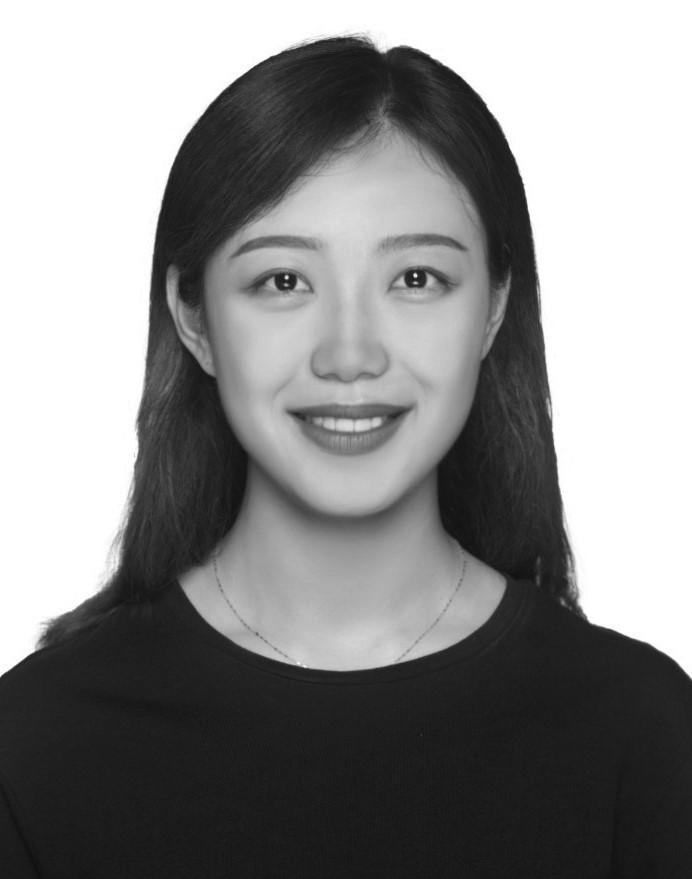}}]{Yuxi Zhang} received the B.E. degree in software engineering from Dalian University of Technology, Dalian, China, in 2017. She is currently pursuing the M.S. degree in software engineering at Dalian University of Technology, Dalian, China. Her research interests include computer vision, image enhancement and machine learning. 
\end{IEEEbiography}

\begin{IEEEbiography}[{\includegraphics[width=1in,height=1.25in,clip,keepaspectratio]{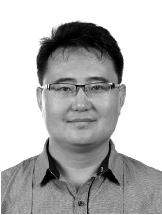}}]{Xin Fan} was born in 1977. He received the B.E. and Ph.D. degrees in information and communication engineering from Xian Jiaotong University, Xian, China, in 1998 and 2004, respectively. He was with Oklahoma State University, Stillwater, from 2006 to 2007, as a post-doctoral research
Fellow. He joined the School of Software, Dalian University of Technology, Dalian, China, in 2009. His current research interests include computational geometry and machine learning, and their applications to lowlevel image processing and DTI-MR image analysis.
\end{IEEEbiography}

\begin{IEEEbiography}[{\includegraphics[width=1in,height=1.25in,clip,keepaspectratio]{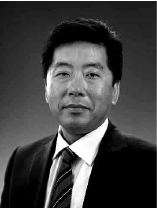}}]{Zhongxuan Luo} received the B.S. degree in Computational Mathematics from Jilin University, China, in 1985, the M.S. degree in Computational
Mathematics from Jilin University in 1988, and the PhD degree in Computational Mathematics from Dalian University of Technology, China, in 1991. He has been a full professor of the School of Mathematical Sciences at Dalian University of Technology since 1997. His research interests include computational geometry and computer vision.
\end{IEEEbiography}




\end{document}